\documentclass[11pt]{article}

\usepackage{amsmath}
\usepackage{amsfonts} 
\usepackage{amssymb}
\usepackage{amsthm}
\usepackage{fullpage}

\usepackage{graphicx}
\usepackage{array}
\usepackage{tabularx}

\usepackage{color}
\usepackage{hyperref}
\usepackage{subfigure}

\usepackage{epstopdf}
\usepackage{epsfig}

\usepackage{xspace} 
\usepackage{indentfirst}

\usepackage{float}
\floatstyle{boxed}
\floatstyle{ruled}
\newfloat{listing}{H}{loi}
\floatname{listing}{Listing}

\ifpdf\DeclareGraphicsExtensions{.pdf,.png,.jpg}\fi

\theoremstyle{remark}

\newcommand{\m}{\bs m} 
\newcommand{\atom}{\psi}  
\newcommand{\Basis}{\mathcal{B}}

\newcommand{\NN}{\mathbb{N}}
\newcommand{\CC}{\mathbb{C}}

\newcommand{\LL}{\mathbb{L}}

\newcommand{\RR}{\mathbb{R}}

\newcommand{\Bb}{\mathcal{B}}

\newcommand{\Dd}{\mathcal{D}}

\newcommand{\Hh}{\mathcal{H}}

\newcommand{\Ll}{\mathcal{L}}
\newcommand{\Mm}{\mathcal{M}}

\newcommand{\Rr}{\mathcal{R}}

\newcommand{\la}{\lambda}

\newcommand{\La}{\Lambda}

\newcommand{\be}{\beta}

\renewcommand{\th}{\theta}
\newcommand{\om}{\omega}

\DeclareMathOperator{\sign}{sign}

\newcommand{\Cdeux}{\text{C}^{2}}

\newcommand{\lun}{\ell^1}

\newcommand{\Ldeux}{\mathbb{L}^2}

\newcommand{\lzero}{\ell^0}

\renewcommand{\leq}{\leqslant}
\renewcommand{\geq}{\geqslant}

\newcommand{\ThreshH}{H}

\renewcommand{\epsilon}{\varepsilon}

\newcommand{\dotp}[2]{\langle #1,\,#2\rangle}

\newcommand{\norm}[1]{\|#1\|}

\newcommand{\normz}[1]{\norm{#1}_{0}}
\newcommand{\normu}[1]{\norm{#1}_{1}}

\newcommand{\eq}[1]{\begin{equation*}#1\end{equation*}}
\newcommand{\eql}[1]{\begin{equation}#1\end{equation}}

\DeclareMathOperator{\argmin}{argmin}

\newcommand{\uargmin}[1]{\underset{#1}{\argmin}\;}

\newcommand{\enscond}[2]{ \left\{ #1 \,:\; #2 \right\} }

\newcommand{\qandq}{ \quad \text{and} \quad }

\newcommand{\qwhereq}{ \quad \text{where} \quad }

\newcommand{\qwithq}{ \quad \text{with} \quad }

\newlength{\restsubwidth}
\newlength{\restsubheight}
\newlength{\restsubmoreheight}
\newcommand{\rest}[2]{%
        \settowidth{\restsubwidth}{\ensuremath{#2}}
        \settoheight{\restsubheight}{\ensuremath{{}_{#2}}}
        \ensuremath{{#1\hskip 0.5pt}_{\vrule\kern2pt\parbox[b][%
        4pt][b]{\the\restsubwidth}{%
                        \ensuremath{{}_{#2}}}}}
        }

\newcommand{\lagrangian}{\Ll}

\newcommand{\basis}[1]{\Bb^{#1}}

\newcommand{\sqij}{S_{j,i}}

\newcommand{\GSteer}{{\ensuremath{\mathcal G}}}

\newcommand{\NMs}[1][M]{\ensuremath{{\mathbb N}^{\star}_{#1}}}

\newcommand{\xx}{{\ensuremath{\boldsymbol x}}}
\newcommand{\yy}{{\ensuremath{\boldsymbol y}}}
\newcommand{\icomplex}{{\rm i}}

\newcommand{\nDim}[1]{$#1$-D}

\newcommand{\Nbb}{\mathbb{N}}
\newcommand{\Zbb}{\mathbb{Z}}

\newcommand{\Rbb}{\mathbb{R}}
\newcommand{\Cbb}{\mathbb{C}}
\newcommand{\ud}{{\rm d}}
\newcommand{\scp}[2]{\langle #1, #2 \rangle}
\newcommand{\inv}[1]{\frac{1}{#1}}
\newcommand{\tinv}[1]{{\textstyle\frac{1}{#1}}}

\newcommand{\bs}{\boldsymbol}

\newcommand{\etal}{\emph{et al.}\xspace}

\newcommand{\ie}{\emph{i.e.}, }
\newcommand{\eg}{\emph{e.g.}, }

\newcommand{\imageHRP}{of the Haar-Riesz Memorial plaque\xspace}

\title{A Panorama on Multiscale Geometric Representations,\\Intertwining Spatial, Directional and Frequency Selectivity}

\author{Laurent Jacques$^1$, Laurent Duval$^2$, Caroline Chaux$^3$, Gabriel Peyr{\'e}$^4$\\[2mm]
\small $^1$\,ICTEAM Insitute, ELEN Department, Universit\'e catholique Louvain, Belgium\\
\small $^2$\,IFP Energies nouvelles, 1 et 4 avenue de Bois-Pr\'eau F-92852 Rueil-Malmaison, France\\
\small $^3$\,Universit\'e Paris-Est, Laboratoire d'Informatique
Gaspard Monge and UMR--CNRS 8049,\\ 
\small 77454 Marne-la-Vall\'ee, France\\
\small $^4$\,CNRS, CEREMADE, Universit\'e Paris-Dauphine, France}

\date{\today}

\begin{document}

\maketitle

\begin{abstract}
  The richness of natural images makes the quest for optimal
  representations in image processing and computer vision
  challenging. The latter observation has not prevented the design of
  image representations, which trade off between efficiency and
  complexity, while achieving accurate rendering of smooth regions as
  well as reproducing faithful contours and textures. The most recent
  ones, proposed in the past decade, share an hybrid heritage
  highlighting the multiscale and oriented nature of edges and
  patterns in images. This paper presents a panorama of the
  aforementioned literature on decompositions in multiscale,
  multi-orientation bases or dictionaries. They typically exhibit
  redundancy to improve sparsity in the transformed domain and
  sometimes its invariance with respect to simple geometric
  deformations (translation, rotation). Oriented multiscale
  dictionaries extend traditional wavelet processing and may offer
  rotation invariance. Highly redundant dictionaries require specific
  algorithms to simplify the search for an efficient (sparse)
  representation. We also discuss the extension of multiscale
  geometric decompositions to non-Euclidean domains such as the sphere
  or arbitrary meshed surfaces. The etymology of panorama suggests an
  overview, based on a choice of partially overlapping
  ``pictures''. We hope that this paper will contribute to the
  appreciation and apprehension of a stream of current research
  directions in image understanding.
\end{abstract}

\begin{quote}
\end{quote}

\hrule
\tableofcontents
\ \\
\hrule

\section{Introduction: Vision Aspects, Scope and Notations}
\label{sec:intro}
\subsection{Background on Vision Aspects of Scale}

Many natural-world object features are substantive only over a certain
spatial extent.  In other words, the scale of observation is crucial
in object recognition and understanding. For instance, a chair would
be easily recognizable in the scale of a few meters.
But neither at a centimeter scale which captures the chair's texture  and not its object appearance, or   at a hectometer scale, where the chair's appearance is  hardly distinguished from other surrounding objects.

Accordingly, early neurophysiological
studies in biologic perception reveal that those objects are generally
apprehended differently according to the scale of observation by the
sensory receptors and the cortex of mammalians
\cite{Daugman_J_1985_j-opt-soc-am-a_unc_rrssfootvcf,DeValois_R_1982_j-vis-res_spa_fscmvc}.
Efficient information extraction is thus required for artificial
sensing systems to mimic standard biologic tasks such as object
recognition.

Pixel-based  representations as  linear combinations of
``delta'' functions suffice for simple data manipulation but are very
limited for higher level tasks.  
Only assuming some sufficient
resolution in the data, the lack of prior knowledge in the extent of
objects to be analyzed calls for tools able to unveil the appropriate
scales and to allow a hierarchical representation of 
the underlying  features
\cite{Hildreth_E_1980_tr_imp_ted,Marr_D_1979_j-roy-stat-soc-b_com_thsv,Marr_D_1980_j-proc-roy-soc-b_the_ed}.
Disregarding the peculiar fractal formalism
\cite{Massopust_P_1994_book_fra_ffsw,Wornell_G_1995_book_sig_pfwba}
where similar phenomena appear at different scales (what is called
\emph{self-similarity}),  special attention has been paid to data
transformations able to capture object features over a range of scales
in a more compact form.  Sparsity, amounting to a reduced number of
parameters in a suitable domain, is thus used as a heuristic guide to
image understanding.  Bearing analogies with findings in vision
processes \cite{Olshausen_B_1997_j-vis-res_spa_cobssev1}, several
sparse decompositions have proven efficient in image compression, with
the discrete wavelet transform (DWT) as their most 
well-known 
avatar,
often intermingled with information theory and technical 
wizardry, from bit plane arithmetic coding
\cite{Shapiro_J_1993_j-ieee-tsp_emb_iczwc} to trellis coded
quantization. A compact history and a paper collection are given in
\cite{Davis_G_1998_incoll_wav_bico,Topiwala_P_1998_book_wav_ivc},
respectively.

\begin{figure}[htb]
  \centering
  \subfigure[\label{fig:cartoon}]{
    {\includegraphics[height=4cm,keepaspectratio]{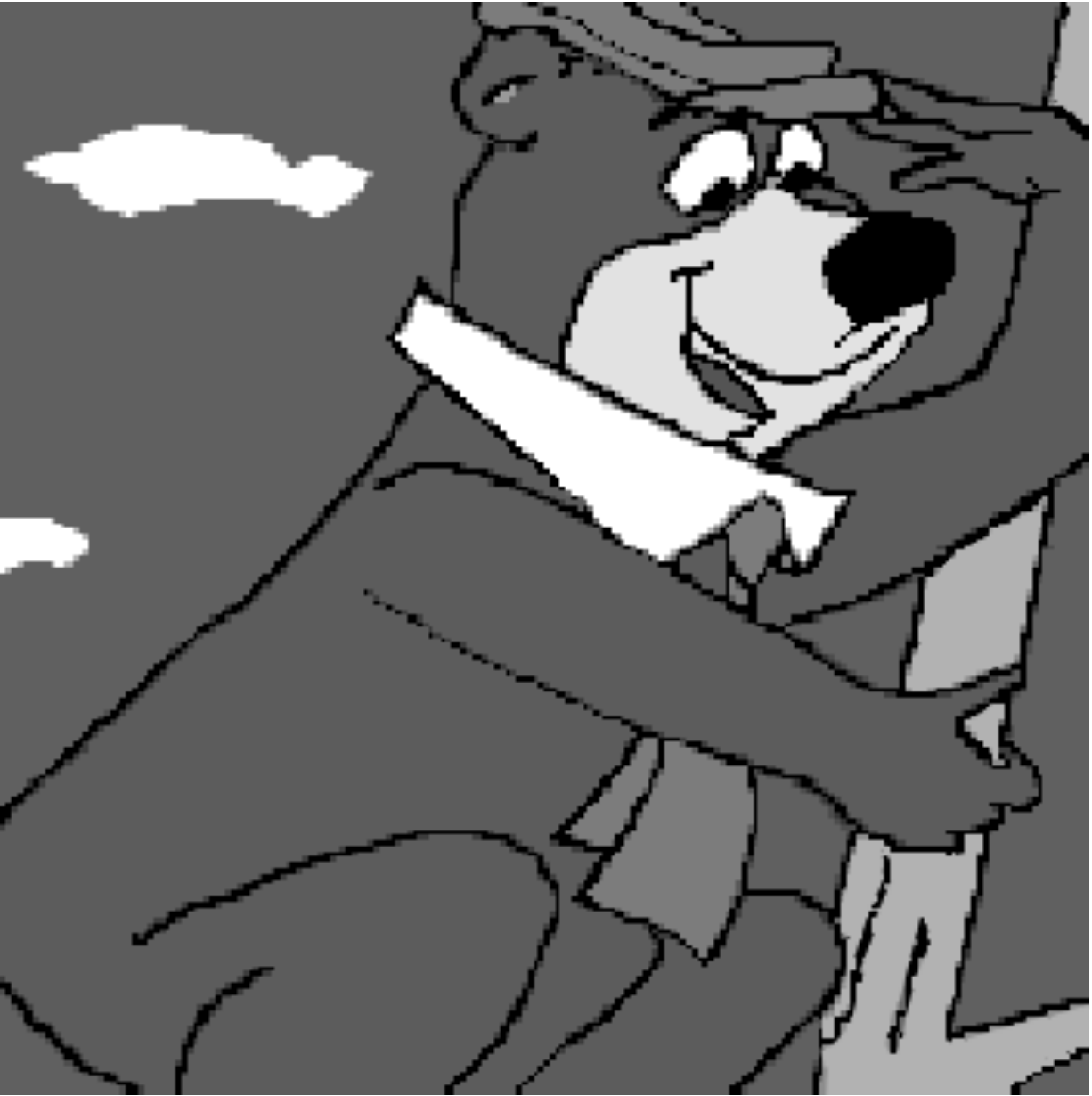}}
  }
  \subfigure[\label{fig:fingerprint}]{
    \includegraphics[height=4cm,keepaspectratio]{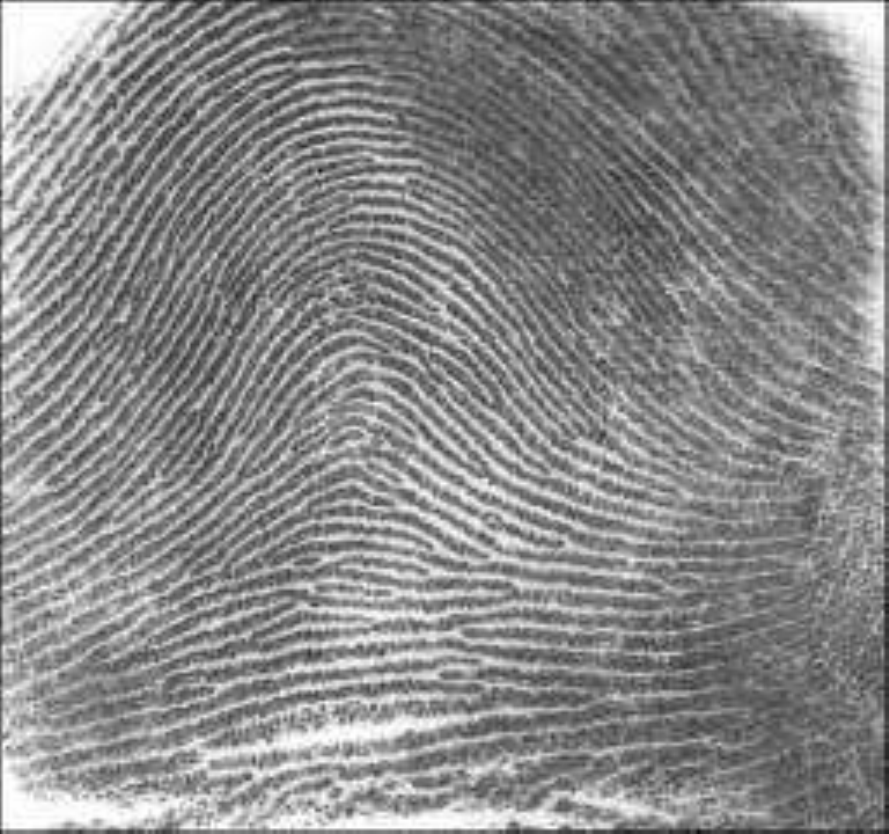}
  }
  \caption{Two faces
  of the cartoon-texture model:  (a) Yogi bear (b) Fingerprint.}
\end{figure}

Yet, beyond image compression transforms, other decomposition
techniques are needed, with more resolving power in complex scene
detection, denoising, segmentation or, in a broad sense,
scene understanding. As a matter of fact, standard separable wavelet
transforms appropriately detect point-like (\nDim{0}) singularities and
address mild noise levels. Still they generally lack performance in dealing
with higher dimensional features combining both regularity and
singularity such as edges, contours or regular textures, that may 
also be anisotropic.
Amongst their limitations are shift
sensitivity, limited orientation selectivity, rigid and uneven atom
shapes (\eg fractal-looking asymmetric Daubechies wavelets), crude
frequency direction selection.  Major challenges reside in a proper
definition of the underlying regularity (with respect to each feature)
and 
corresponding
singularities. These challenges are amplified
  by additional degradations from which acquired  data may suffer such as blur,
jitter and  noise.  Descriptive mathematical models of images combining
cartoon and textures become increasingly popular
\cite{Meyer_Y_2001_incoll_osc_pipnee,Aujol_J_2005_j-math-imaging-vis_ima_dbvcoc} and progressively yield
tractable algorithms.
We note that
there exists a continuum of real-world images between cartoon and
textures, ranging from  cartoon-ish Yogi bear in Fig.~\ref{fig:cartoon} to
``textural'' fingerprints in Fig.~\ref{fig:fingerprint}. 
In between these two extreme image types, there exists
many possible variations in image object complexity.

Moreover, both contours and (even regular) textures are known to
be ill-defined. 
They are indeed viewer- and scale-dependent concepts in discrete
images or volumes.  Consider an image resulting from a
combination of  piecewise smooth components,   contours,   geometrical
textures and   noise.  Their
discrimination  
 is required for high level image processing
tasks.  Each of these four components could be detected,
described and modeled by different formalisms: smooth curves or
polynomials, oriented regularized derivatives, discrete geometry, parametric curve
detectors (such as the Hough transform), mathematical morphology, local frequency estimators,
optical flow approaches, smoothed random models, etc. They have progressively influenced the 
hybridization  of standard multiscale transforms towards 
 more geometric and sparser representations of 
such components, with improved localization, orientation
sensitivity, 
frequency selectivity
or noise robustness.

\subsection{Scope of the Paper}
\label{sec:scope-paper}

Geometry driven  ``$\star$-let'' transforms \cite{Duval_L_2005_url_wits}
have been popular in the past decade, with a seminal ancestor in
\cite{Daugman_J_1980_j-vis-res_two_dsacrfp}. 
Early \cite{Candes_E_1999_curves-surfaces_cur_senroe}, a debate 
 opened on the relative strength of Eulerian (non-adaptive)
versus Lagrangian (adaptive) representation, now
pursued with the growing interest in dictionary learning
\cite{Rubinstein_R_2010_j-proc-ieee_dic_srm}.  

\begin{figure}[htbp]
  \centering
  \includegraphics[height=8cm]{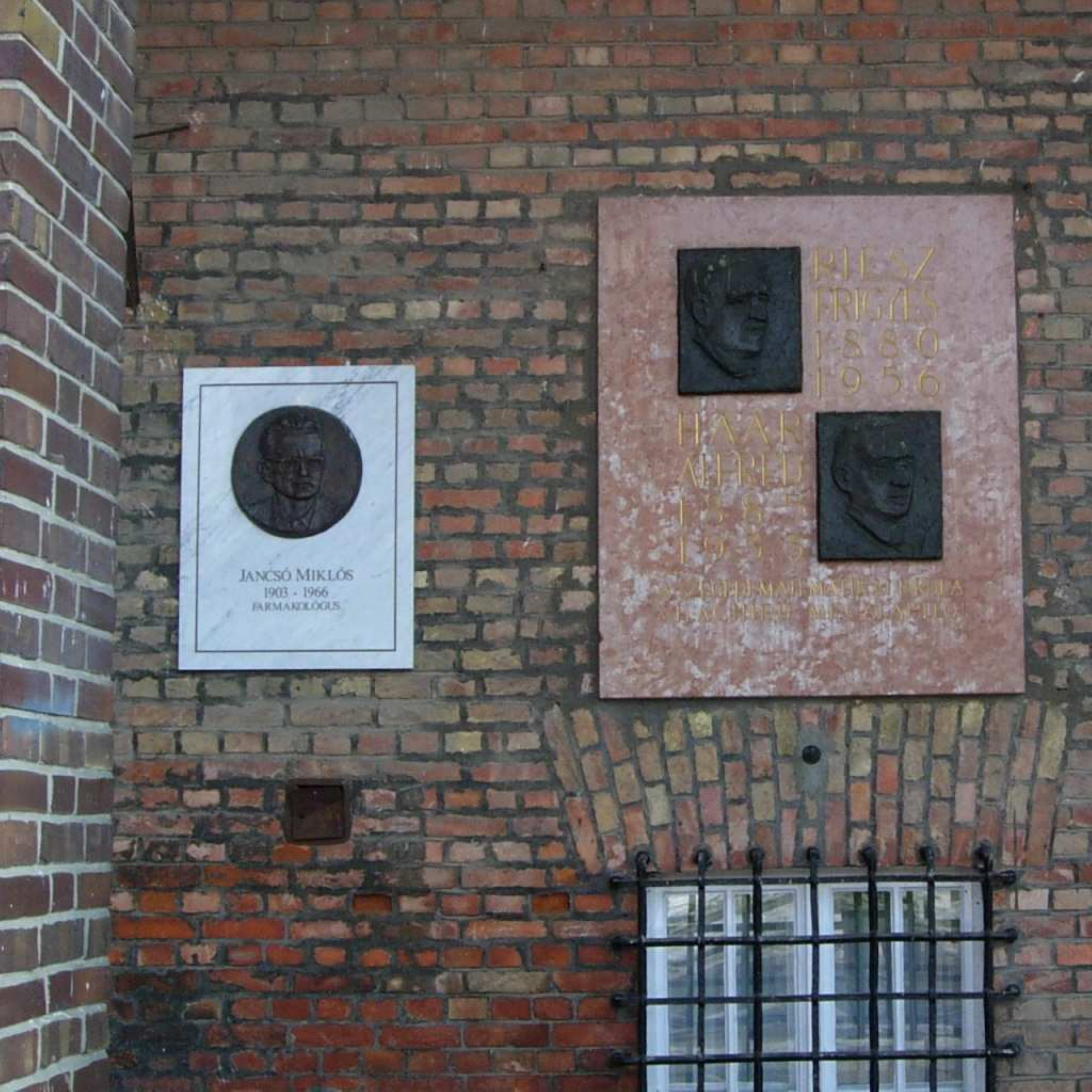}
  \caption{Szeged University Memorial plaque in honor of A. Haar and F. Riesz: \emph{A szegedi matematikai iskola vil{\'a}gh{\'i}r{\H u} megalap{\'i}t{\'o}i} (The world-wide famous founders of the mathematical school in Szeged).}
  \label{fig:fig_SzegedOriginalColor}
\end{figure}

As of today, the authors believe that the 
discussion is not fully settled in 
 the various
different uses of sparsity in images.
Neither has the
trade-off between redundancy and sparsity.  
A number of early
 papers on geometric multiscale methods appear in
\cite{Welland_G_2003_book_bey_w}. Comparisons  are drawn in \cite{Romberg_J_2003_phd_mul_gip,Lisowska_A_2005_phd_geo_wgdicp}, while 
\cite{Fuhr_H_2006_incoll_bey_wnirp,Ma_J_2010_j-ieee-spm_cur_t,Fadili_J_2009_incoll_cur_r,Ma_J_2010_j-ieee-spm_cur_t,Starck_J_2010_book_spa_ispwcmd} focus on ridgelets, curvelets and
wedgelets, as representative of fixed and adaptive decompositions.
The present paper  aims at
providing a broader panorama 
of the recent developments in multiscale decompositions targeted to efficient representation of
 geometric features in images: smooth content
(multiscale or hierarchical), edges and contours (locally spatial) and
textures (locally spectral). We emphasize the main characteristics and differences pertaining to spatial, directional and frequency selectivity
of the selected methods. 
The paper  therefore cites a dense set of references, 
ranging from continuous to discrete
representations, from (nearly) orthogonal to (fully) redundant. 
As a guiding thread to this panorama, we  illustrate some of the reviewed geometric multiscale
decompositions on a memorial plaque\footnote{Courtesy of Professor K\'aroly Szatm\'ary,
  \url{http://astro.u-szeged.hu/szatmary.html}, who performed
  scalograms analysis of variable stars as early as in 1992
  \cite{Szatmary_K_1992_j-mon-not-roy-astron-soc_per_lcsvsyl}.} in Szeged University, Hungary,
depicted in Fig.~\ref{fig:fig_SzegedOriginalColor}. It features simple objects (embedded rectangles and a disk), a few
differently oriented features and regular textures at different
scales. Since some of the illustrations have been
slightly enhanced  to improve the clarity of details, they are
available in original resolution online \cite{Jacques_L_2011_url_panorama_addendum}.
This picture finally  honors  Alfr\'ed Haar's
originative paper \cite{Haar_A_1910_ma_zur_tofs} \emph{Zur {Theory der
    orthogalen Funktionen Systeme} (On the Theory of Orthogonal
  Function Systems)} and his  eponymous wavelet. He also
founded  \emph{Acta Scientiarum Mathematicarum}  together
with Frigyes Riesz, whose works percolated wavelet theory
\cite{Christensen_O_2001_j-bull-amer-math-soc_fra_rbdgwe}.

The paper is organized as follows: the remaining of
Section~\ref{sec:intro} is devoted to context and notations for image
representations. Then, as a preliminary to geometric tools, a quick
survey of early multiscale decompositions is presented in
Section~\ref{sec:introtools}. More recent transforms, termed directional
or geometrical, circumventing aforementioned drawbacks, are discussed in
Section~\ref{sec:oriented}. Owing to the additional degrees of freedom
provided by these representations, a discussion is carried out in
Section~\ref{sec:red_adapt} 
on 
redundancy and
adaptivity. The extension of frequency, scale and directionality to
non-Euclidean spaces or grids such as the sphere, are presented in
Section~\ref{sec:noneuclidian}. Finally, 
concluding remarks are
given in Section~\ref{sec:conclusion}.

\subsection{Mathematical Framework}

\subsubsection{Notations and Conventions}
\label{sec:notat-conv}

This paper describes numerous mathematical methods designed for
different spaces and geometries. We have tried therefore to adopt
coherent representations for the many mathematical notions that
coexist in this text. For instance, functions and vectors in high
dimensional spaces are generally referring to some signal of interest
(\eg \nDim{1} signals or images). They must therefore share the same
notations and we thus decided to write them as simple lowercase Roman or
Greek letters. However, coordinate systems, vectors in 2 or 3
dimensions and multi-indices are denoted in bold symbols.

The (Hilbert) space $\Ldeux(\mathcal{X})$ is the space of square
integrable functions on the space $\mathcal{X}$, \ie given the
(Lebesgue) integration measure $\ud\rho$ on that space,
$\Ldeux(\mathcal{X})=\Ldeux(\mathcal{X},\ud\rho)=\{f:\mathcal{X}\to
\Cbb: \norm{f}^2:=\int_{\mathcal{X}}\ |f(\bs u)|^2\,\ud\rho(\bs u) <
\infty\}$. In $\Ldeux(\mathcal{X})$ the inner product between two
functions $g,h\in \Ldeux(\mathcal{X})$ is denoted by
$\scp{g}{h}=\int_{\mathcal{X}} g^*(\bs u)\, h(\bs u)\ \ud\rho(\bs u)$
with $^*$ the complex conjugation. By extension, for $p \geq 1$, we also use
 the (Banach) spaces
$\LL^p(\mathcal{X})=\LL^p(\mathcal{X},\ud\rho)=\{f:\mathcal{X}\to \Cbb:
\norm{f}_p^p:=\int_{\mathcal{X}}\ |f(\bs u)|^p\,\ud\rho(\bs u) <
\infty\}$, with $\norm{\!\cdot\!}_2=\norm{\!\cdot\!}$.

We also use some discrete spaces as the common $\ell^p_N=(\Cbb^N,
\norm{\cdot}_p)$ with $\norm{v}^p_p:=\sum_i |v_i|^p$ for $p\geq 1$ and
$v\in\Cbb^N$, with again the shorthand
$\norm{\!\cdot\!}=\norm{\!\cdot\!}_2$. In $\ell^2_N$, the inner
product between $u, v\in \ell^2_N$ is written $\scp{u}{v}= u \cdot v =
\sum u_i^* v_i$. Whether the overused notations $\scp{\cdot}{\cdot}$
or $\norm{\!\cdot\!}_p$ are applied to continuous or discrete
mathematical objects will remain clear from the context. The spaces
$\ell^p$ are the generalization of the previous finite spaces to
infinite sequences, \ie $\ell^p=\{ v = (v_i)_{i\in\Nbb}:
\|v\|^p_p=\sum_{i\geq 0} |v_i| <\infty \}$.

For functions $f\in \Ldeux(\mathcal{X})$ or discrete sequences
$v\in\ell^2_N$, $\hat{f}$ and $\hat{v}$ denote the Fourier transform
of $f$ or $v$ respectively. For instance, for $\mathcal{X}=\Rbb$ and
$f\in \Ldeux(\Rbb)$, $\hat{f}(\omega)=\tinv{\sqrt{2\pi}}\int_{\Rbb}
f(t)\,e^{-\icomplex \omega t}\ \ud t$ and $f(t) =
\tinv{\sqrt{2\pi}}\int_{\Rbb} \hat{f}(\omega)\,e^{\icomplex \omega t}\
\ud \omega$ are the Forward and Inverse Fourier transform respectively.
For $f\in \Ldeux(\Rbb^2)$ and $\bs x,\bs \omega\in\Rbb^2$, the same
transforms are $\hat{f}(\bs \omega)=\tinv{{2\pi}}\int_{\Rbb^2} f(\bs
x)\,e^{-\icomplex \bs \omega\cdot\bs x}\ \ud^2\bs x$ and $f(\bs x) =
\tinv{{2\pi}}\int_{\Rbb^2} \hat{f}(\bs \omega)\,e^{\icomplex \bs
  \omega \cdot \bs x}\ \ud^2 \bs\omega$. For $v\in\ell^2_N$, the same
transforms are $\hat{v}_k = \inv{\sqrt{N}}\,\sum_j v_j \exp(-
2\pi\icomplex\,jk/N)$ and $v_j = \inv{\sqrt{N}}\,\sum_k \hat v_k \exp(
2\pi\icomplex\, jk/N)$. In matrix algebra notations, this can be
rewritten as $v = \mathcal{F} \hat v$ and $\hat v = \mathcal{F}^* v$,
where the Fourier matrix $\mathcal{F}\in \Cbb^{N \times N}$ is given
by $\mathcal{F}_{jk} = \inv{\sqrt N}\exp( 2\pi\icomplex\,jk/N)$, and
$\mathcal{F}^*$ is its complex adjoint. The convolution by
time-invariant filter $h$ operates as $(f \star h) (t) =
\int_{-\infty}^{\infty} f(u) h^*(t-u) \ud u $ and $(v \star h)_n =
\sum_{n'} h^*_{n'} v_{n-n'}$ in continuous and discrete sample
domain\footnote{With periodization for finite length vectors.}
respectively.  The ubiquitous Gaussian kernel with scale parameter
$\sigma>0$ is denoted by $G_\sigma(\bs x) =
\exp(-\tinv{2\sigma^2}\|\bs x\|^2)$, with $G(\bs x)=G_1(\bs x)$.

\subsubsection{Image Representations in Bases and Frames}
\label{sec:introtools-representation}
\medskip
\paragraph{Stability and Frames}

This paper describes processing methods that make use of a
decomposition of the image $f \in \Ldeux([0,1]^2)$ into a family of
atoms $\Basis = \{\atom_{\m}\}_{\m}$. Each atom $\atom_{\m} \in
\Ldeux([0,1]^2)$ is parameterized by a multi-index $m$ (that might
take into account its frequency, position, scale and
orientation). Numerical processing is performed on discretized images
which are vectors $f \in \RR^N$, where $N$ stands for the number of
pixels. The atoms of $\Basis$ are also discretized and the continuous
inner products are replaced by the standard discrete inner product in
$\RR^N$.

To guarantee a stable reconstruction from the coefficients
$\{ \dotp{\atom_{\m}}{f} \}_{\m}$, the family $\Basis$ is assumed to
be a frame \cite{Duffin_R_1952_tams_cla_nhfs,Casazza_P_2000_tjm_art_ft,Christensen_O_2001_j-bull-amer-math-soc_fra_rbdgwe,Kovacevic_J_2007_ieee-spm_lif_bbaf1,Kovacevic_J_2007_ieee-spm_lif_bbaf2} of $\Ldeux([0,1]^2)$ or $\RR^N$, which means that there
exist two constants $0<\mu_1\leq\mu_2<\infty$  such that for all $f$
\eql{\label{eq:trame} \mu_1 \norm{f}^2 \leq \sum_{\m}
  |\dotp{\atom_{\m}}{f}|^2 \leq \mu_2 \norm{f}^2.  } 

Atoms are allowed to be linearly dependent, thus corresponding to a
redundant representation. Redundancy enables atoms to meet certain
additional constraints, for instance smoothness, symmetry and
invariance to translation or rotation.

\paragraph{Thresholding for Approximation and Processing}

Using a dual frame $\{\tilde\atom_{\m}\}_{\m}$
\cite{Christensen_O_2001_j-bull-amer-math-soc_fra_rbdgwe}, an image is
recovered from the set of coefficients as $f = \sum_{\m}
\dotp{\atom_{\m}}{f} \tilde\atom_{\m}$. The computation of the set of
coefficients $\{\dotp{\atom_{\m}}{f} \}_{\m}$ for a discrete image $f
\in \RR^N$ is usually performed using a fast algorithm, that also
enables a fast reconstruction of an image from coefficients.

The basic processing operation, used in denoising and compression applications, is the thresholding
\eql{\label{eq-thresholding}
	f_M = \ThreshH_T(f,\Basis) = \sum_{ \m\ :\ |\dotp{\atom_{\m}}{f}|>T } \dotp{\atom_{\m}}{f}\ \tilde\atom_{\m}
}
where $M = \#\enscond{\m}{|\dotp{\atom_{\m}}{f}|>T}$ counts the number
of non-zero coefficients in (\ref{eq-thresholding}).  

When $\mu_1=\mu_2$, the frame is said to be \emph{tight} (Parseval tight
frame). If furthermore $\mu_1=\mu_2=1$, then one can choose
$\tilde\atom_{\m}=\atom_{\m}$, and $\Basis = \{\atom_{\m}\}_{\m}$ is
then an orthonormal basis if $\|\psi_{\m}\|=1$ for all $\m$.  In this
last case, $\Basis$ performs the least energy reconstruction of $f_M$
in \eqref{eq-thresholding}, or equivalently, $f_M$ is the best
$M$-terms approximation of $f$.  
The decay of the approximation error $\norm{f-f_M}$ is related to both the average risk of a
denoiser, and the distortion rate decay of a coder, see for instance
\cite{Mallat_S_2009_book_wav_tspsw}. This motivates the search for
bases or frames $\Basis$ which can efficiently approximate large classes of
(natural) images.
When the frame is redundant, more complicated decomposition methods
improve the sparsity of the representation
(see Sec.~\ref{sec:red_adapt-pursuit}).

\section{Early Scale-Related Representations}
\label{sec:introtools}
\subsection{Frequency, Heat Kernel and Scale-Space Formalism}
\label{sec:introtools-contss}

At the heart of modern signal processing techniques is the concept of
\emph{signal representation}, \ie the selection of an efficient
``point of view'' in the study of signal properties that is not
restricted to straightforward spatial descriptions.

\begin{figure}[htbp]
  \centering
  \includegraphics[height=6cm]{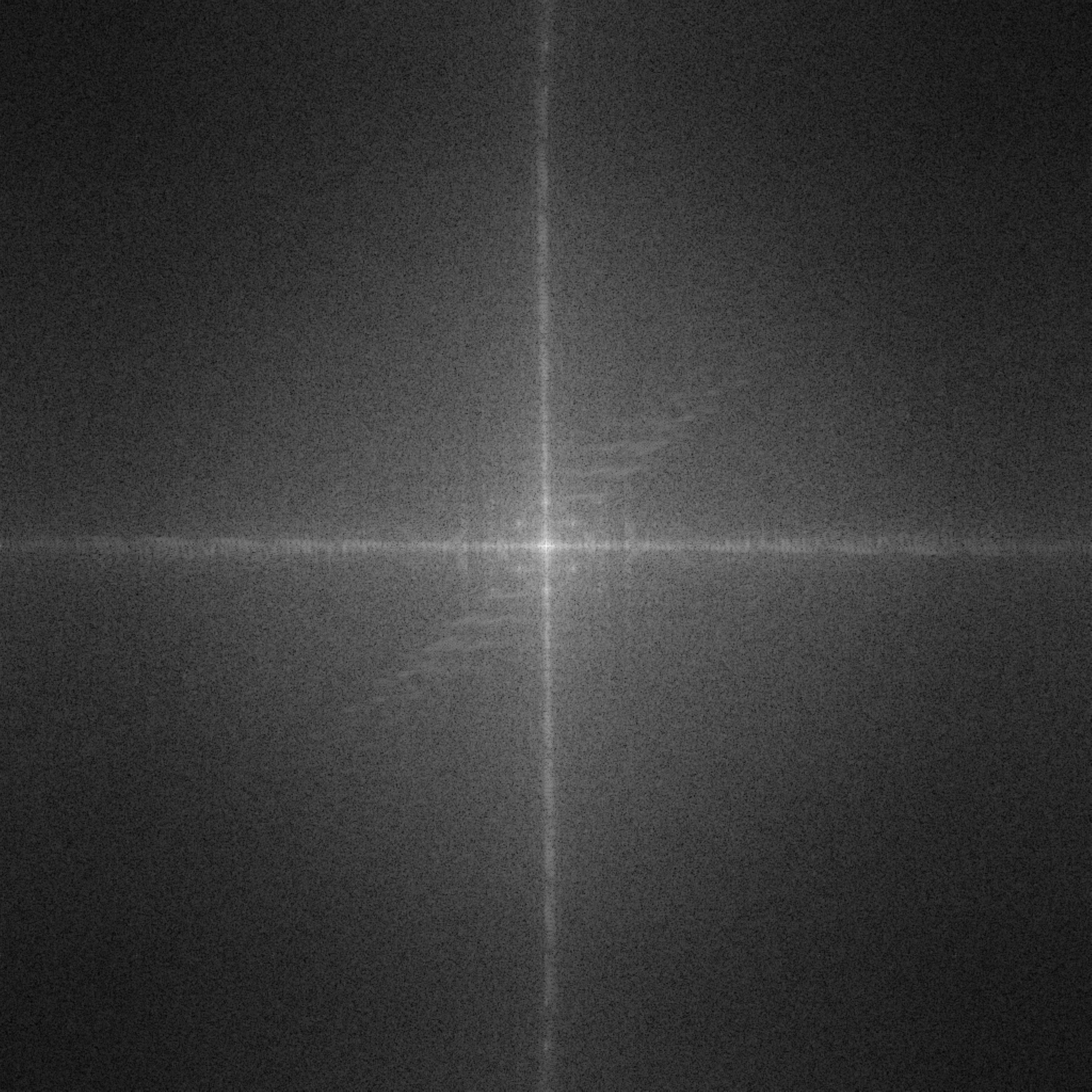}
  \caption{Magnitude of the \nDim{2} Fourier transform  \imageHRP in Fig.~\ref{fig:fig_SzegedOriginalColor}.}
  \label{fig:fig_SzegedFourierMagnitude}
\end{figure}

The most obvious alternative signal representation is its frequency
reading, \ie the one provided by the Fourier transform of the signal
explained in Sec.~\ref{sec:notat-conv}
\cite{Bracewell_R_1986_book_fou_ta,Bremaud_P_2002_book_mat_pspfwa}. However,
this representation is not sufficiently ``local''. It is indeed rather
difficult to detect what spatial part of an image contributes to high
peaks in the Fourier spectrum.
Fig.~\ref{fig:fig_SzegedFourierMagnitude} represents the amplitude
spectrum\footnote{The original image has been multiplied by a \nDim{2}
raised-cosine type apodizing window in order to reduce border discontinuity
effects.} of the luminance component from
Fig.~\ref{fig:fig_SzegedOriginalColor}. It exhibits a mixture of
prominent vertical and horizontal directions with tiny fuzzy diagonal
ones.

\begin{figure}[htbp]
  \centering
\subfigure[\label{fig:gauss_sp4}]{
    {\includegraphics[height=5cm,keepaspectratio]{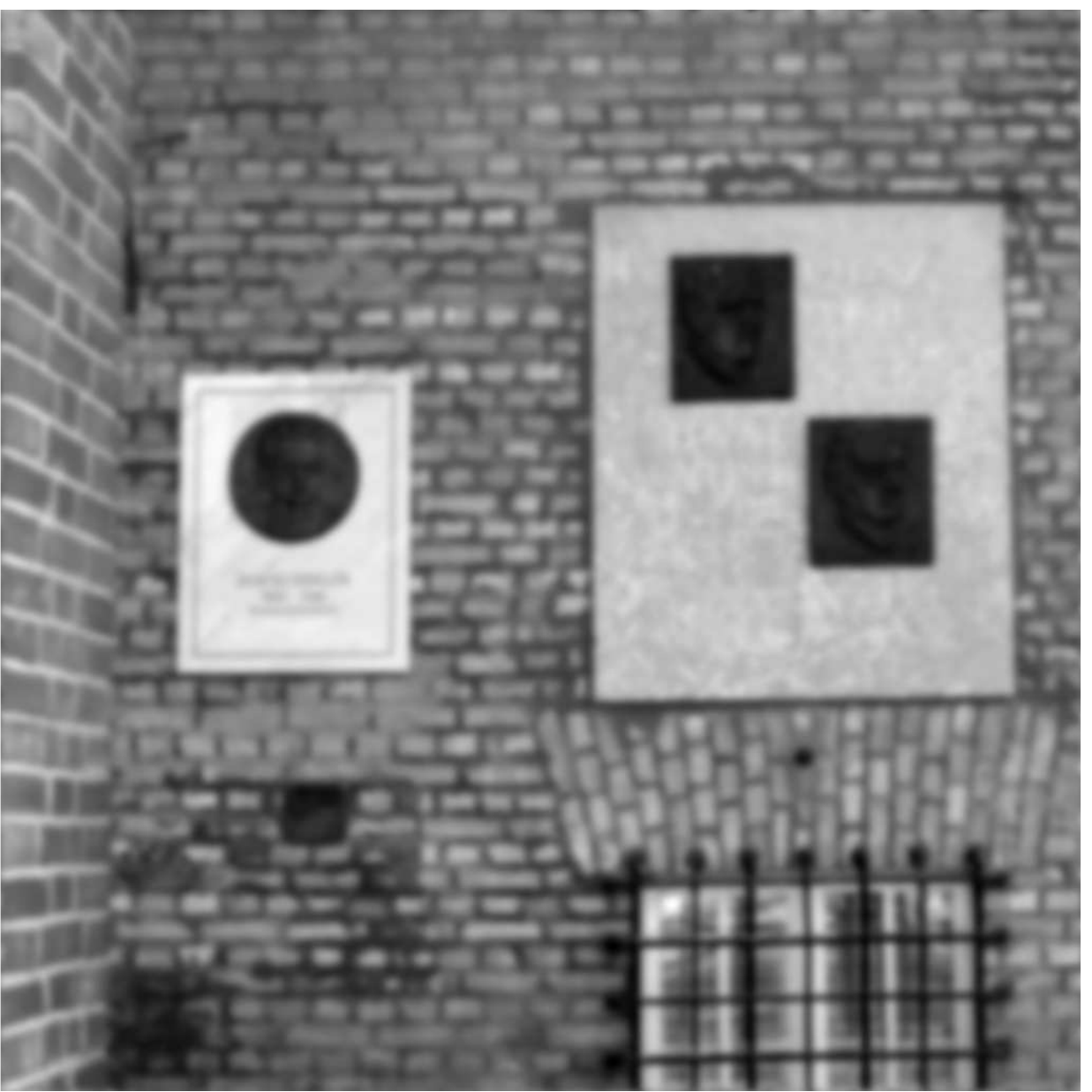}}
  } 
\subfigure[\label{fig:gauss_sp8}]{
    {\includegraphics[height=5cm,keepaspectratio]{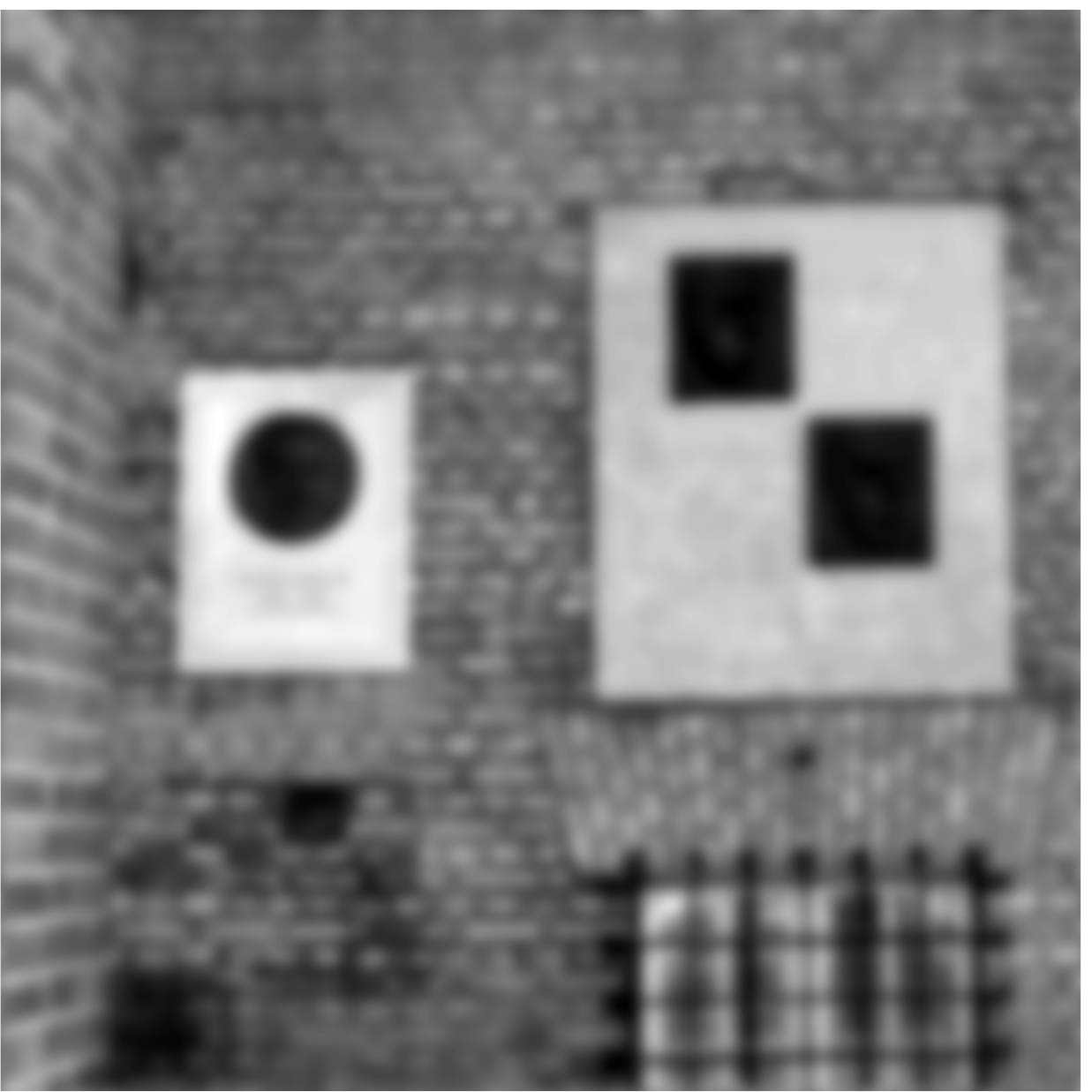}}
  }
\subfigure[\label{fig:gauss_sp16}]{
    {\includegraphics[height=5cm,keepaspectratio]{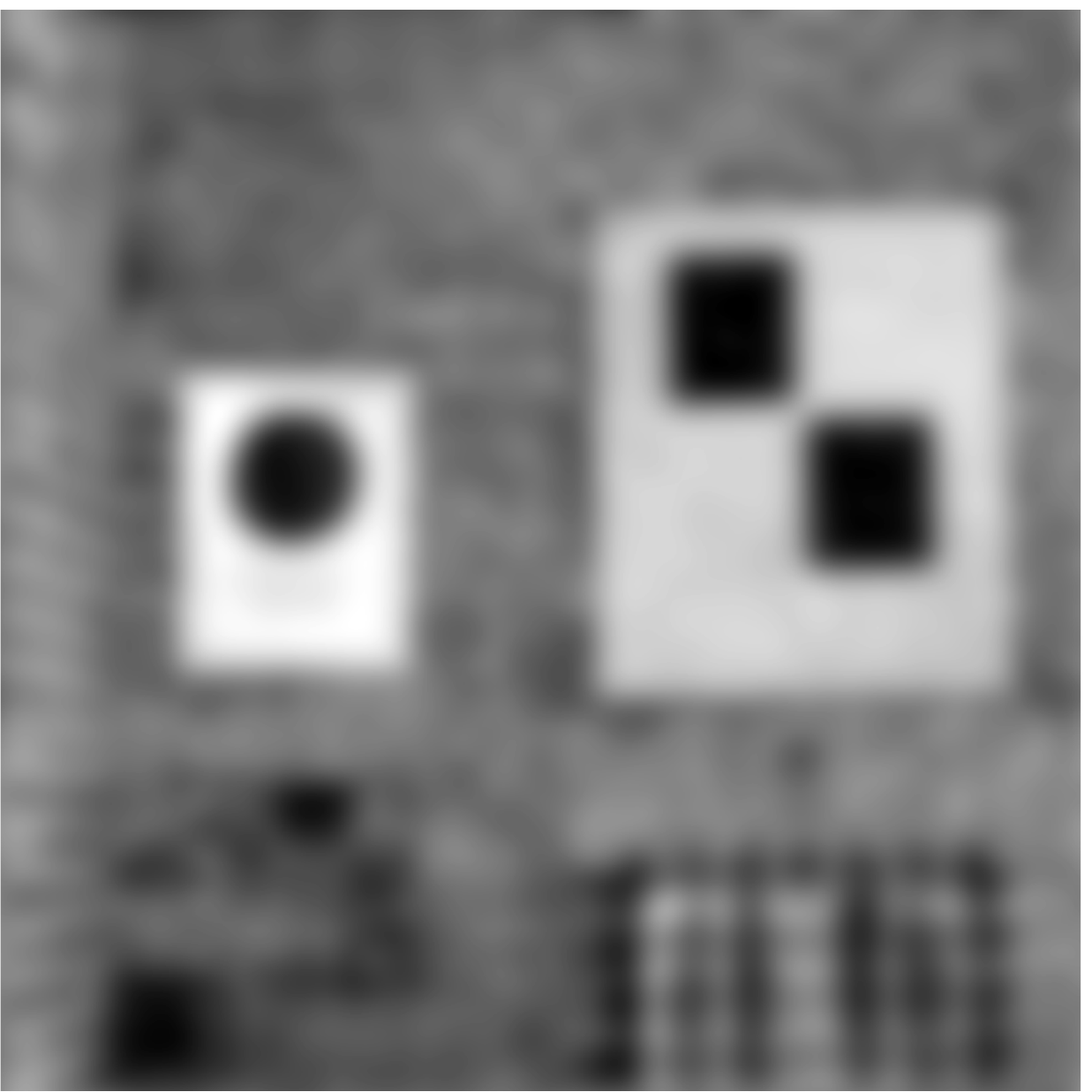}}
  }
  \caption{Gaussian scale-space decomposition \imageHRP at three different scales.}
  \label{fig:fig_SzegedGaussianScaleSpace}
\end{figure}

An approach for obtaining a better localization is to introduce a
notion of ``scale'' in the image observation. This has been performed
very early in image and signal processing by either windowing  or introducing scales in the
Fourier transform \cite{Allen_J_1977_tassp_sho_tsasmdft,Wilson_R_1992_j-ieee-tit_gen_wtfamftaiasa} or observing
a well-known diffusion process like the heat dynamics governed by the
famous Heat equation. The idea relies on considering the image as an
initial configuration of heat that is diffused with a time variable
$\tau>0$ and in interpreting this time parameter as the
``scale''. Indeed, in this dynamic diffusion, small image structures
will be smoothed early at small evolution time while larger ones
persist for a larger duration. Interestingly, this diffusion is
equivalently described by a filtering process: the convolution of the
image by a Gaussian function $G_\sigma$ of width $\sigma=\sqrt{2\tau}$
\cite{Witkin_A_1984_p-icassp_sca_sfnamsd,Babaud_J_1986_tpami_uni_gkssf,Bredies_K_2005_j-acha_mat_cms}. This
image unfolding into a scale-space domain has led to many new image
processing techniques such as edge, ridge and feature detection
\cite{Lindeberg_T_1993_j-math-imaging-vis_dis_dasspbllfe,Florack_L_1998_tr_top_sssi}.
This is illustrated in Fig.~\ref{fig:fig_SzegedGaussianScaleSpace},
where the original image is convolved with three different Gaussian
kernels in dyadic progression. Large objects such as the white
rectangular plaques persist across all scales, while brick and grid
textures vanish in Fig.~\ref{fig:gauss_sp16}.  The overall redundancy
of the Gaussian pyramid is given by the number of smoothing
kernels. Taking advantage of the resolution loss, the redundancy
factor may be reduced by sub-sampling, leading to the ``Gaussian
pyramid'' construction.

\begin{figure}[htbp]
  \centering
  \includegraphics[height=6cm]{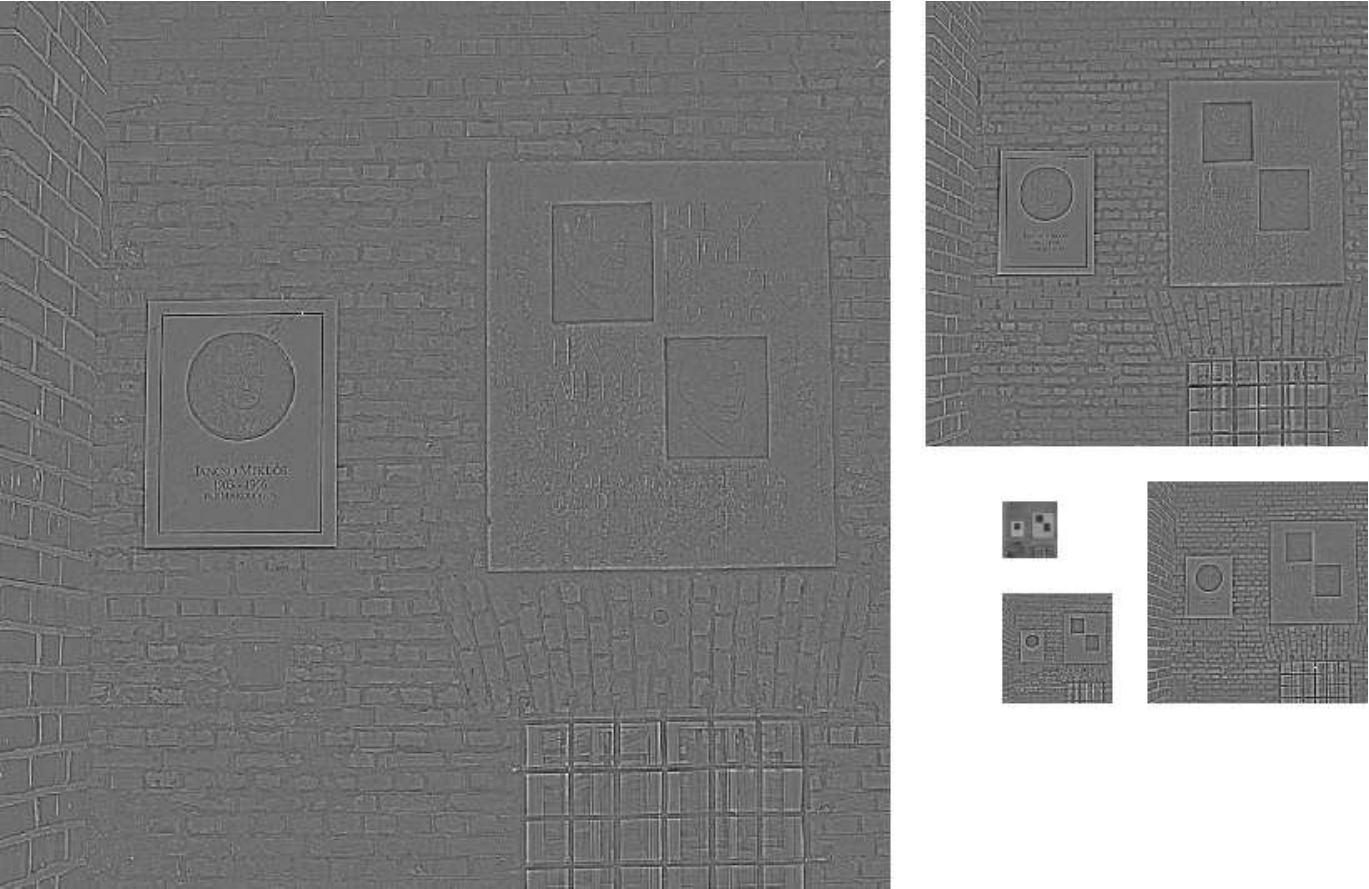}
   \caption{Laplacian pyramid decomposition \imageHRP.}
  \label{fig:fig_SzegedPyramidLaplacian}
\end{figure}

The scale content of the image can  be decomposed further by computing, for
instance, differences between two filterings performed at two different
scales.  This led to the famous Littlewood-Paley decomposition, or to
the (invertible) Laplacian pyramid conveniently combining  multiple
sub-sampled low-pass filterings of images, creating a pyramidal scale 
hierarchy \cite{Burt_P_1983_tcom_lap_pcic}. Interestingly, the
resulting decomposition represented in
Fig.~\ref{fig:fig_SzegedPyramidLaplacian} is a complete image
representation that can advantageously be processed before
reconstructing a new ``restored'' image (\eg in image denoising). Additionally, image singularities are enhanced at
fine scales, with low activity regions
  associated with coefficients being close to zero.
  Fast implementations of deformable (steerable or
scalable) decompositions \cite{Treitel_S_1971_j-ieee-tge_des_mspf} are
available for instance with recursive filters
\cite{Deriche_R_1993_tr_rec_igd} or efficient multirate filter banks
\cite{Manduchi_R_1998_tsp_eff_dfb,Adelson_E_1984_j-rca-eng_pyr_mip,Ogden_J_1985_j-rca-eng_pyr_bcg,Do_M_2003_j-ieee-tsp_fra_p}.

Remarkably, the notion of Scale-Space has been defined and
``axiomatized'' more than 50 years ago by the Japanese mathematicians
Iijima and Otsu, as presented in \cite{Weickert_J_1997_tr_sca_sdj}.  As
we will realize throughout this paper, this scale-space representation (refer to  \cite{Lindeberg_T_2011_j-math-imaging-vis_gen_gssaclssassstss} for a recent  overview and axiomatic generalization)
was the starting point of many new ways to represent images.

\subsection{Isotropic Continuous Wavelet Transform}
\label{sec:introtools-contss-wavelet}

The continuous wavelet transform somehow generalizes the previous
scale-space formalism driven by the Gaussian kernel to any
``function'' with enough regularity. The continuous wavelet
transform was initially developed for the transformation of \nDim{1}
signals \cite{Grossmann_A_1984_misc_dec_fwcsrt} and further extended
in \nDim{2} first with \emph{isotropic} wavelets. The case of
non-isotropic (directional) wavelets was defined later
\cite{Antoine_J_1993_j-sp_ima_atdcwt} (see
Sec.~\ref{sec:direct-wavel-fram}).

In one dimension, a wavelet $\psi$ is an integrable and well-localized
function of $\LL^2(\RR)$, generally described as locally oscillating,
\ie $\int_\Rbb \psi(t) \ud t = 0$. It may be dilated or contracted by
a scale factor $a>0$ and translated to a position $b \in
\RR$: $\psi_{(b,a)}(t)=\tfrac{1}{\sqrt{a}}\,\psi(\tfrac{t-b}{a})$.

\begin{figure}[t!]
  \centering
\subfigure[\label{fig:mexicanhat}]{
  \raisebox{5mm}{\includegraphics[width=5cm]{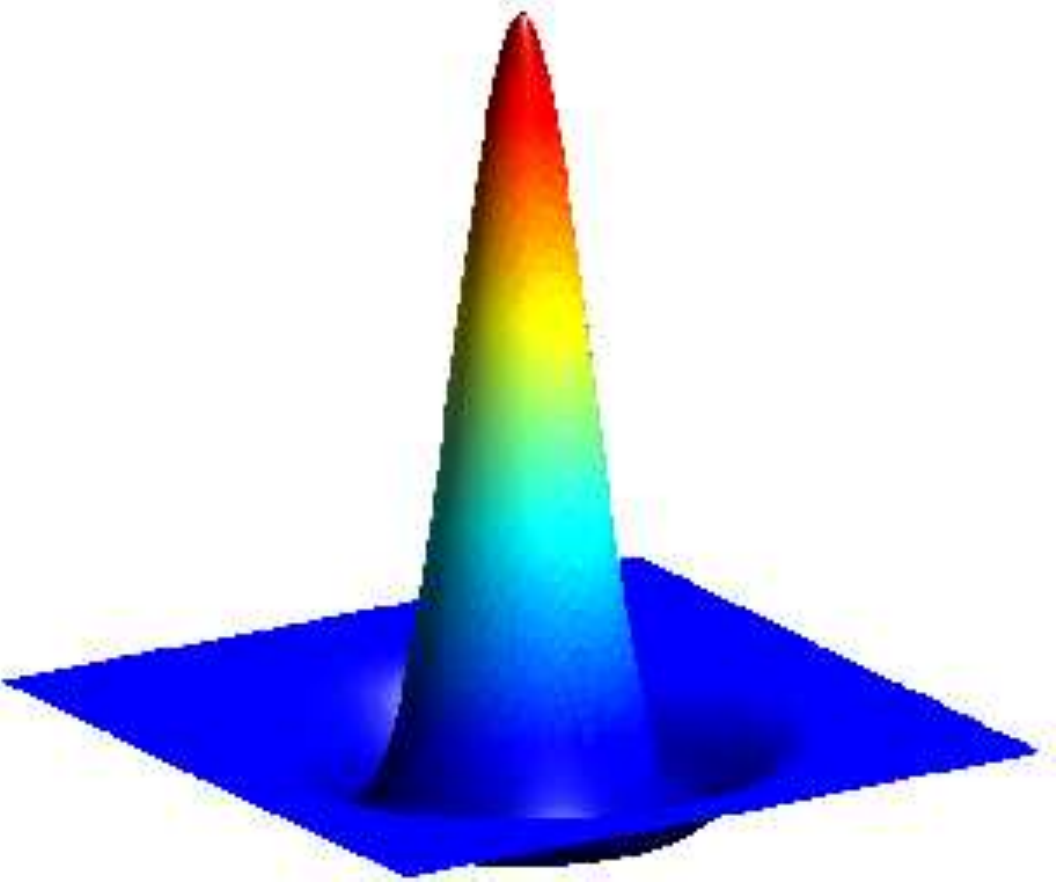}}}\hspace{5mm} 
\subfigure[\label{fig:fig_SzegedMarrWavelet}]{
  \includegraphics[height=5.5cm]{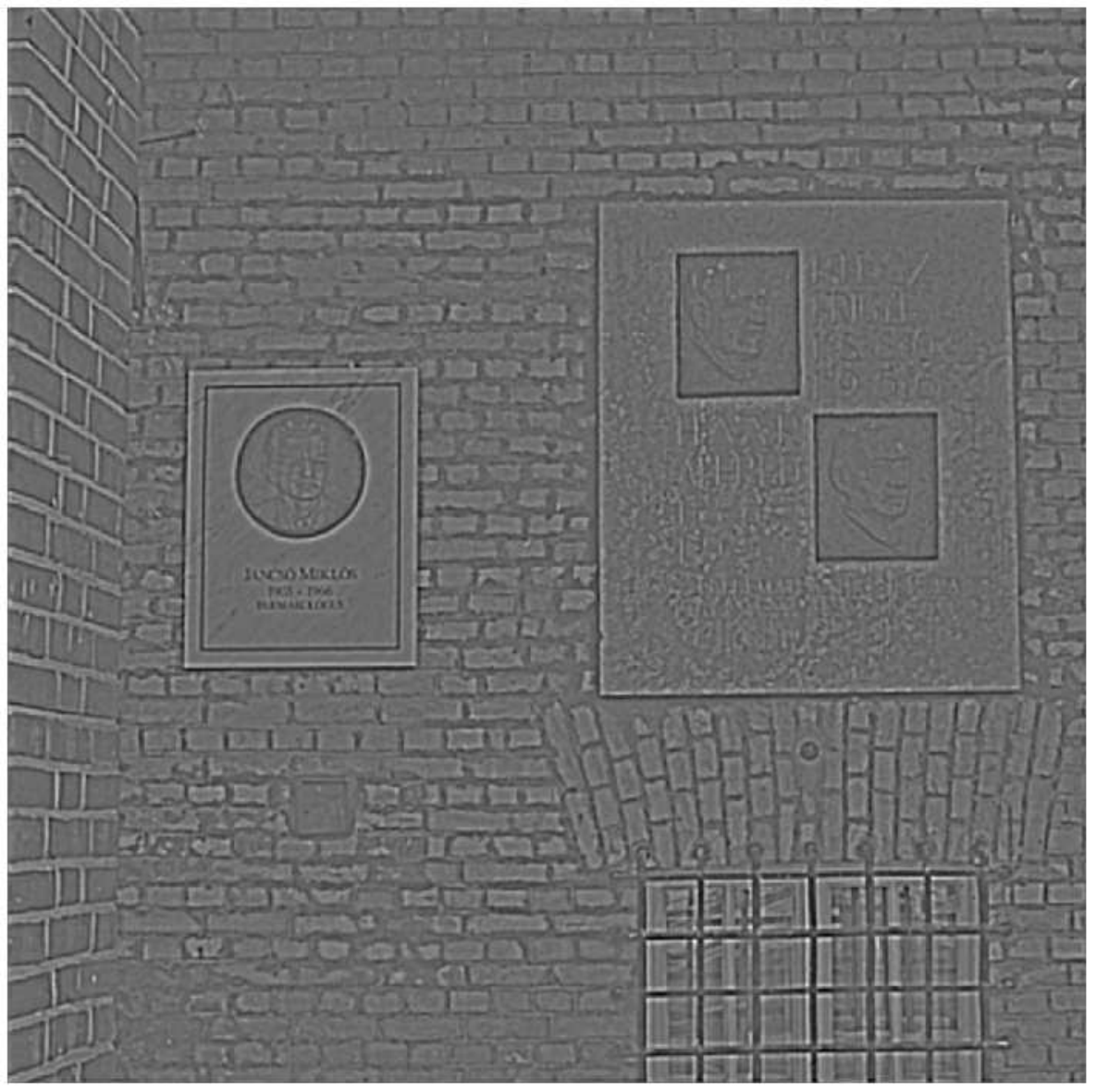}
}
  \caption{(a) The Marr wavelet (or Mexican hat). (b) Marr Wavelet singularity detector \imageHRP.}
\end{figure}

The continuous wavelet transform of a signal $f\in\LL^2(\Rbb)$ probes
its content with a ``lens'' $\psi_{(b,a)}$ of zoom factor $a$ and
location $b$. Mathematically,
\begin{equation}\label{eq-continuous-wavtransf}
W_f(b,a)\ =\ \int_{\Rbb}\ f(t)\ \tfrac{1}{\sqrt{a}}\,\psi^{*}(\tfrac{t-b}{a})\ \ud
t\ =\ \scp{\psi_{(b,a)}}{f}.
\end{equation}
Interestingly, provided that $\psi$ is \emph{admissible}, \ie when the
two constants $c^{\pm}_{\psi} =
2\pi \int_{0}^{+\infty}\tfrac{|\hat\psi(\pm\omega)|^{2}}{\omega} \ud\omega
< \infty$ are finite and equal\footnote{When $\psi$ is sufficiently
  regular, this condition reduces to a zero-average requirement, that is,
  $\int_{\Rbb} \psi(t)\ \ud t = 0$}, that is,
$c^+_\psi=c^-_\psi=c_\psi < \infty$, the signal $f$ may be recovered
from the coefficients $W_f(b,a)$:
\begin{equation}
f(t)\ =\
\tfrac{1}{c_{\psi}}\int_{0}^{+\infty}\!\!\!\int_{\Rbb}\ W_f(b,a)\
\psi_{(b,a)}(t)\ \ud b\,\tfrac{\ud
a}{a^{2}}.\label{eq:rec-ond}
\end{equation}
This integral representation involves wavelets at every location and
all positive dilations, \ie $f$ is decomposed on the continuous set of
functions $\{\psi_{(b,a)}:a\in\Rbb^*_+,\,b\in\Rbb\}$. Many different
kinds of (admissible) wavelets may be selected. We may cite the
derivatives of Gaussian (DoG), the Morlet and the Cauchy
wavelets, etc. Their selection is driven by the features  to
be elucidated in the data, \eg  frequency content with the Morlet
wavelet or  singularities with DoGs
(Fig.~\ref{fig:mexicanhat}) as illustrated\footnote{The YAWTb toolbox has been used, see \url{http://rhea.tele.ucl.ac.be/yawtb/}.} in
Fig.~\ref{fig:fig_SzegedMarrWavelet}.

In two dimensions, the most natural extension of the \nDim{1}-CWT is
obtained by considering isotropic wavelets, \ie wavelets $\psi\in
\LL^2(\Rbb^2)$ such that $\psi(\bs x)=\psi_{\rm rad}(\|\bs x\|)$, with
$\bs x=(x_1,x_2)$, for some radial function $\psi_{\rm
  rad}:\Rbb_+\to\Rbb$. In that case, the wavelet family is generated
by \nDim{2} dilations and translations, \ie we work with $\psi_{(\bs
  b,a)}(\bs x)=\tinv{a}\psi(\tfrac{\bs x - \bs b}{a})$ that are copies
of $\psi$ translated to $\bs b=(b_1,b_2)\in\Rbb^2$ and dilated by
$a>0$. The \nDim{2} CWT of the image $f\in \LL^2(\Rbb^2)$ is then simply
$W_f(\bs b,a)=\scp{\psi_{(\bs b,a)}}{f}$ and the reconstruction of $f$
is guaranteed by
\begin{equation}
  f(\bs x)\ =\
  \tfrac{2\pi}{c_{\psi}}\int_{0}^{+\infty}\!\!\!\int_{\Rbb^2}\ W_f(\bs b,a)\
  \psi_{(\bs b,a)}(\bs x)\ \ud^2 \bs b\,\tfrac{\ud
    a}{a^{3}},\label{eq:rec-ond-2d}
\end{equation}
if $c_\psi = (2\pi)^2 \int_{\Rbb^2} |\hat\psi(\bs k)|^2/\|\bs k\|^2\
\ud^2\bs k< \infty$. 
The isotropic CWT is a useful analysis tool for edge detection in
images.  For instance, by taking the (admissible) Marr Wavelet
$\psi(\bs x) = \Delta[\exp -\tinv{2}\|\bs x\|^2]$ (with $\Delta$ the
\nDim{2} Laplacian) also called Laplacian of Gaussian or Mexican Hat
(see Fig.~\ref{fig:mexicanhat}), the CWT of an image $f$ acts as a
multiscale edge detector.  The topic of \nDim{1} and \nDim{2}
continuous wavelet transforms is covered in more details in
\cite{Grossmann_A_1984_misc_dec_fwcsrt,
  Antoine_J_1993_j-sp_ima_atdcwt, Daubechies_I_1992_book_ten_lw,
  Holscheider_M_1995_book_wav_at,Mallat_S_2009_book_wav_tspsw}.

\subsection{Discrete Scale-Space Representations}
\label{sec:introtools-discss}

Numerical computation requires that continuous expansions such as \eqref{eq-continuous-wavtransf} and \eqref{eq:rec-ond-2d} be discretized.
In this
section, we detail some parameter samplings, such as dyadic or
translation invariant grids. Together with a suitable choice of the
wavelet function, they lead to stable representations where the
original signal can be perfectly reconstructed from its coefficients.

\subsubsection{Multiresolution Analysis (MRA)}
\label{sec:introtools-discss-mra}

In the context of a dyadic sampling where $a=2^{j}$ and $b=n2^{j}$ for
$j,n\in\mathbb{Z}$, the canonical way to design a suitable wavelet
function $\psi$ in \nDim{1} makes use of a multi-resolution analysis
(MRA). It is defined as a nested sequence of closed vector subspaces
$\left(V_{j}\right)_{j\in\mathbb{Z}}$ in $\LL^{2}(\mathbb{R})$
verifying standard properties \cite{Mallat_S_1989_tpami_the_msdwr}.
Multiresolution analysis of a signal $f$ consists of successively
projecting the signal onto subspaces $V_{j}$ in a series of
increasingly coarser approximations as $j$ grows.  The difference
between two successive approximations represents \emph{detail}
information. It amounts to the information loss between two
consecutive scales, which lies in the subspace $W_{j}$, the orthogonal
complement of $V_{j}$ in $V_{j-1}$ such that:
\[
	V_{j-1}=V_{j}\oplus W_{j}.
\] 
Then, with additional stability
properties, there exists a wavelet $\psi \in \LL^2(\RR)$ such that
$\Basis = \{2^{-j/2}\psi(2^{-j}x-n):n\in\Nbb\}$ is an orthonormal
basis for $W_j$.  

\subsubsection{Separable Orthogonal Wavelets}
\label{subsubsec-iso-wav}

\begin{figure}[htbp]
  \centering
  \includegraphics[height=8cm]{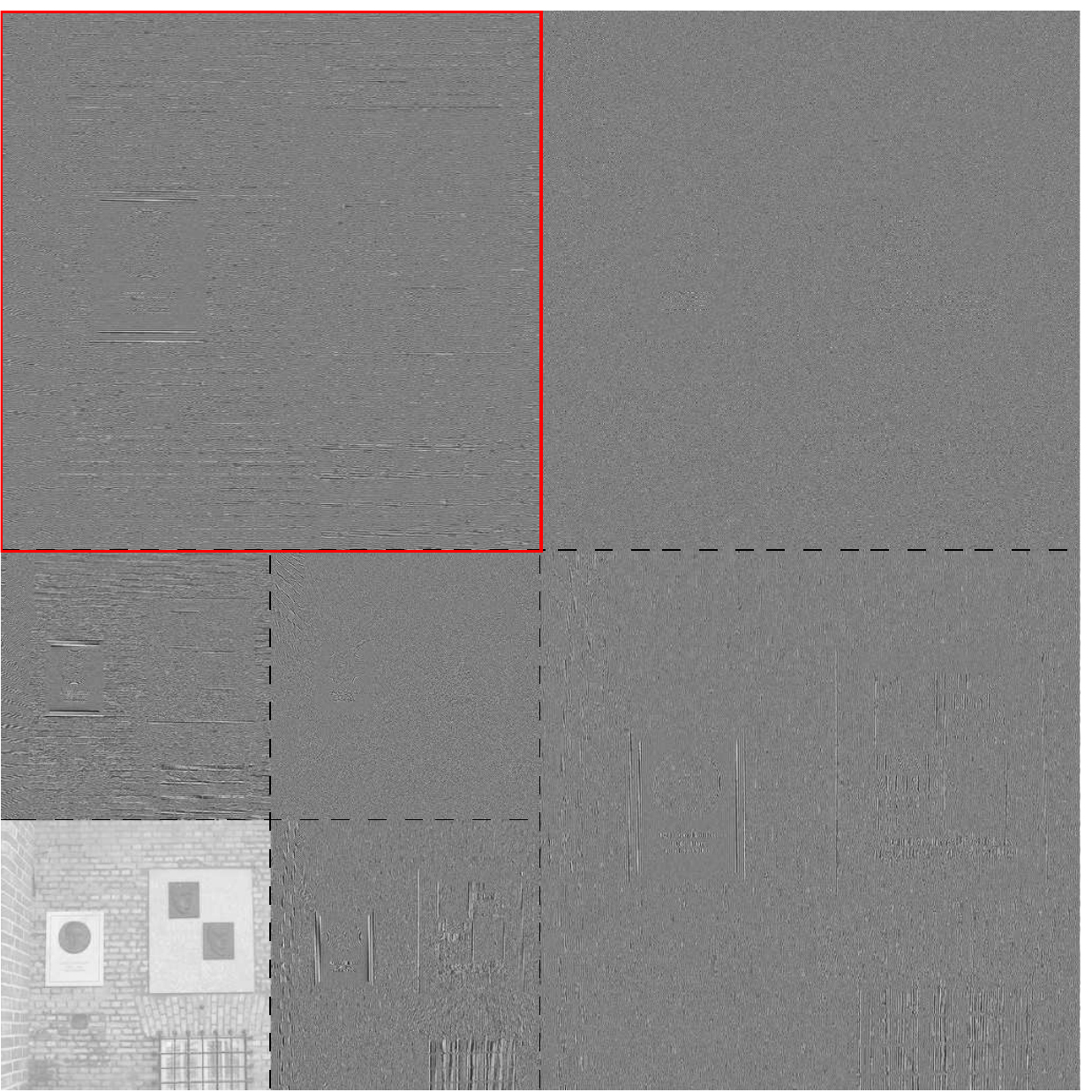}
   \caption{Dyadic wavelet decomposition \imageHRP.}
  \label{fig:fig_SzegedDyadicWavelet}
\end{figure}

A \nDim{2} orthogonal wavelet basis $\Basis = \{\atom_{\m}\}_{\m}$ of
$\Ldeux(\Rbb^2)$ for $\m=(j,\bs n,k)$ is parameterized by a
scale\footnote{Here and throughout the rest of the paper, we use the
  convention that scale increases with $j$, as in $s=2^j$. The
  converse convention is also often used in the literature.}  $2^j$
($j\in\Zbb$), a translation $2^j \bs n = 2^j(n_1,n_2)$ ($\bs
n\in\Zbb^2$) and one of three possible orientations $k \in \{V,H,D\}$,
loosely denoting the vertical, horizontal and (bi) diagonal
directions, the latter being poorly representative.  Wavelet atoms are
defined by dyadic scalings and translations $\atom_{\m}(\bs x) =
2^{-j} \atom^k( 2^{-j}\bs x-\bs n )$ of three tensor-product \nDim{2}
wavelets
$$
\atom^V(\bs x) = \psi(x_1)\phi(x_2), \quad \atom^H(\bs x) =
\phi(x_1)\psi(x_2), \qandq \atom^D(\bs x) = \psi(x_1)\psi(x_2),
$$
where $\phi$ and $\psi$ are respectively \nDim{1} orthogonal scaling
and wavelet functions, see
\cite{Daubechies_I_1992_book_ten_lw,Vetterli_M_1995_book_wav_sc,Mallat_S_2009_book_wav_tspsw}.
When the scale interval is limited to $j<J$ for some $J\in\Zbb$, the
basis $\Basis$ is completed by the functional set $\mathcal
A=\{\phi_{(J,\bs n)}\}_{\bs n}$, with the \nDim{2} separable scaling
function $\phi(\bs x)=\phi(x_1)\phi(x_2)$. This set gathers all the
coarse scale wavelet atoms with $j\geq J$.  The standard cascade image
is depicted in Fig.~\ref{fig:fig_SzegedDyadicWavelet}. It is now
critically sampled, \ie free from redundancy (compare
Fig.~\ref{fig:fig_SzegedPyramidLaplacian} and
\ref{fig:fig_SzegedMarrWavelet}). The approximation coefficients in
$\mathcal A$, a coarse image approximation at scale $J$, are
represented in the bottom-left square of
Fig.~\ref{fig:fig_SzegedDyadicWavelet}. The other squares in this
picture, associated to the ``bands'' $\{V,H,D\}$ for $j<J$, exhibit
some sparsity (few important coefficients), and horizontal and
vertical edges are relatively well captured.

A non-linear approximation $f_M = \ThreshH_T(f, \Basis)$ in an
orthogonal separable wavelet basis is efficient for smooth images or
images with point-wise singularities. The approximation of a piecewise
smooth image with edges of finite length decays like $\norm{f-f_M}^2 =
O(M^{-1})$. This result extends to functions with bounded variations
\cite{Cohen_A_1999_j-am-j-math_non_lasbvr2}, and is asymptotically
optimal. This decay is nevertheless not improved when the edges are
smooth curves, because of the fixed ratio between the horizontal and
the vertical sizes of the orthogonal wavelet support.

\subsubsection{Fast Algorithms for Finite Images}

A finite discretized image $f \in \CC^{N_1\times N_2}$ of $N=N_1N_2$
pixels fits into the MRA framework by assuming that the pixel values
of $f_{\bs n}$ on $\bs n=(n_1,n_2)$ are the coefficients
$\scp{\phi_{(J,\bs n)}}{\tilde f}$ of some continuous
function $\tilde f \in \Ldeux(\RR^2)$ at a fixed resolution $V_J$,
where $2^{2J} = N$.

The coefficients $\dotp{\atom_{j,\bs n}^k }{f}$ of $\tilde f$ for $j
\geq J$ are computed from the discrete image $f$ alone.  This computation is
performed using a cascade of filters interleaved with downsampling
operators \cite{Mallat_S_1989_tpami_the_msdwr}. For compactly
supported wavelets, this requires $O(N)$ operations. Symmetric bi-orthogonal wavelet bases with compact support ease the implementation of
non-periodic boundary conditions \cite{Cohen_A_1992_j-comm-acm_bio_bcsw}.
For infinite
impulse response (IIR) wavelet filters,  computations in the Fourier
domain require $O(N \log(N))$ operations \cite{Rioul_O_1992_tit_fas_adcwt}, while recursive implementations 
\cite{Smith_M_1995_j-ieee-tip_rec_tvfbsic} allow signal-adaptive implementation.

While separable wavelets are not optimal for approximating generic
edges, they lie at the heart of early state-of-the-art
methods for compression and denoising. The JPEG 2000 coding standard
\cite{Taubman_D_2002_book_jpe_2000icfsp} performs an embedded
quantization of wavelet coefficients, and uses an adaptive entropic
coding scheme that takes into account the local dependencies across
wavelet coefficients.  The sub-optimality of wavelets for 
the sparse representation of edges
can be alleviated using block thresholding of
groups of wavelet coefficients \cite{Cai_T_1999_ann-stat_ada_webtoia},
that gives improvements over scalar thresholding. Advanced statistical modeling
of wavelet coefficients leads to denoising methods close to the
state-of-the-art, see for instance
\cite{Muller_P_1999_book_bay_iwbm,Portilla_J_2003_tip_ima_dsmgwd,Chaux_C_2008_j-ieee-tsp_non_sbemid}.

\subsubsection{Translation Invariant Wavelets}

Given a discrete frame $\Basis = \{ \psi_{\m} \}_m$ of $\CC^N$,
$\Basis$ is \emph{translation invariant} if $\psi(\cdot-\tau) \in \Basis$
for any $\psi\in\Basis$ and any integer translation $\tau$. This
property tends to reduce artifacts in image restoration problems like
denoising, since, for such invariant frame, the thresholding operator
$\ThreshH_T(f, \Basis)$ becomes itself translation invariant.
Discrete orthogonal wavelet bases described in the previous sections
are not translation invariant and many authors have worked on
recovering this useful capability.

For instance, cycle spinning, proposed by Coifman and Donoho in
\cite{Coifman_R_1995_was_tra_id}, reduces wavelet artifacts by
averaging the denoising result of all possible translates of the
image, thus resulting in a translation invariant processing. For an
orthogonal basis $\Basis = \{ \psi_{\m} \}_{\m}$, this is equivalent
to considering a tight frame which is the union of all translated
bases $\{ \psi_{\m}(\cdot-\tau) \}_{\m, \tau}$. For a generic basis,
this frame has up to $N^2$ atoms. For a wavelet basis, the frame has
$O(N \log(N))$ atoms, and the coefficients are computed with the fast
``\`a trous'' algorithm in $O(N \log(N))$
\cite{Rioul_O_1992_tit_fas_adcwt,Shensa_M_1992_j-ieee-tsp_dis_wtwatma}.
  The translation invariant paradigm additionally draws a connection
  between the scale-space formalism (Sec. \ref{sec:introtools-contss})
  \cite{Chambolle_A_2001_j-ieee-tip_int_tiwsnisss} and thresholding
  (Sec. \ref{sec:introtools-representation}). Several \nDim{2} design
  described in the next sections attempt to (approximately) address
  invariance (translation/rotation) without sacrificing computational
  efficiency.

\section{Oriented and Geometrical Multiscale Representations}
\label{sec:oriented}
The variety of oriented and geometric multiscale representations
proposed over the last few years requires broad grouping, arranged as
follows:  Sec.~\ref{sec:dir-outcrops} presents directional
methods closely related to \nDim{1} decompositions. In
Sec.~\ref{sec:non-separ-direct}, the directionality is addressed with
diverse non-separable schemes. Finally, in
Sec.~\ref{sec:direct-anis-scal}, directionality is 
attained
by an
anisotropic scaling of the atoms that yields various efficient edge and curve
representations.

\subsection{Directional Outcrops from Separable Representations}
\label{sec:dir-outcrops}
\subsubsection{Improved Separable Selectivity by Relaxing Constraints}
\label{sec:adaptedgeometry-mband}

As discussed in Sec.~\ref{sec:introtools-discss-mra}, discrete orthogonal
wavelets may be viewed as a peculiar instance of orthogonal filter
banks \cite{Vaidyanathan_P_1993_book_mul_sfb}.  A well-known
limitation in \nDim{1} is that orthogonality (hence non-redundant),
realness, symmetry and finite support properties cannot coexist with
pairs of low- and high-pass filters, except for the Haar
wavelet. 

We decide to briefly mention here some of the early steps taken to
  tackle this limitation. These have also been employed in
  more genuine non-separable transforms, as seen later, typically
  relaxing one of the aforementioned properties, such as using
  infinite-support filters \cite{Blu_T_2000_icassp_fra_swtdi}, semi-
  or biorthogonal decompositions \cite{Cohen_A_1992_j-comm-acm_bio_bcsw}
  or complex filter banks \cite{Zhang_X_1999_tsp_ort_cfbwpd}.

  For instance, instead of a two-band filter bank, $M$-band wavelets
  \cite{Steffen_P_1993_tsp_the_rmbwb} with $M> 2$ provide alternatives
  where symmetry, orthogonality and realness are compatible with
  finitely supported atoms. In this setting, the approximation and the
  $M$-band detail spaces are $V_j$ and $(W^m_j)_{m \in \NMs}$ related
  through $V_{j-1}=V_j \oplus \bigoplus_{m=1}^{M-1}W^m_j$ for a
  resolution level $j$. This versatile design provides filters that
  suffer less aliasing artifacts with increased regularity. Their
  finer subband decomposition is also beneficial for detecting
  orientations in a more subtle fashion than with the $\{V,H,D\}$
  quadrants obtained with standard wavelets
  (Sec.~\ref{subsubsec-iso-wav}). Yet, more general $M$-adic MRAs are
  possible, for instance with a rational $M= p/q, M > 1$  \cite{Auscher_P_1992_in-coll-wav_bl2rrdf,Blu_T_1993_tsp_ite_fbrccdwt,Blu_T_1998_tsp_new_datborforw,Baussard_A_2004_j-sp_rat_mafwtawsd,Bayram_I_2009_j-ieee-tsp_fre_ddordwt}. Note
  that for specific purposes such as compression, $M$-band filter
  banks with $M = 2^J, J\in \NN$ may be treated like a $J$-level
  dyadic tree and combined in a hierarchical transform
  \cite{Xiong_Z_1996_j-ieee-spl_dct_beic,Malvar_H_2000_p-dcc_fas_picww}. Satisfying
  the MRA axioms is not necessary in practice  in order to
  yield high performance results. This is suggested by recent image
  and video coders focusing on ``simpler'' transforms, closer to
  ancient Walsh-Hadamard transforms than to more involved wavelets \cite{Malvar_H_2003_j-ieee-tcsvt_low_ctqh264avc}.

  Alternatively, the \nDim{1} decomposition on rows and columns of
  images may be performed in a more anisotropic manner, as in
  \cite{Rosiene_C_1999_p-iscas_ten_pwmdca,Xu_D_2003_p-spie-wasip_ani_2dwprtta}.
  An additional relaxation comes from lifting the critically sampled
  scheme, yielding oversampled, translation-invariant (see Sec.~\ref{sec-tree-wavcospackets}) multiscale
  wavelets, wavelet/cosine packets  or frames  \cite{Coifman_R_1995_was_tra_id,Nason_G_1995_incoll-was_sta_wtsa,Pesquet_J_1996_tsp_tim_iowr,Cohen_I_1997_j-sp_ort_sialtd,Chui_C_2002_p-acha_com_stsfmvm,Daubechies_I_2003_j-acha_fra_mrabcwf}. Multidimensional
  oversampled filter banks in $\nDim{n}$ with limited redundancy may
  be designed as well \cite{Aach_T_2000_sp_lap_dtsiaare,Tanaka_T_2004_tsp_gen_lpbtolpprfbls,Zhou_J_2005_spie-wav_mul_ofb,Tanaka_T_2006_tsp_dir_doprfirfbfof,Gauthier_J_2009_tsp_opt_socfb}.

\subsubsection{Pyramid-related wavelets}
\label{sec:oriented-marr}
Notably influenced by \cite{Simoncelli_E_1992_tit_shi_mst,Simoncelli_E_1995_icip_ste_pfamdc}, Unser and Van de Ville propose a slightly redundant transform
\cite{Unser_M_2008_tip_pai_wbmrasr} based on a pyramid-like wavelet
analysis. This decomposition constitutes a wavelet frame with 
mild redundancy, which is nevertheless not steerable. 
Subsequently, the same authors propose a steerable analysis
\cite{VanDeVille_D_2008_tip_com_wbsmlp} based on polyharmonic
$B$-splines \cite{Forster_B_2008_j-acha_shi_isrcf} and the Maar-like
\cite{Marr_D_1980_j-proc-roy-soc-b_the_ed,Marr_D_1982_book_vis_cihrpvi} wavelet
pyramid. Such multiresolution analysis can easily be implemented via filter
banks as detailed in \cite{VanDeVille_D_2008_tip_com_wbsmlp} and the
total redundancy of this decomposition is $8/3$ (a redundancy of
$4/3$ is introduced by the pyramid structure and the complex nature of
the coefficients increases the redundancy by a factor of $2$). A similar approach based  on Riesz-Laplace wavelets is proposed in 
\cite{Unser_M_2011_j-ieee-tip_ste_ptwfl2rd}. The latter constructions are related to Hilbert and Riesz transforms.

\subsubsection{Complexifying Discrete Wavelets with Hilbert and Riesz}
\label{sec:adaptedgeometry-complex}
Different kinds of complexification are indeed a possible option in order to tackle the
problem of poor directionality with  classical wavelet transforms.
The common basic idea leans toward 
analytic wavelets and their combination  to improve the \nDim{2}
directionality. Behind a generic notion  of complex wavelets reside different
approaches detailed hereafter, which require the  definition of   some basic tools.

We first introduce the Hilbert transform, termed  ``complex signal'' in \cite{Gabor_D_1946_j-iee_the_c} and exhaustively mapped in \cite{King_F_2009_book_hil_t}. While the \nDim{1} Hilbert transform is unambiguously defined, 
there exists  multidimensional extensions, often obtained by tensor products, thus leading to
approximations. In order to increase the directionality
property, other multidimensional constructions (discussed in
\cite{Hahn_S_1992_proc-ieee_mul_cssos}) have also been proposed.
\begin{itemize}
\item The \nDim{1} Hilbert transform $\Hh$ of a signal $f$ is easily expressed in the Fourier domain as
\begin{equation}
 \Hh\{f\}(\omega) = -\icomplex \sign (\omega) \widehat{f}(\omega).
\label{eq:HT}
\end{equation}

\item The \nDim{1}  fractional Hilbert transform  $\Hh_\theta$ of $f$  is similarly defined in \cite{Chaudhury_K_2010_j-ieee-tsp_shi_dtcwt} by 
\begin{equation}
 \Hh_\theta\{f\}(\omega) = \exp(\icomplex \pi \theta \sign (\omega)) \widehat{f}(\omega).
\label{eq:fHT}
\end{equation}

\item The \nDim{2} directional Hilbert transform $\Hh_\theta$ of $f$ is one of the \nDim{2} extensions defined in \cite{Chaudhury_K_2010_j-ieee-tsp_shi_dtcwt} as
\begin{equation}
 \Hh_\theta\{f\}(\omega_1,\omega_2) =  -\icomplex \sign \big(\cos(\theta)\omega_1 + \sin(\theta) \omega_2\big) \widehat{f}(\omega_1,\omega_2).
\label{eq:dHT}
\end{equation}
See also \cite{Antoine_J_1999_j-acha_dir_wrcwsdp}.

\end{itemize}

\begin{figure}[htbp]
  \centering
  \includegraphics[height=7cm]{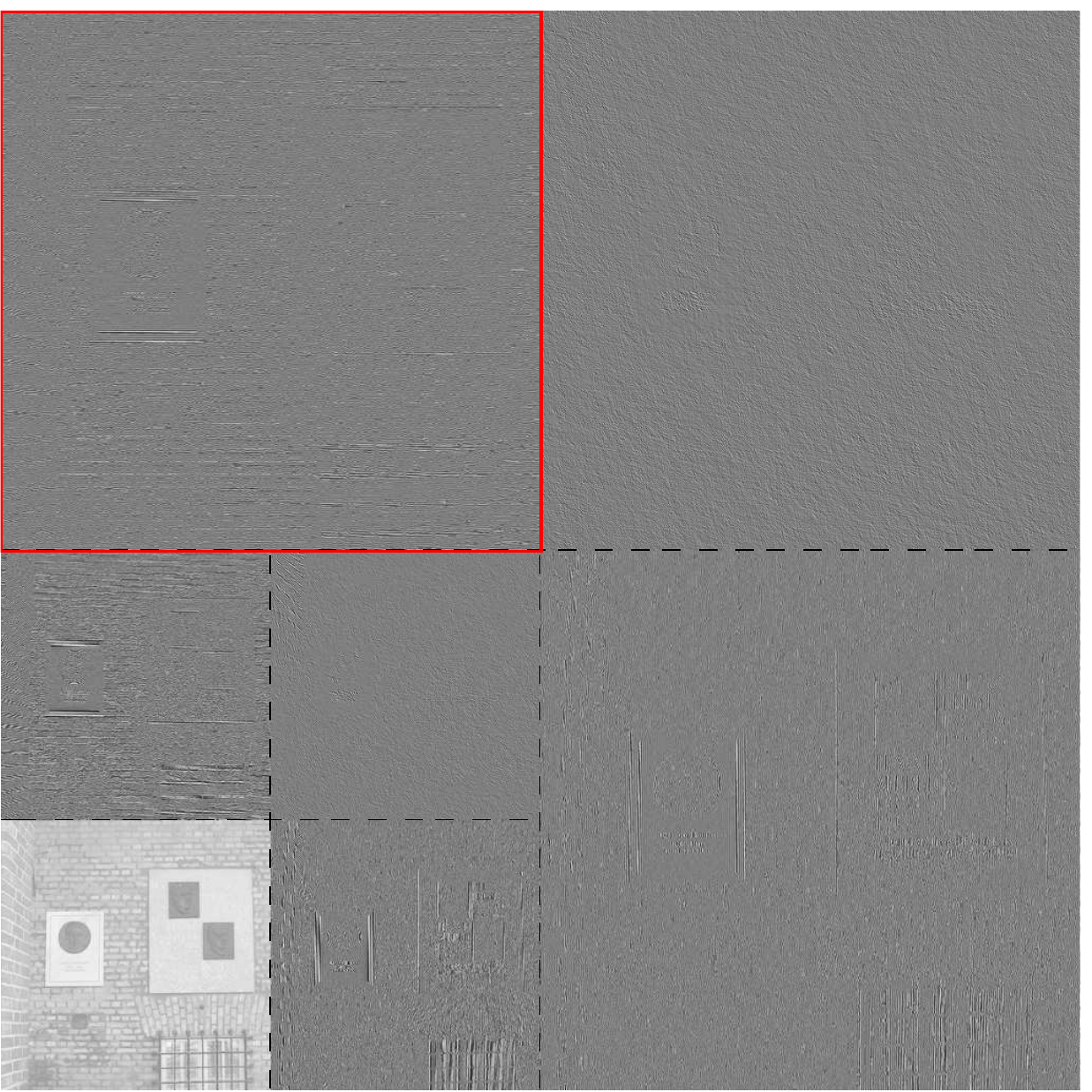}
  \includegraphics[height=7cm]{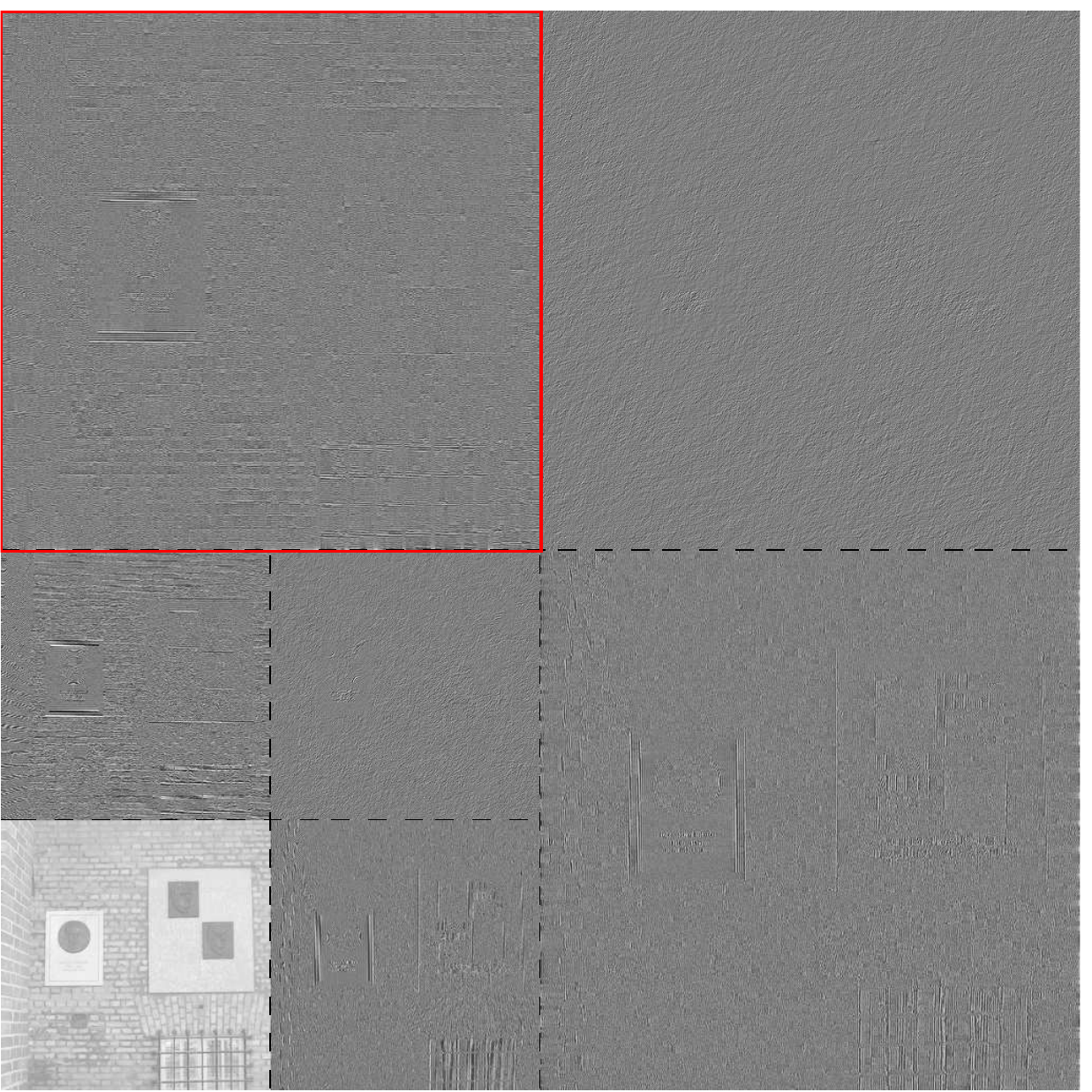}    \caption{Dyadic dual-tree wavelet decomposition \imageHRP.}
  \label{fig:fig_SzegedDualTreeDyadic}
\end{figure}

\begin{figure}[htb!]
  \centering
  \subfigure[\label{fig:fig_SzegedDualTreeDyadic-proj}]{
    	{\includegraphics[height=4cm,keepaspectratio]{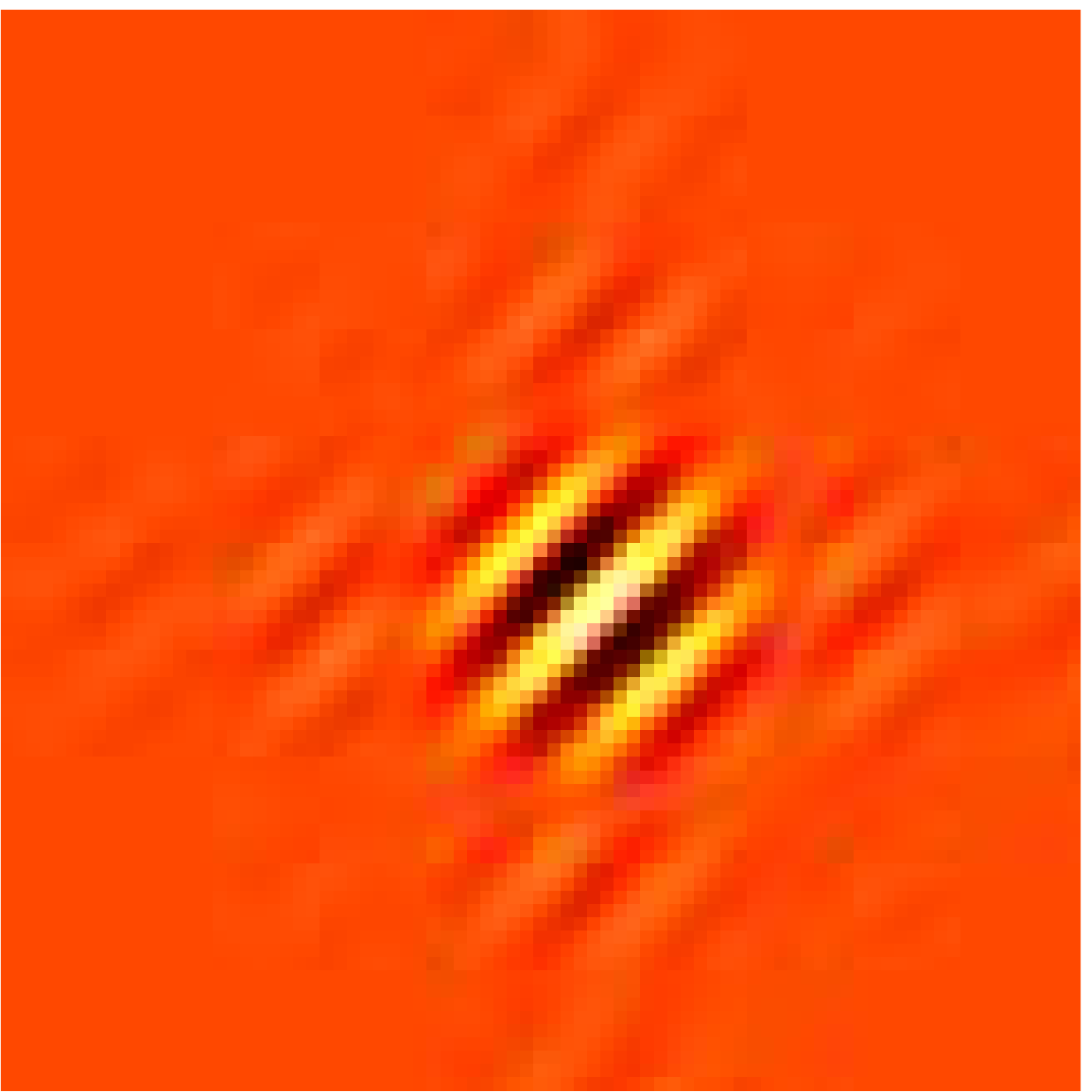}}}
  \subfigure[\label{fig:fig_SzegedDualTreeDyadic-freq}]{
    \includegraphics[height=5cm,keepaspectratio]{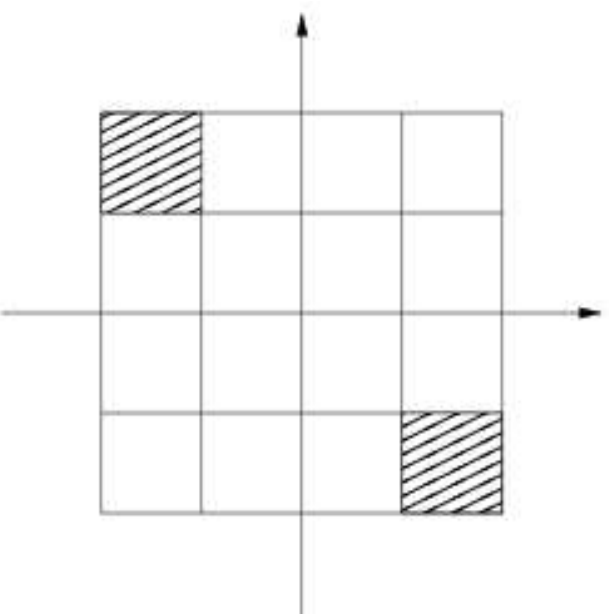}}
  \caption{The dyadic dual-tree wavelet. (a) Example of atom (diagonal
    wavelet). (b) Associated frequency partitioning.}
  \label{fig:fig_SzegedDualTreeDyadic-projfreq}
\end{figure}

\begin{figure}[htb]
  \centering
\subfigure[]{
\includegraphics[width=5cm]{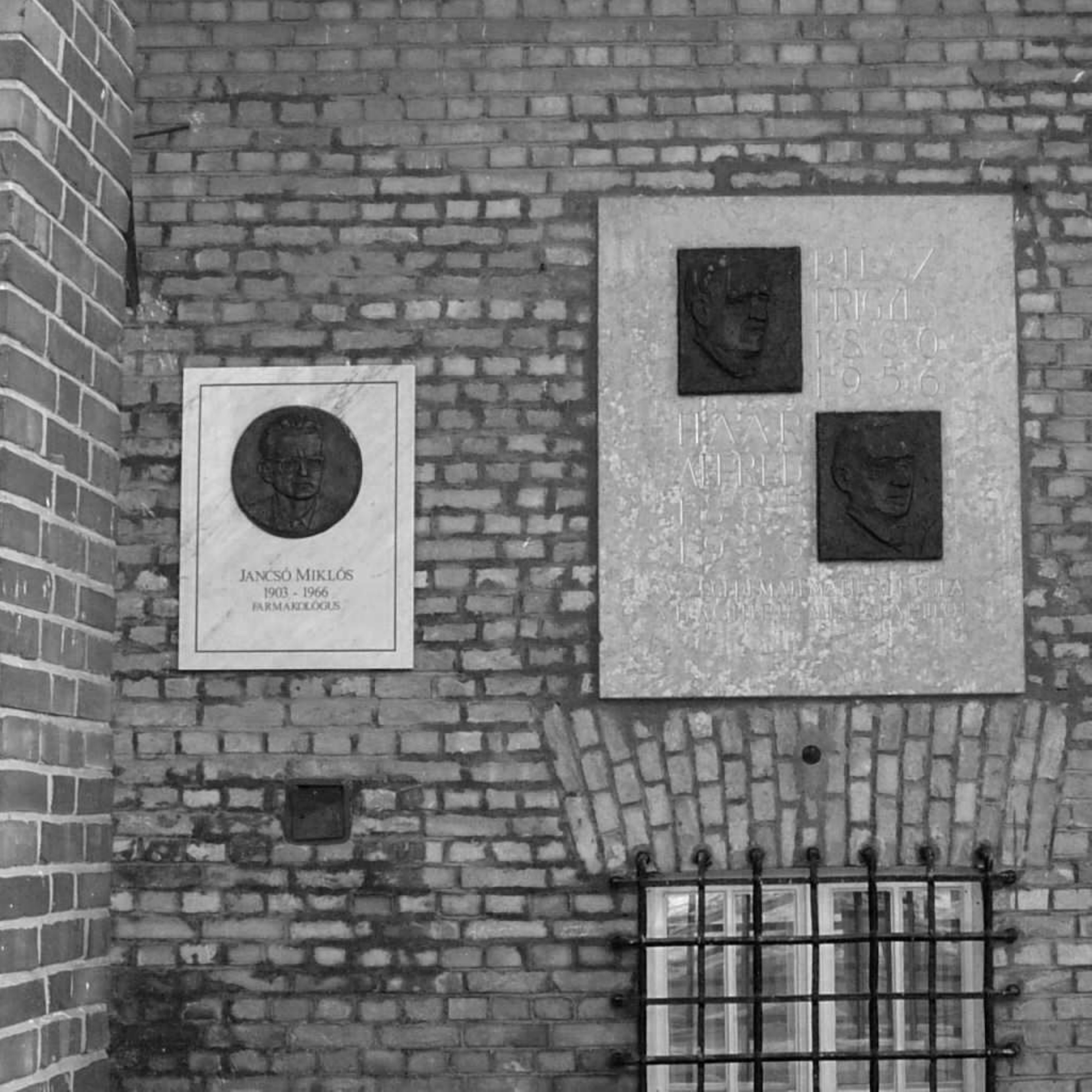}
}\addtocounter{subfigure}{1}\subfigure[\label{fig:fig_SzegedDualTreeDyadic-sub}]{
\includegraphics[width=5cm]{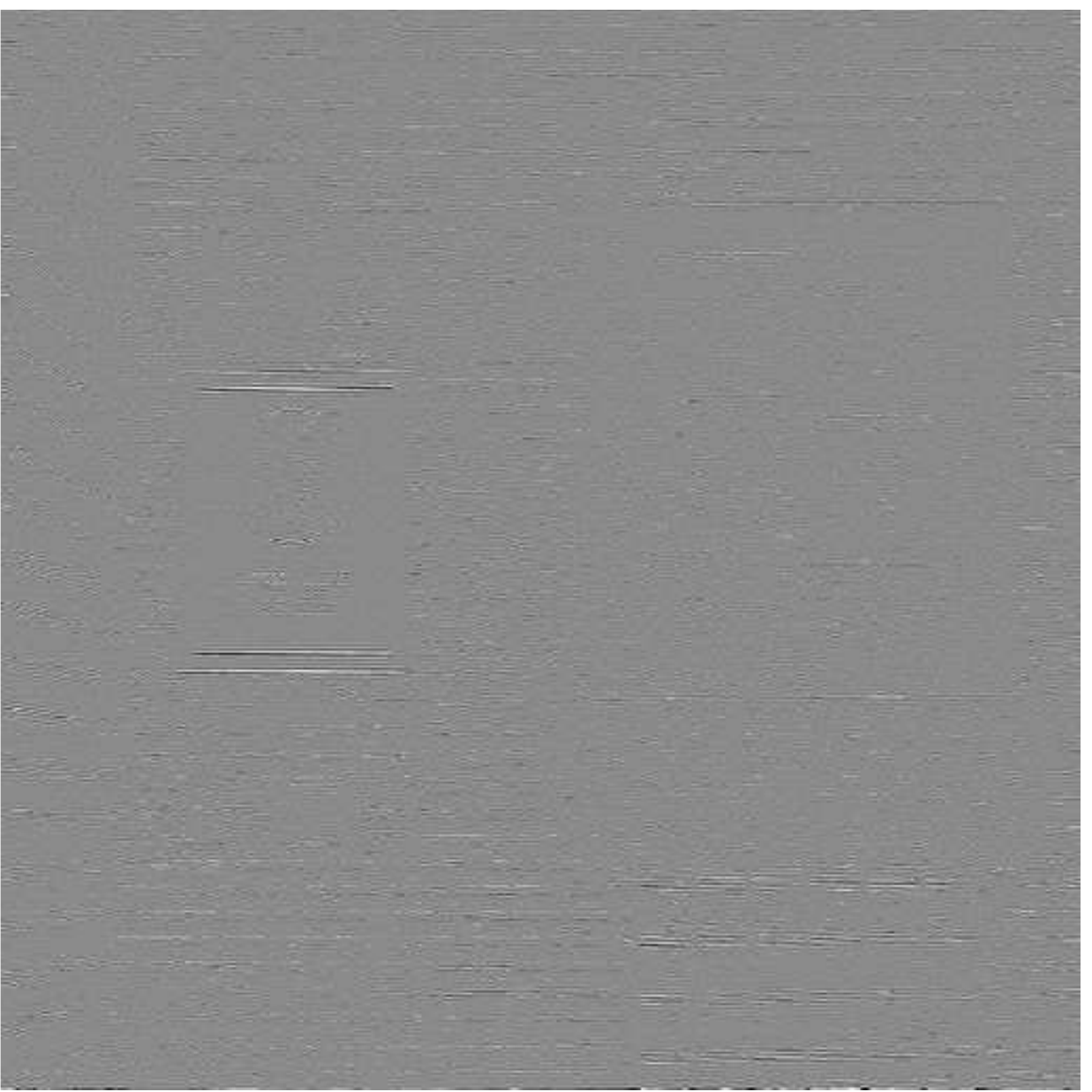} 
\includegraphics[width=5cm]{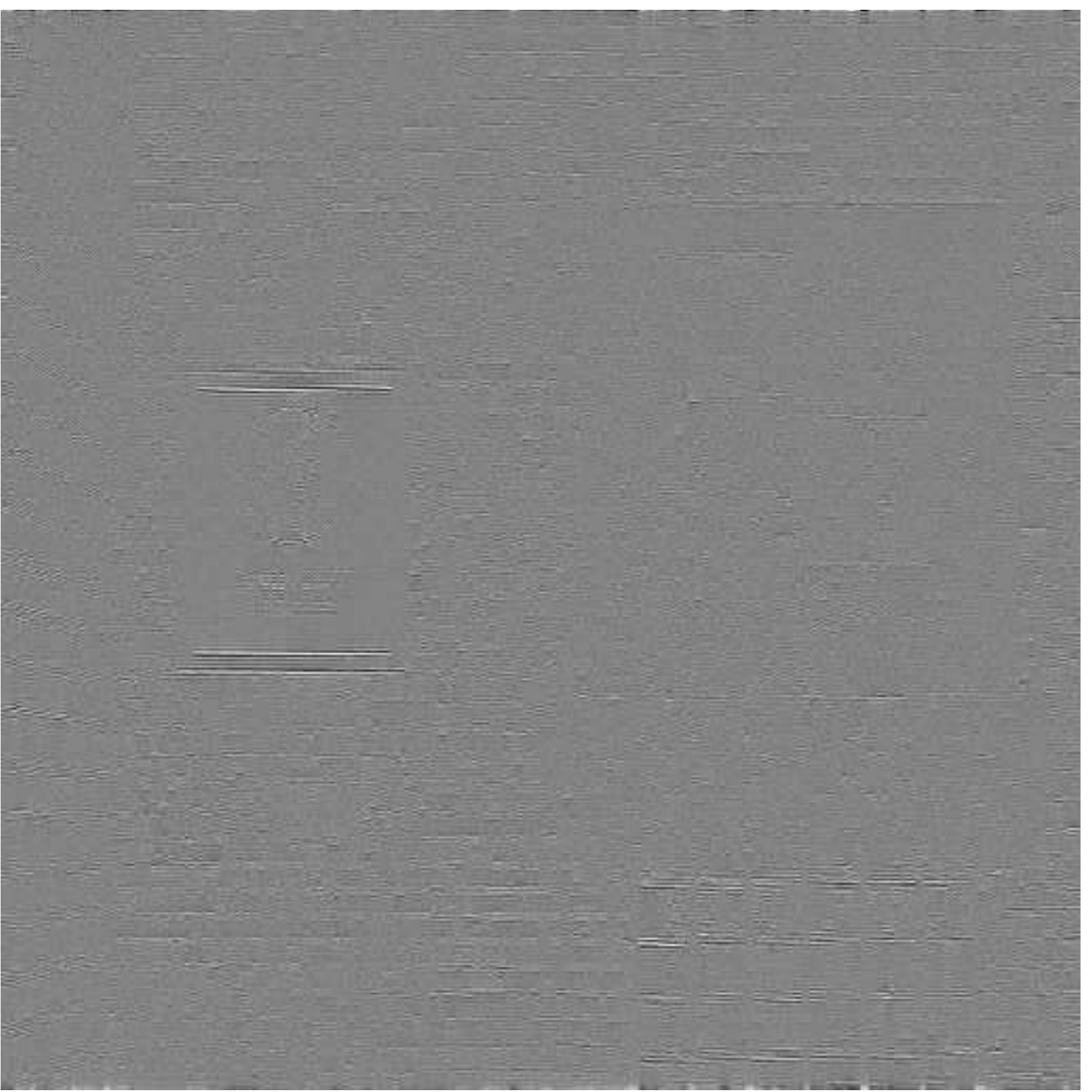}
}\\ \addtocounter{subfigure}{-2}\subfigure[\label{fig:fig_SzegedDWTDyadic-sub}]{
\includegraphics[width=5cm]{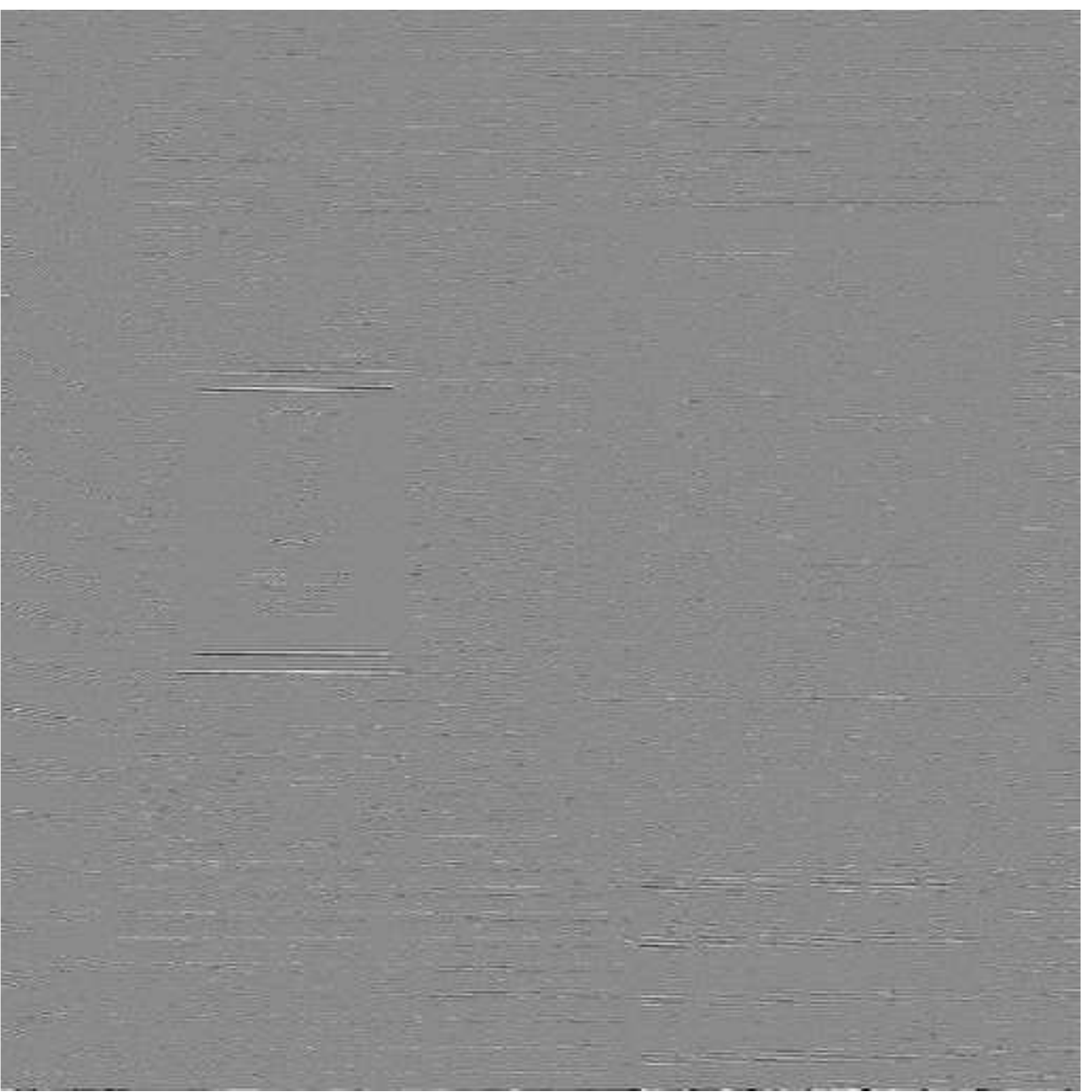}
}\addtocounter{subfigure}{1}\subfigure[\label{fig:fig_SzegedDualTreeMband-sub}]{
\includegraphics[width=5cm]{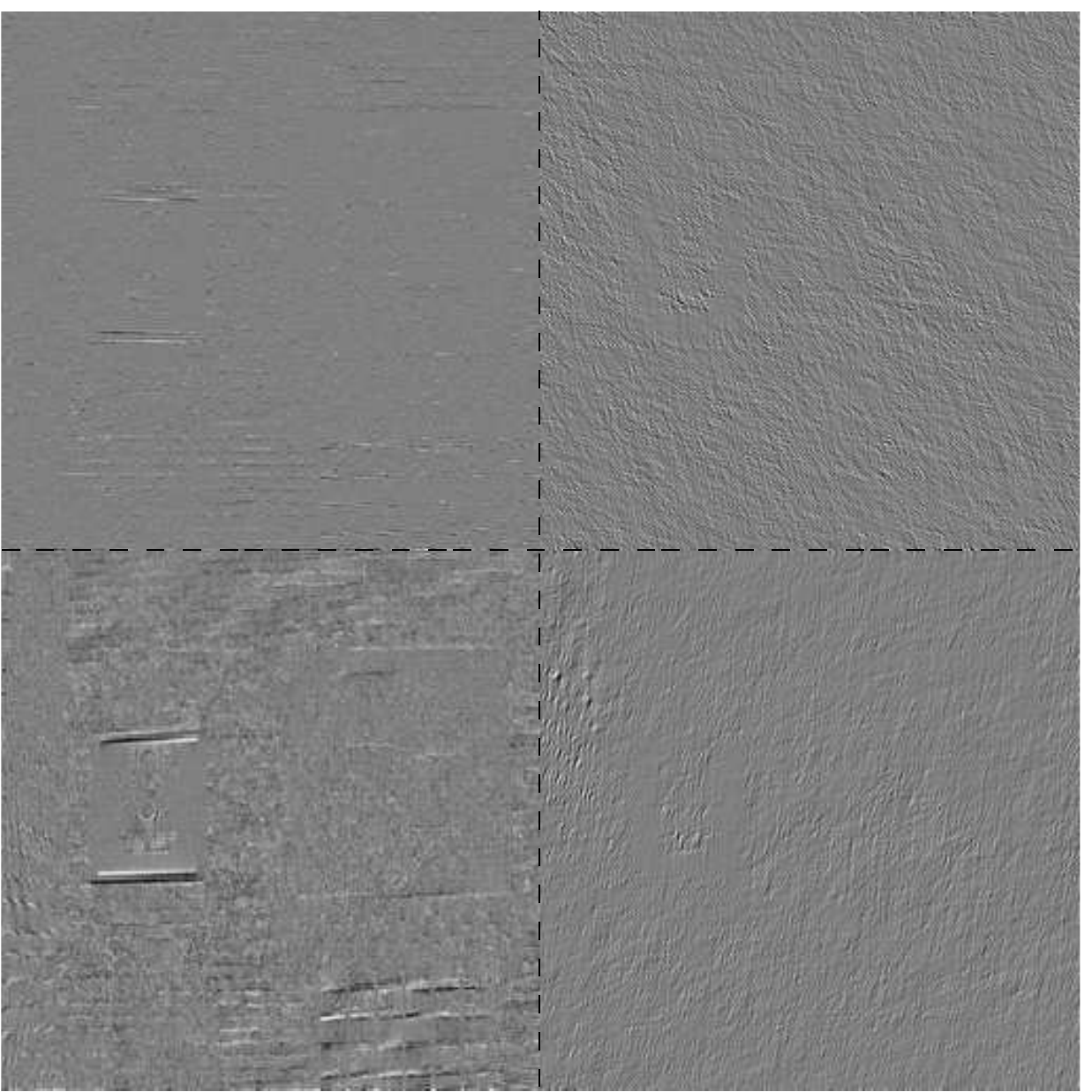}
\includegraphics[width=5cm]{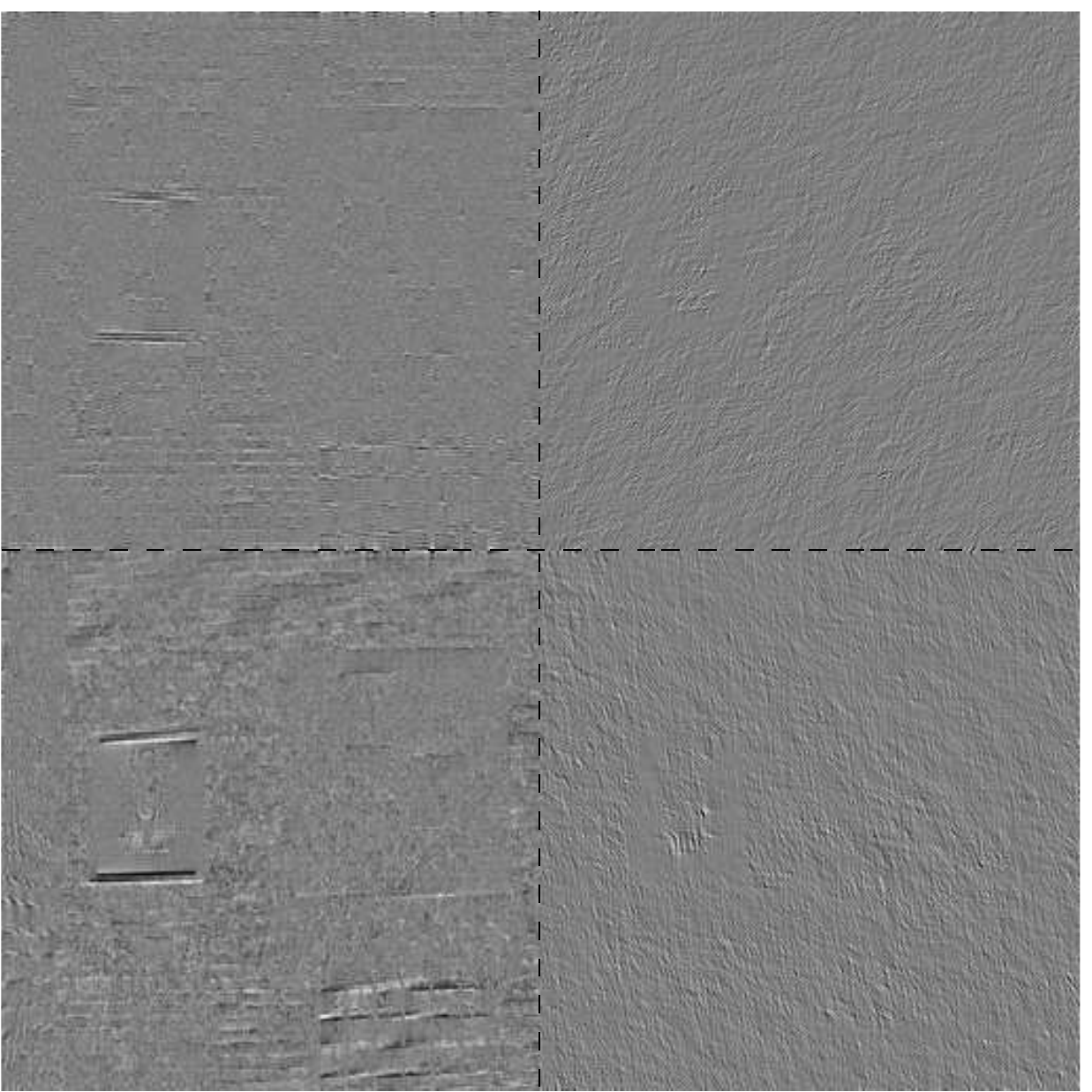}
} \caption{The original image (a) and the horizontal subband(s) at first
  resolution level for (b) Dyadic wavelet transform, (c) Dyadic
  dual-tree transform (primal+dual) and (d) $M$-band dual-tree wavelet
  decomposition (primal+dual) \imageHRP.}
  \label{fig:fig_SzegedDWTDTT}
\end{figure}

\begin{figure}[htb!]
  \centering
  \subfigure[\label{fig:fig_SzegedDualTreeMBand-projection}]{
    	{\includegraphics[height=4cm,keepaspectratio]{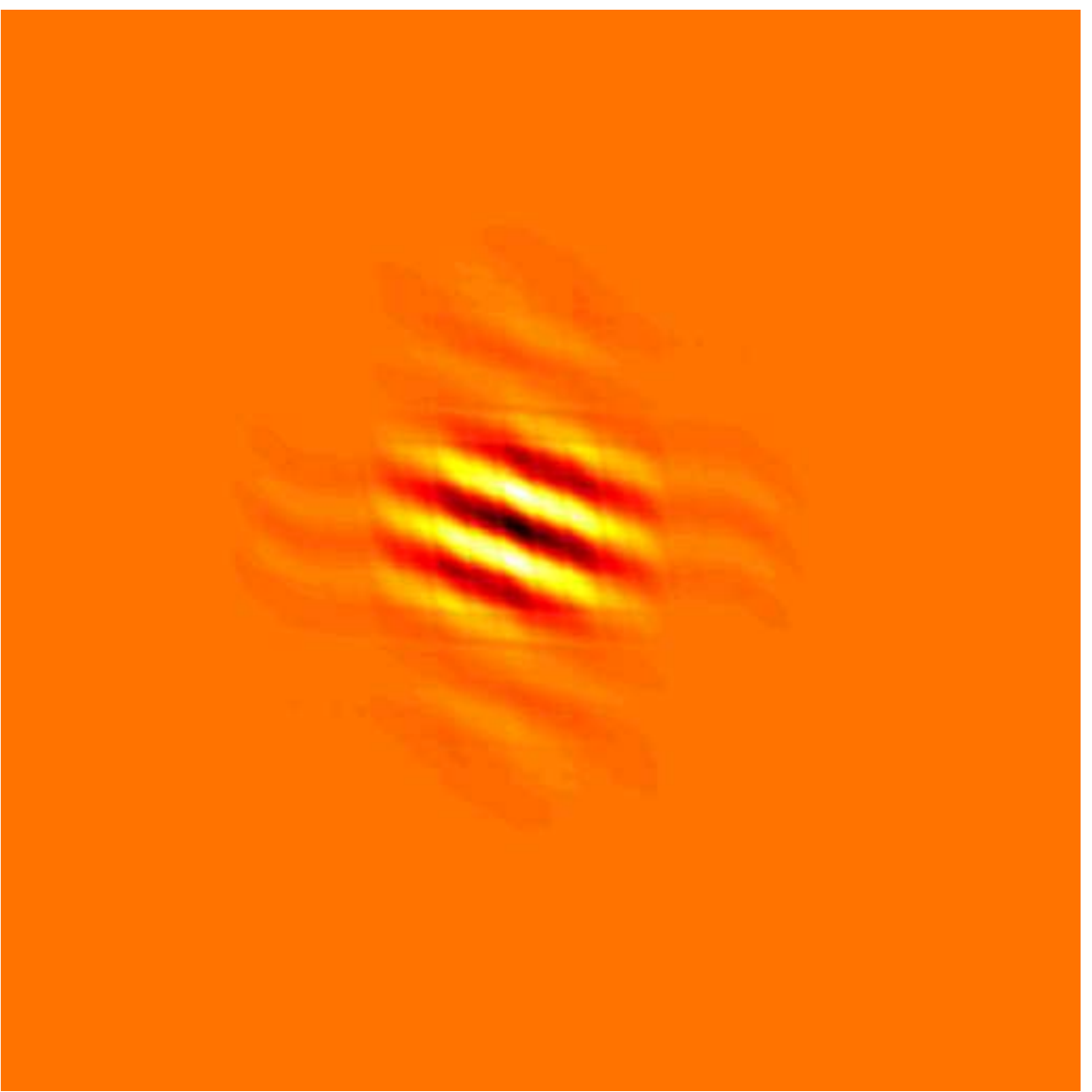}}}
  \subfigure[\label{fig:fig_SzegedDualTreeMBand-frequency}]{
    \includegraphics[height=5cm,keepaspectratio]{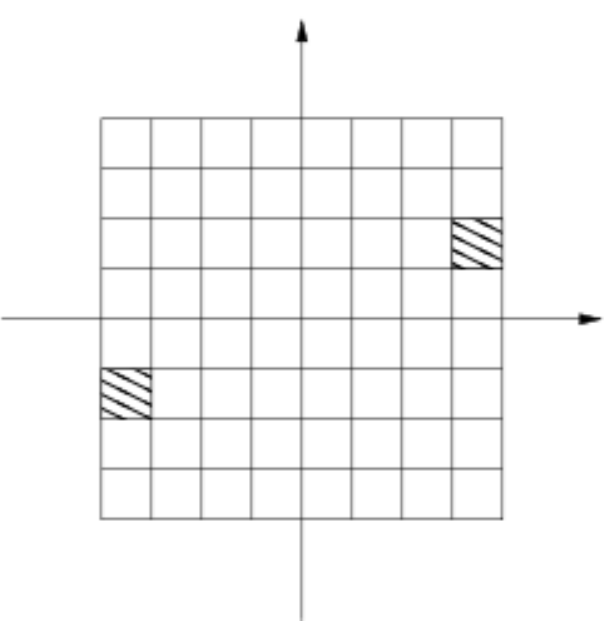}}
  \caption{The $M$-band  dual-tree wavelet. (a) Example of atom. (b) Frequency partinioning.}
  \label{fig:fig_SzegedDualTreeMBand-projfreq}
\end{figure}

The Hilbert transform was already associated with wavelets for transient
detection by Abry \etal \cite{Abry_P_1994_stfts_mul_td}. Others early connections  between wavelets and the Hilbert transform are drawn in 
\cite{Beylkin_G_1994_tfom-tma_tra_hbf,Weiss_J_1995_tr_hil_tww,Beylkin_G_1996_acha_imp_ofbashw}.
 At the end of the 1990's, Kingsbury
proposed the dual-tree transform based on even and odd filters 
\cite{Kingsbury_N_1998_p-ieee-dspw_dua_tcwtntsidf,Kingsbury_N_1999_j-phil-trans-roy-soc-lond-a_ima_pcw}. An
alternative construction is given by
Selesnick \cite{Selesnick_I_2001_spl_hil_tpwb}.  It amounts to
performing two discrete classical wavelet transforms in parallel, the
wavelets generated by the trees forming Hilbert pairs. An atom of the
corresponding basis (here the diagonal wavelet) and its
corresponding frequency plane tiling are depicted in
Fig.~\ref{fig:fig_SzegedDualTreeDyadic-projfreq}. The corresponding
dual-tree of wavelet coefficients is represented in
Fig.~\ref{fig:fig_SzegedDualTreeDyadic}, which clearly shows the separation of oriented structures with 
different orientations.  The resulting oriented wavelet dictionary has a small redundancy
and is also computationally efficient.  The corresponding wavelet is
approximately shift invariant, see~\cite{Selesnick_I_2005_spm_dua_tcwt} for more details.
It is extended to the $M$-band setting by Chaux \etal \cite{Chaux_C_2006_tip_ima_adtmbwt} and to wavelet packets in 
\cite{Jalobeanu_A_2001_p-icip_ima_deconv,Bayram_I_2008_tsp_dua_tcwpmbt}. 
In Fig.~\ref{fig:fig_SzegedDWTDTT}, one subband of the wavelet transform (red square in Fig. \ref{fig:fig_SzegedDyadicWavelet}),
two subbands (primal+dual) of the dyadic dual-tree transform (red squares in Fig. \ref{fig:fig_SzegedDualTreeDyadic}), as well as the corresponding
eight subbands (4 primal+4 dual) of the $4$-band dual-tree wavelet decomposition are depicted.
In Fig.~\ref{fig:fig_SzegedDualTreeMband-sub}, the fine oriented textures from the left side of the image are (slightly) better separated in some non-horizontal subbands. The wavelet/frequency tiling 
corresponding to the $4$-band dual-tree wavelet decomposition are depicted in
Fig.~\ref{fig:fig_SzegedDualTreeMBand-projfreq}. The main advantage
of this decomposition is that it achieves a directional image analysis with a small redundancy of a factor $2$ ($4$ for  the complex transform).

Gopinath \cite{Gopinath_R_2003_tsp_pha_tirnsiwt,Gopinath_R_2005_tsp_pha_f} has designed phaselets which is an extension of the dyadic dual-tree wavelet transform~\cite{Kingsbury_N_1998_p-ieee-dspw_dua_tcwtntsidf,Selesnick_I_2001_p-ciss_cha_dhtpwb}. They aim at improving translation invariance with a given redundancy, and are built by carefully observing the effects of shifts in a discrete wavelet transform. \nDim{2} phaselets are easily obtained by tensor products.

More recently, the shiftability of the dual-tree transform has been
studied by Chaudhury \etal
\cite{Chaudhury_K_2010_j-ieee-tsp_shi_dtcwt} by introducing the
fractional Hilbert transform \eqref{eq:fHT}. A \nDim{2} extension has
been proposed in \cite{Chaudhury_K_2009_p-spie-wasip_gab_wafht} and the
construction of Hilbert transform pairs of wavelet bases can be found
in \cite{Chaudhury_K_2009_j-ieee-tsp_con_htpwbglt}. Note that previous
works dealing with multidimensional extensions have been first reported
for instance in \cite{Bulow_T_2001_j-ieee-tsp_hyp_sneasmc} and then in
\cite{Chan_W_2004_icassp_dir_hwmsap,Wedekind_J_2007_p-icspc_ste_fghdtwt}
using the notion of hypercomplex wavelets.

Numerous extension to multidimensional signals have been proposed, see for instance~\cite{Unser_M_2009_tip_mul_msarlwt,Unser_M_2009_p-icip_hig_ortswf}. They, for instance, use the Riesz transform $\Rr$, 
which is defined in the frequency domain as follows:
\begin{equation}
\widehat{\Rr}\{f\}= (\widehat{\Rr}_1\{f\},\dots,\widehat{\Rr}_N\{f\}).
\end{equation}
where
\begin{equation}
\forall n\in\{1,\,\cdots,N\},\qquad \widehat{\Rr}_n\{f\}(\boldsymbol{\omega}) = -\icomplex \frac{\omega_n}{\|\boldsymbol{\omega}\|} \widehat{f}(\boldsymbol{\omega}).
\end{equation}
Other recent extensions of multidimensional oriented wavelets are based on the notion of monogenic signal/wavelet
\cite{Felsberg_M_2002_tr_low_lipsm,Olhede_S_2009_j-ieee-tsp_mon_wt,Held_S_2010_j-ieee-tip_ste_wfrt}.
We finally mention that other methods have
been developed in order to achieve directional analytic wavelets such
as softy space projections
\cite{VanSpaendonck_R_2000_p-icip_non_rdscw,Fernandes_F_2001_p-icip_dir_silrwt,Fernandes_F_2003_tsp_new_fcwt,Fernandes_F_2004_icassp_non_rlpsodcw,Fernandes_F_2005_tip_mul_mbcwt}
or the Daubechies complex wavelets
\cite{Gagnon_L_1995_p-embs_sha_edmcsdw,Belzer_B_1995_tsp_com_lpfeic,Clonda_D_2004_sp_com_dwpsim}.
Complex wavelets have also been shown to provide robust
  image similarity measures
  \cite{Wang_Z_2005_p-icassp_tra_iiscwd,Sampat_M_2009_tip_com_wssnisi}.

\subsection{Non-Separable Directionality}
\label{sec:non-separ-direct}

\subsubsection{Non-separable Decomposition Schemes}
\label{sec:nonsep-design}

In contrast to the separable constructions detailed in
Sec.~\ref{sec:adaptedgeometry-mband} where \nDim{n} representations
are composed of \nDim{1} transforms applied separately along each
dimension (sometimes recombined, as in the dual-tree wavelet case or
in \cite{Shen_L_2006_tsp_ima_dtf}), non-separable constructions are
directly performed in \nDim{n}. Since the literature on this
topic is large, this section is focussed on a limited number of
references dealing with directional multiscale decompositions.

These works are often related to non-diagonal subsampling operators,
non-rectangular lattices (\eg quincunx grids or integer lattices)
\cite{Bamberger_R_1992_j-ieee-tsp_fil_bdditd,Smith_M_1984_p-icassp_pro_derfbtssc},
or non-separable \nDim{n} windows
\cite{Xia_X_1995_acha_f_tdnmw,Coulombe_S_1996_p-icassp_mul_walafirfd}. Complementary
standard references can be found in \cite[p. 558
sq.]{Vaidyanathan_P_1993_book_mul_sfb} or
\cite{Kovacevic_J_1992_tit_non_mprfbwbrn,Kovacevic_J_1995_tsp_non_ttdw,Antoine_J_2004_book_two_dwr}. Some of these constructions are defined using the lifting scheme, see Sec.~\ref{sec-lifting} and \ref{sec:lift-scheme-wavel} for more details.
While directional filter banks do not provide a multiscale
representation in general, $2$-band
\cite{Feauveau_J_1990_ts_ana_mfrs2,Faugere_J_1998_tsp_des_rnbwgbt,Ayache_A_2001_j-acha_som_mcnocswb}
or even $M$-band non-redundant directional discrete wavelets
\cite{Durand_S_2007_acha_m_bfndw} have been proposed. 
Non-separable
schemes are used for instance as building blocks for multiscale
geometric decompositions such as:
\begin{itemize}
\item directional filter banks in
  \cite{Nguyen_T_2007_j-ieee-tsp_cla_mdfb}, and their combination with a
  Laplacian pyramid in contourlets
  \cite{Do_M_2005_tip_con_tedmir,Cunha_A_2006_tip_non_cttda} or
  surfacelets \cite{Lu_Y_2007_tip_mul_dfbs},
\item (pseudo-) polar fast Fourier transform (FFT) \cite{Averbuch_A_2006_j-acha_fas_apft} in first generation curvelets described in Sec.~\ref{sec:adaptedgeometry-curvelet}, or the loglets in
  \cite{Knutsson_H_2005_j-spic_imp_iulsafs} that exhibit a polar
  separability.
\end{itemize}

In order to overcome the limited efficiency of the standard \nDim{2}
separable DWT for representing non-horizontally or vertically directed
edges (see Sec.~\ref{subsubsec-iso-wav}), several authors have adapted
\nDim{1} concepts for local edge representation. Reissell
\cite{Reissell_L_1996_j-graph-model-image-process_wav_mrcs} develops,
for instance, a pseudo-coiflet scheme that addresses numerically
efficient interpolation for a parametric wavelet representation of
curves.  Moreover, for digital images it would be beneficial to follow
contours on more appropriate discrete paths (see
\cite{Taubman_D_1994_j-ieee-tip_ori_asci} for an early application)
such as discrete lines
\cite{Bresenham_J_1998_j-ibm-syst-j_alg_ccdp,Rosenfeld_A_2001_j-entcs_dig_s,Daragon_X_2003_incoll_dis_f}. While
discrete lines are adapted to digital ridgelets in
\cite{Andres_E_2002_p-dgci_rid_trdl}, Velisavljevi{\'c} \etal
propose multidirectional anisotropic directionlets
\cite{Velisavljevic_V_2006_tip_dir_amdrsf},
based on skewed lattices, with directional vanishing moments along
direction with rational slopes, still relying on a simple separable
implementation. This approach is refined in
\cite{Chappelier_V_2006_tip_ori_wticd} by taking lifting steps of
\nDim{1} wavelets along an explicit orientation map defined on a
quincunx multiresolution sampling grid, and in
\cite{Chang_C_2007_tip_dir_adwtic} with a more efficient
representation for sharp features.  A combination of
\nDim{2} filter banks and \nDim{1} directional filter bank is devised in
\cite{Tanaka_Y_2009_tip_mul_irc2d1ddfb,Tanaka_Y_2010_j-ieee-tip_ada_dwtbdp}.  Similar ideas have been
recently applied to edge detection in
\cite{Zhang_Z_2009_j-comput-math-appl_edg_dabdwt}.  In
\cite{Krommweh_J_2009_j-acha_dir_hwft}, non-adaptive directional
wavelet frames are constructed with Haar wavelets and a finite
collection of ``shear'' matrices. Krommweh also proposes tetrolets,
an adaptive variation (akin to digital wedgelets) of Haar-like
wavelets on compact tetrominoes (geometric shapes composed of four
squares, connected orthogonally, see \cite{Golomb_S_1994_book_pol}).
These last constructions may further sparkle the growing interest of
the association of multiscale analysis and discrete geometry
\cite{Said_M_2009_p-dgci_mul_dg}.

\subsubsection{Steerable Filters}
\label{sec:steerability}

Steerable filters \cite{Freeman_W_1990_p-iccv_ste_feviawd,Freeman_W_1991_tpami_des_usf,Freeman_W_1992_phd_ste_flais} were developed in order to achieve more precise
feature detectors adapted to image edge junctions (often termed ``X'',
``T'' and ``L'' junctions).
Their construction allows one to compute multiscale derivatives at any orientation (steerability) from a linear
combination of a small number of fixed filters.  In
\cite{Freeman_W_1991_tpami_des_usf}, the construction starts from a
bidimensional Gaussian $G(\bs x)=\exp(-\inv{2}\|\bs x\|^2)$ for $\bs
x=(x_1,x_2)$ with associated base (differential) filters
$\GSteer^{0}(\bs x) = \tfrac{\partial}{\partial x_1} G(\bs x)$ and
$\GSteer^{\pi/2}(\bs x)=\tfrac{\partial}{\partial x_2} G(\bs x)$.

\begin{figure}[htb!]
  \centering
  \begin{tabular}{@{}c@{\hspace{1mm}}c@{\hspace{1mm}}c@{\hspace{1mm}}c@{}}
   \includegraphics[height=.23\linewidth]{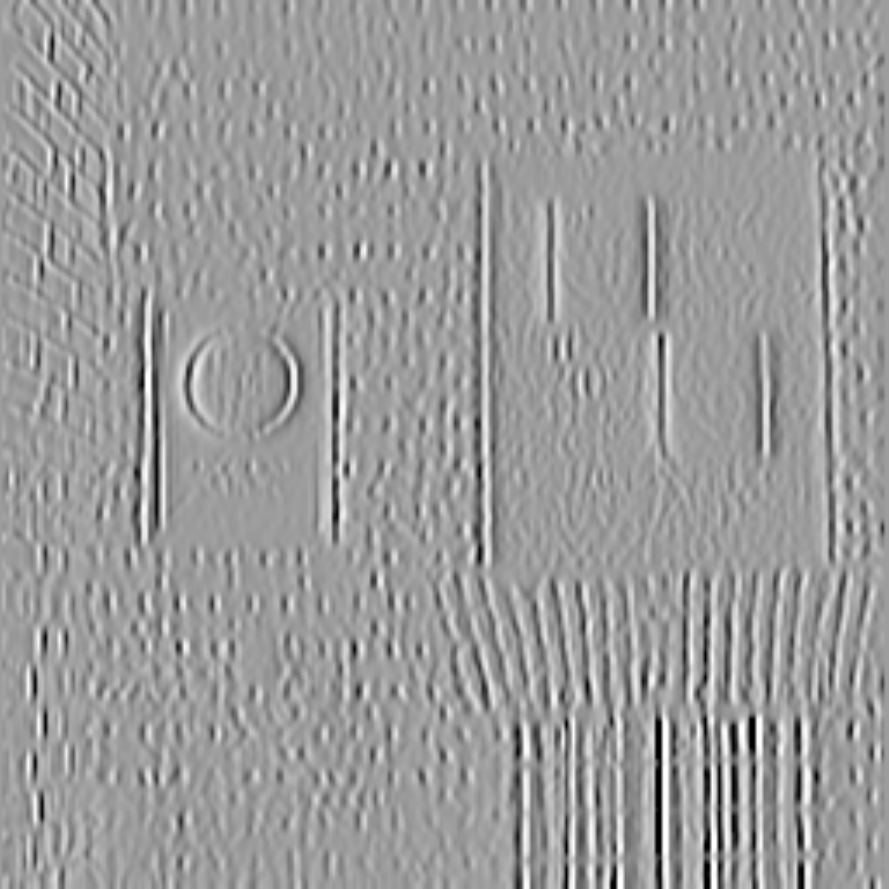}&
   \includegraphics[height=.23\linewidth]{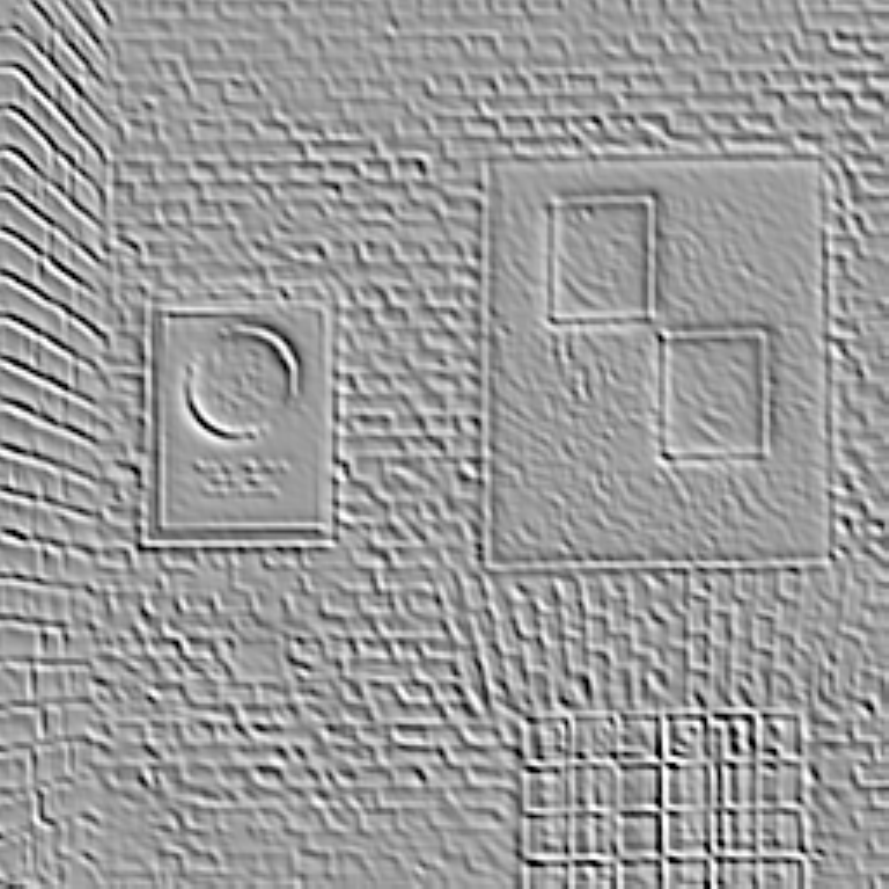}&
   \includegraphics[height=.23\linewidth]{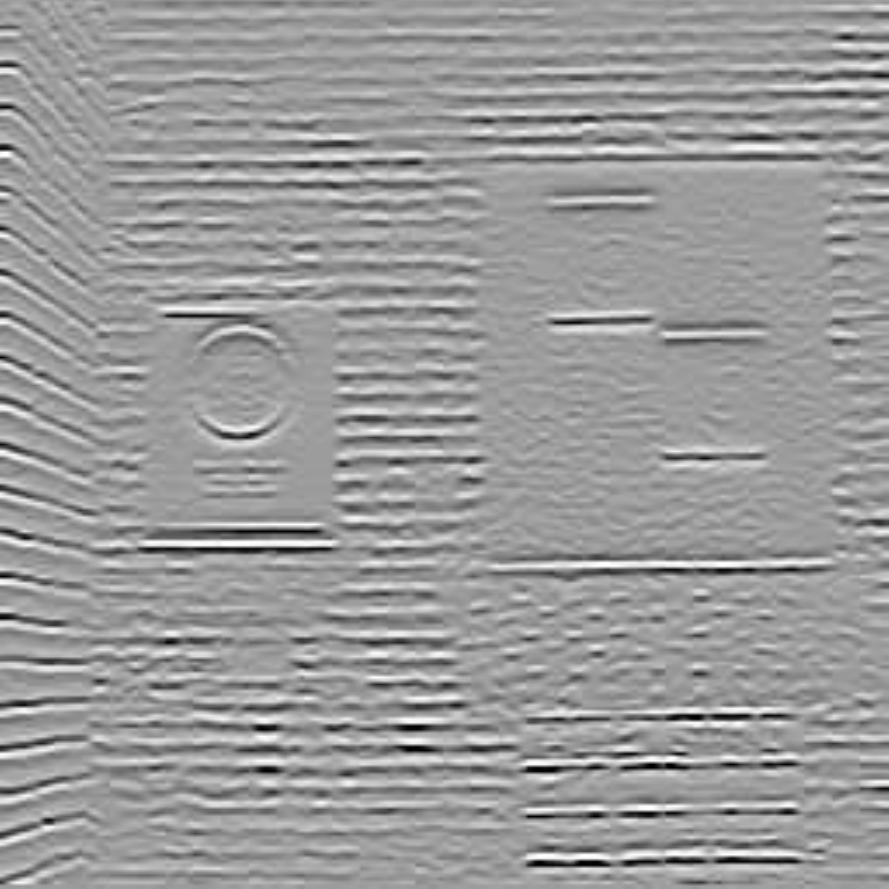}&
   \includegraphics[height=.23\linewidth]{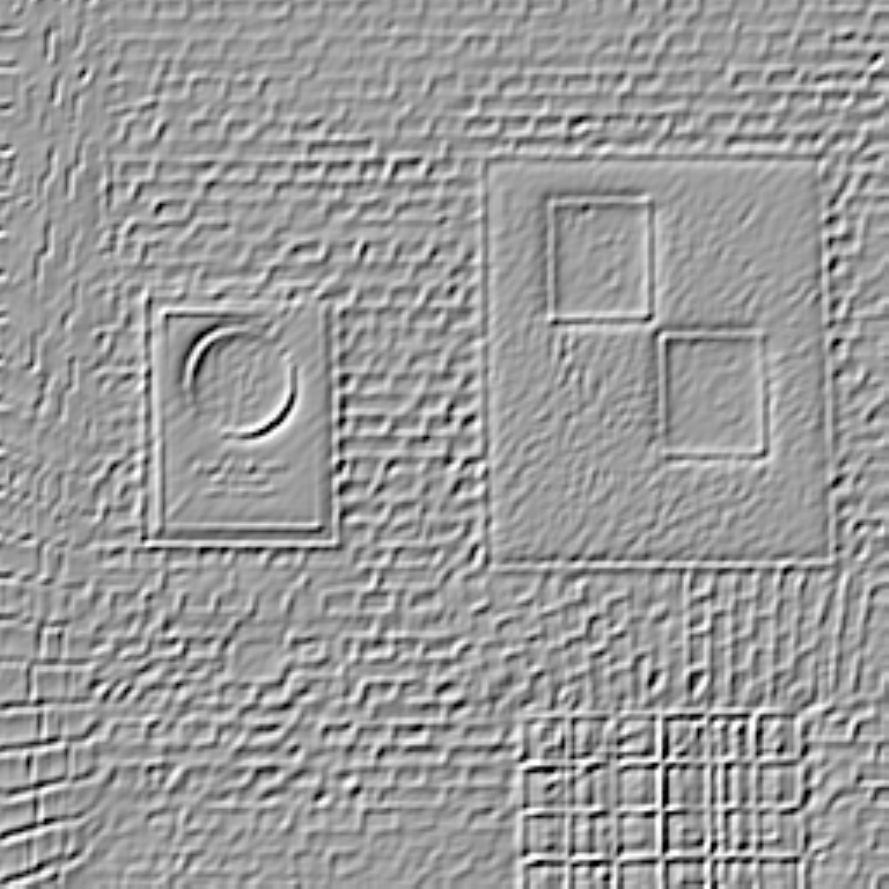}\\
   \includegraphics[height=.12\linewidth]{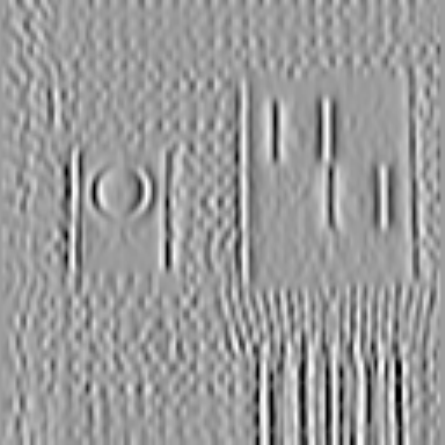}&
   \includegraphics[height=.12\linewidth]{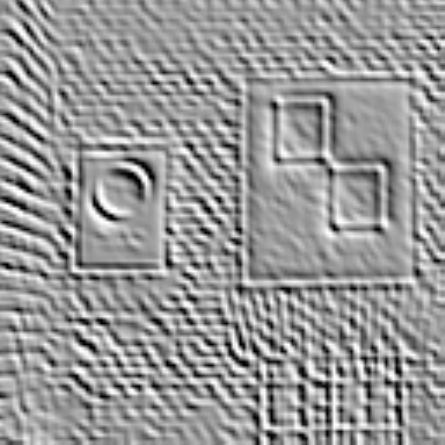}&
   \includegraphics[height=.12\linewidth]{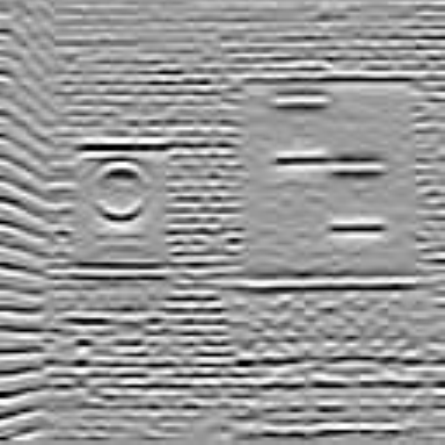}&
   \includegraphics[height=.12\linewidth]{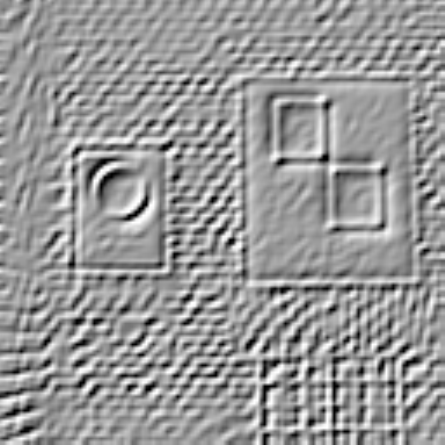}
   \end{tabular}
  \caption{Steerable pyramid decomposition \imageHRP, over two scales, with four orientations.}
  \label{fig:fig_SzegedPyramidSteerable-bis}
\end{figure}

\begin{figure}[htb!]
  \centering
  \includegraphics[height=.2\linewidth]{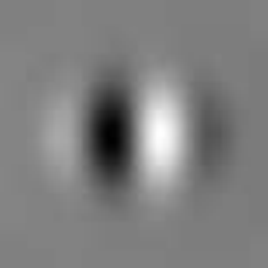}
  \includegraphics[height=.2\linewidth]{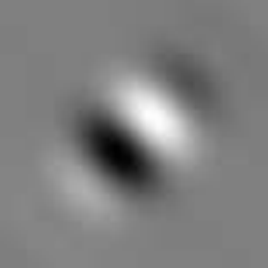}
  \includegraphics[height=.2\linewidth]{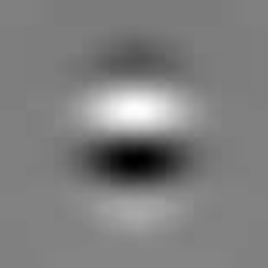}
  \includegraphics[height=.2\linewidth]{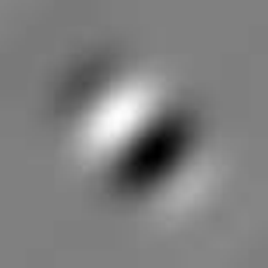}
  \caption{
  Example of steerable pyramid atoms with four orientations.
  }
  \label{fig:fig_SzegedPyramidSteerable-atoms}
\end{figure}

From the properties of the directional derivative, filters ``steered'' at angle $\theta \in [0,2\pi)$ are then built from
\begin{equation}
\GSteer^{\theta}(\bs x)\ =\ \cos(\theta)\,\GSteer^{0}(\bs x)\ +\ \sin(\theta)\,\GSteer^{\pi/2}(\bs x).
\end{equation}
where $\cos(\theta)$ and $\sin(\theta)$ may be interpreted as
interpolators. Since the convolution is linear, the resulting steered
decomposition arises from a combination of images that underwent
$\GSteer^{0}$ or $\GSteer^{\pi/2}$ filters.  A larger class of
asymmetric oriented filters is proposed in
\cite{Simoncelli_E_1996_tip_ste_wfloa}.  Their angular parts are
derived from even and odd functions:
\begin{equation}
\forall \varphi \in [0,2\pi), \qquad h_e(\varphi)=\sum_{n=1}^{N}w_n
\cos(n\varphi) \qquad \mbox{and} \qquad h_o(\varphi)=\sum_{n=1}^{N}w_n
\sin(n\varphi),
\end{equation}
which form  
Hilbert transform pairs (see Sec. \ref{sec:adaptedgeometry-complex}), 
unlike the resulting spatial filters. An angle  $\theta$ rotation is obtained through:
\begin{equation}
h_e(\varphi-\theta)=\bs{k}_e(\theta)^T \bs{f}(\varphi) \qquad
\mbox{and} \qquad h_o(\varphi-\theta)=\bs{k}_o(\theta)^T
\bs{f}(\varphi),
\end{equation}
where  $\bs{k}_e(\theta)$ and  $\bs{k}_o(\theta)$ are interpolating vectors and 
$\bs{f}(\varphi)$ is a weighted Fourier vector, namely:
\begin{align*}
\bs k_e(\theta)&=\big[\,\phantom{-}\cos\theta, \sin\theta, \phantom{-}\cos (2\theta), \sin
(2\theta),\ \cdots\ ,\phantom{-}\cos (N\theta), \sin (N\theta)\big]^T,\\
\bs k_o(\theta)&=\big[-\sin\theta, \cos\theta, -\sin (2\theta), \cos
(2\theta),\ \cdots\ ,-\sin (N\theta), \cos (N\theta)\big]^T,\\
\bs f(\varphi)&=\big[\,w_1\cos\varphi, w_1\sin\varphi, w_2\cos (2\varphi), w_2\sin
(2\varphi), \cdots , w_N\cos (N\varphi), w_N\sin (N\varphi)\big]^T\!\!.
\end{align*}
If we set $\theta=\theta_n=2 \pi n / N$ for $1\leq n \leq N$, filters
$h_e(\cdot-\theta)$ and $h_o(\cdot-\theta)$ may be rewritten as a
linear combination of $h_e(\cdot-\theta_n)$ and $h_o(\cdot-\theta_n),
\, 1\leq n \leq N$. An example of decomposition with four orientations
and two scales is represented in
Fig.~\ref{fig:fig_SzegedPyramidSteerable-bis}, with corresponding
projection atoms in Fig.~\ref{fig:fig_SzegedPyramidSteerable-atoms}.
Steerable filters may be combined with discrete wavelets to improve
their radial properties 
\cite{Bharath_A_2005_tip_ste_cwcaid,Shi_X_2006_spl_rot_iosfb}.

\subsubsection{Directional Wavelets and Frames}
\label{sec:direct-wavel-fram}

In Sec.~\ref{sec:introtools-contss-wavelet}, the two-dimensional
Continuous Wavelet Transform (\nDim{2} CWT) was defined as a
straightforward extension of the \nDim{1} CWT using isotropic
wavelets.  It is however possible to make use of more complicated group actions to drive the 
CWT parameterization in the plane, such as rotations or the \emph{similitude} group SIM$(2)$, see~\cite{Antoine_J_2004_book_two_dwr}.

Consequently, given a mother function $\atom\in\LL^2(\Rbb^2)$ that is well
localized and oriented, we write 
$$
\atom_{(\bs b,a,\theta)}(\bs x)\ =\ \tinv{a}\,\atom(\tinv{a}\,R_\theta^{-1} \big(\bs x
- \bs b)\big),
$$
where $R_\theta$ stands for the $2\times 2$ rotation matrix. For a function
$f\in \LL^2(\Rbb^2)$, the 2-D CWT (non-isotropic) is thus 
$$
W_f(\bs b,a,\theta)\ =\ \scp{\atom_{(\bs b,a,\theta)}}{f}.
$$
If the wavelet is admissible, \ie if $c_\atom = (2\pi)^2 \int_{\Rbb^2}
|\hat\atom(\bs \omega)|^2/\|\bs \omega\|^2\ \ud^2\bs \omega< \infty$, then, the CWT may be
inverted through
$$
f(\bs x)\ =\ c^{-1}_\atom\int_0^\infty\!\!\!\tfrac{\ud a}{a^3}\int_0^{2\pi}\!\!\!\ud\theta\int_{\Rbb^2}\!\!\!\ud^2 \bs
b\quad W_f(\bs b,a,\theta)\ \atom_{(\bs b,a,\theta)}(\bs x),$$
the equality being valid almost everywhere on $\Rbb^2$.

The \emph{selectivity power} of the wavelet, that is, its ability to
distinguish two close orientations in an image, may be measured in the
Fourier domain. Typically, a good directional wavelet is thus a
function whose Fourier transform is essentially or exactly contained
in a cone with apex on the origin: the narrower the cone, the more
selective the wavelet transform using that wavelet
\cite{Antoine_J_2004_book_two_dwr}.

Practically, it is not satisfactory to manipulate a continuum of
wavelets parameterized by continuous parameters. The question is
therefore to know if it is possible to decompose and reconstruct an
image from a discretized set of parameters, \ie on the family
$\mathcal{G}=\{\atom_{(\bs b,a,\theta)}: \bs b\in \mathcal{P}, a\in
\mathcal{A}, \theta\in\Theta\}$ with $\mathcal{P}\subset \Rbb^2$,
$\mathcal{A}\subset \Rbb^*_+$ and $\theta\subset[0,2\pi)$ all discrete
(countable) sets.  As
explained in Sec.~\ref{sec:introtools-representation}, this
question amounts to ask when $\mathcal{G}$ is a frame of $\LL^2(\Rbb^2)$.

\begin{figure}[ht!]
  \centering
  \subfigure[\label{fig:morlet2d-spat}]{
    \raisebox{.42cm}{\includegraphics[width=5cm,keepaspectratio]{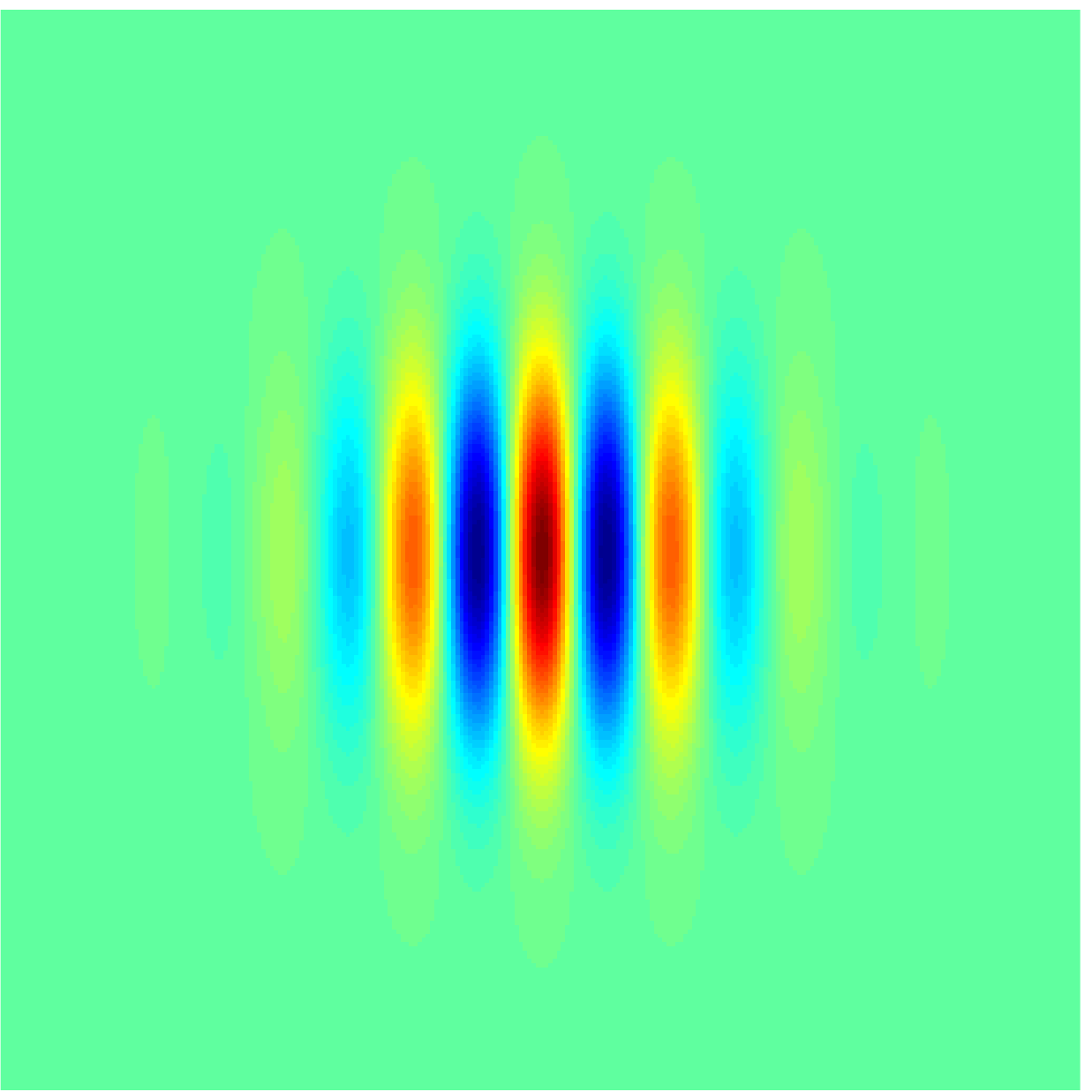}}
  }
  \subfigure[\label{fig:morlet2d-freq}]{
    \includegraphics[width=5.5cm,keepaspectratio]{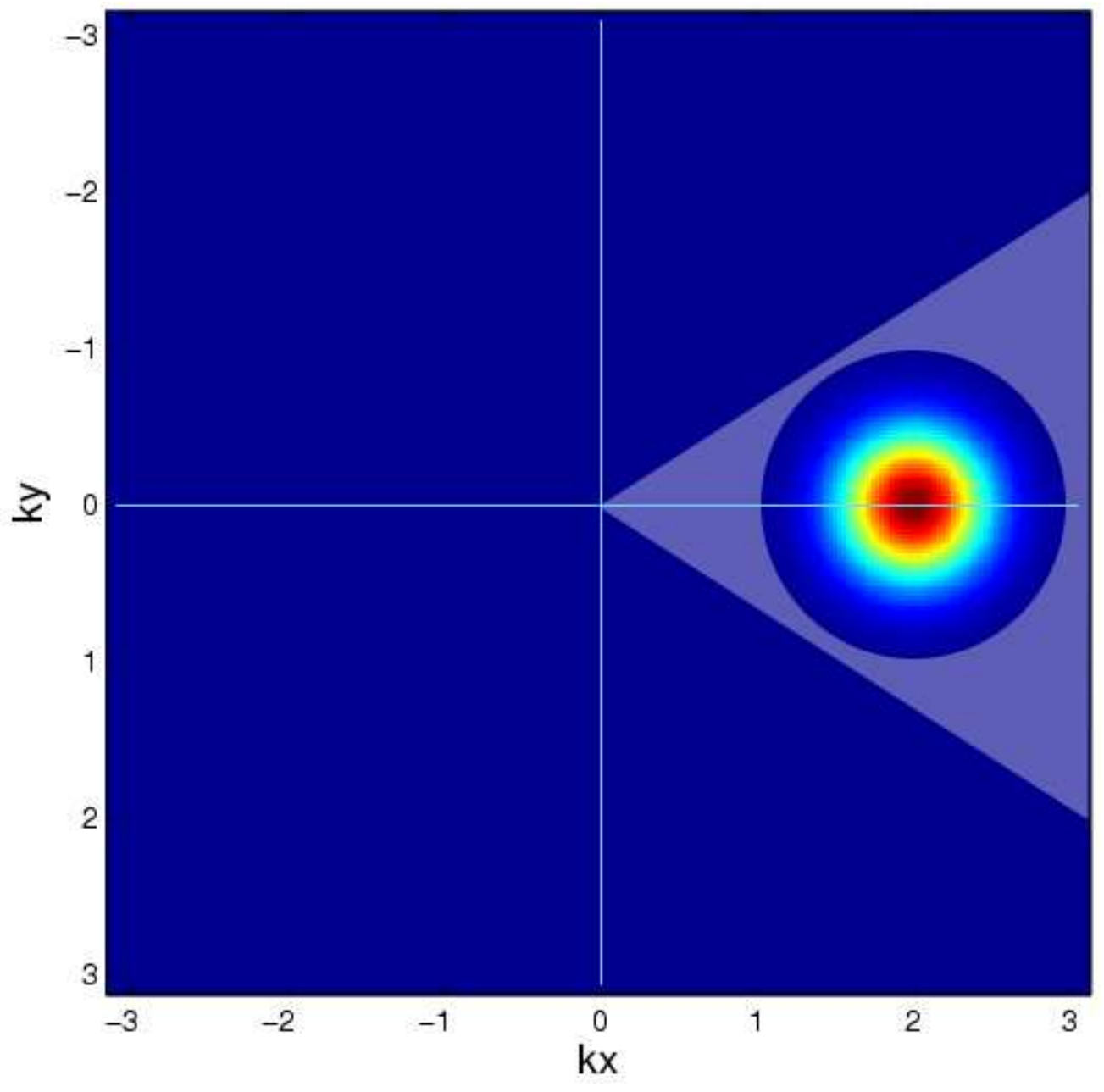}
  }
  \caption{The Morlet Wavelet. (a) Spatial representation (real
    part). (b) Fourier representation. Supporting cone and frequency
    axes  are drawn for illustration.}
  \label{fig:morlet2d}
\end{figure}

Such frames have been built for the Morlet (or Gabor) wavelet \cite{Lee_T_1996_j-ieee-tpami_ima_r2dgw,Nestares_O_1998_j-elec-im_eff_sdimirbgf}:
$$
\atom(\bs x) = G_{\sigma_0} (\bs x) \ e^{\icomplex \bs\omega_0\cdot\bs
x}
= e^{- \|\bs x\|^2/2\sigma_0^2}\ e^{\icomplex \bs\omega_0\,\bs x}, \quad
  \hat{\atom}(\bs \omega) \propto G_{1/\sigma_0} (\bs \omega - \bs
  \omega_0) = e^{- \sigma_0^2\|\bs \omega - \bs \omega_0\|^2 /2},
$$ 
where $\bs \omega_0\in\Rbb^2$ defines the cone axis and $\sigma_0>0$
is related to the cone aperture, as represented in Fig. \ref{fig:morlet2d}.  Notice that approximate quadrature
filters exist to accelerate the computation of the wavelet
coefficients \cite{Vandergheynst_P_2002_j-ieee-tip_dir_dwtda}. The Conic (or
Cauchy) wavelet, whose spectral support is exactly contained into a cone, can also be
used in order to define a frame
\cite{Antoine_J_1999_j-acha_dir_wrcwsdp}.

Finally, a multiresolution structure can also be put on the angular
dependency of the conic wavelets in the frequency domain to define
\emph{multiselective wavelets}
\cite{Jacques_L_2007_j-ijwmip_mul_pdiwaas}. This generates a redundant
basis that may represent jointly a large spectrum of features ranging
from highly directional ones (\eg edges) to isotropic elements (\eg
spots, corners) and including intermediate directional structures such
as textures.

\subsection{Directionality in Anisotropic Scaling}
\label{sec:direct-anis-scal}

\subsubsection{Ridgelets}
\label{sec:adaptedgeometry-ridgelet}

\begin{figure}[htbp]
  \centering
  \includegraphics[height=8cm]{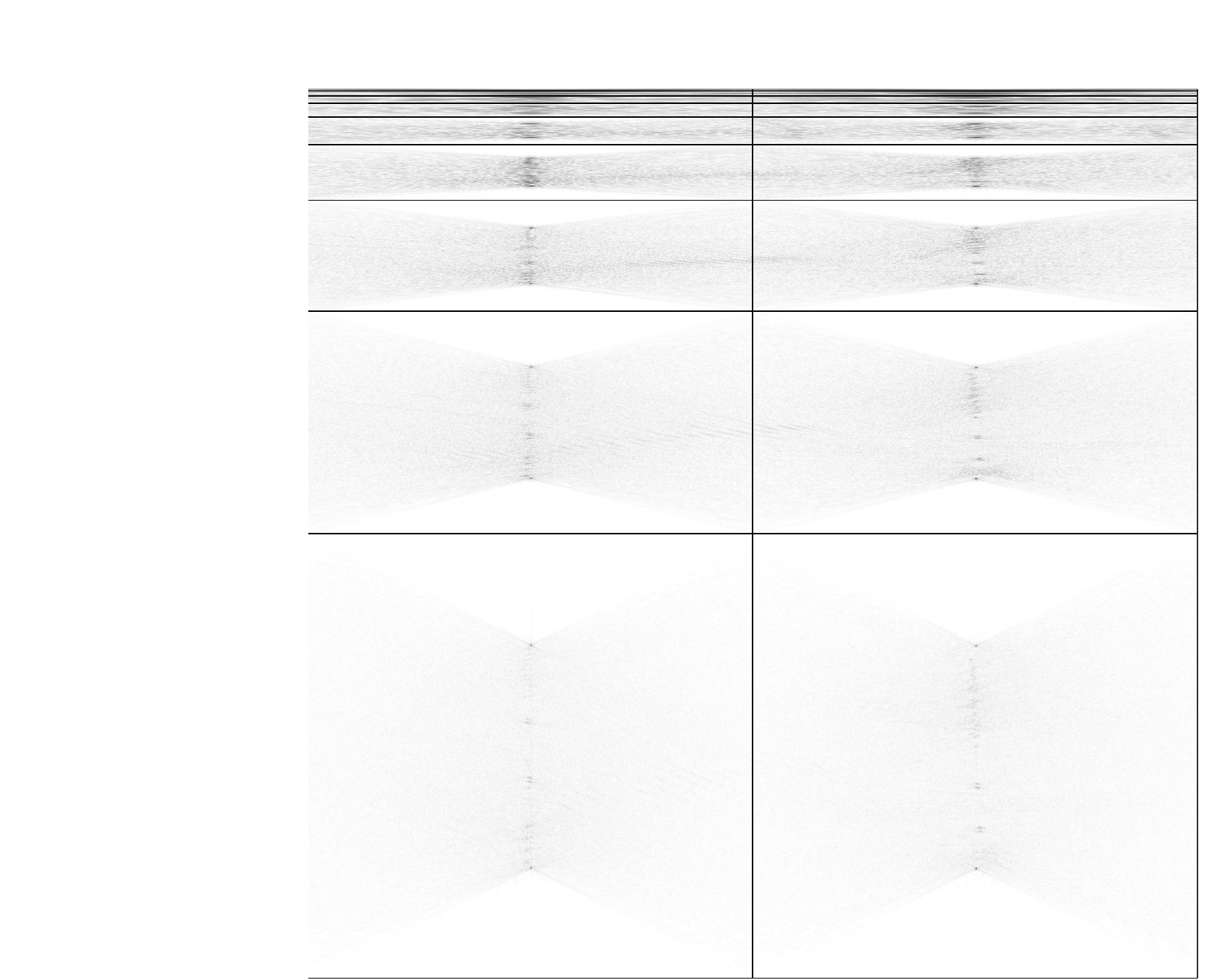}
   \caption{Ridgelet decomposition (square root scale) \imageHRP.}
  \label{fig:fig_SzegedRidgelet}
\end{figure}

\begin{figure}[htb!]
  \centering
  \subfigure[\label{fig:fig_SzegedRidgelet-proj}]{
    	{\includegraphics[height=4cm,keepaspectratio]{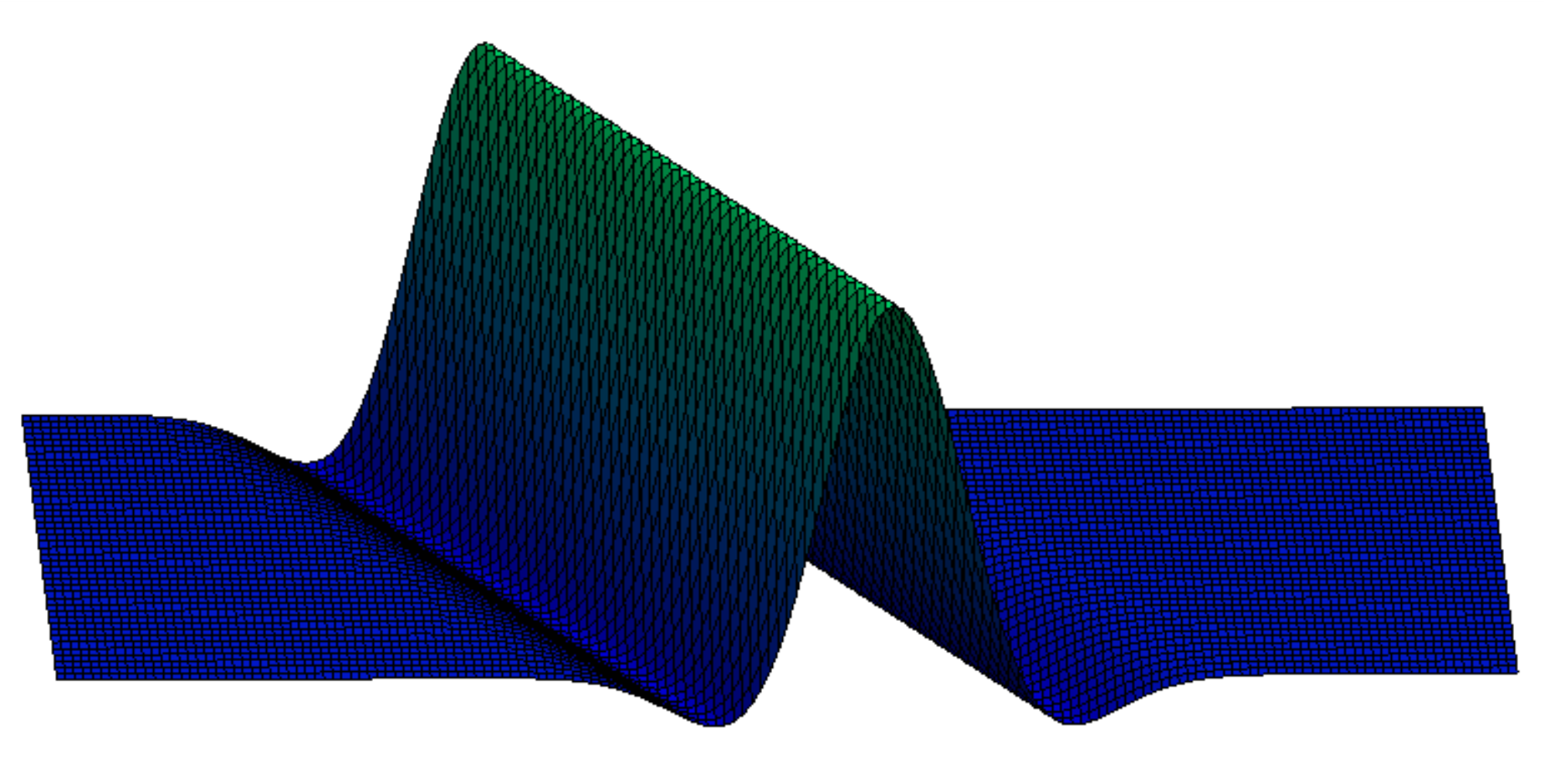}}}\\
  \subfigure[\label{fig:fig_SzegedRidgelet-freq}]{
    \includegraphics[height=8cm,keepaspectratio]{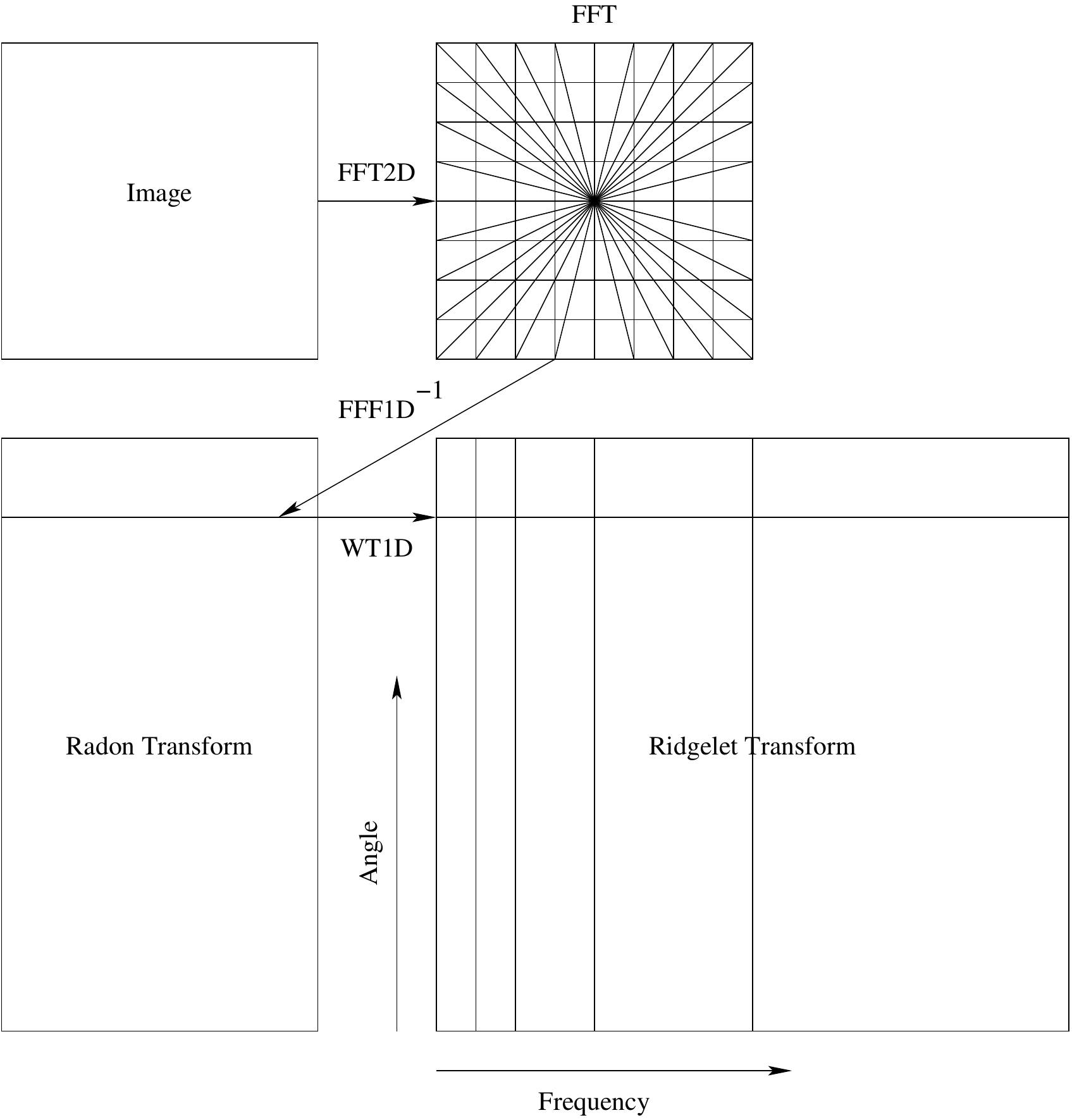}}
  \caption{The Ridgelet transform. (a) Example of atoms. (b) Synthetic implementation description.}
  \label{fig:fig_SzegedRidgelet-projfreq}
\end{figure}

Ridgelets 
\cite{Candes_E_1999_phil-trans-r-soc-lond_rid_khdi,Donoho_D_1999_j-pnas_tig_fkprprosddsrn} and wavelet X-ray transforms  \cite{Zuidwijk_R_2000_j-siam-math-anal_dir_tswa} appear 
as a combination of a \nDim{1} wavelet transform and the Radon
transform \cite{Deans_S_1983_book_rad_tsa}. They are designed for 
efficient representation of discontinuities over straight lines. A
bivariate ridgelet transform is constant along parameterized lines
$x_1 \cos(\theta) + x_2 \sin(\theta)=b$ and defined for $a>0$, $b\in
\RR$ and $\theta \in [0,2\pi)$, by
\begin{equation}
\forall \xx=(x_1,x_2)\in \RR^2, \qquad
\psi_{(\bs b,a,\theta)}(\xx)=a^{-1/2} \, \psi((x_1 \cos(\theta) +
x_2 \sin(\theta) - b)/a).
\end{equation}
Ridgelet coefficients for the image  $f$ are given by
\begin{align}
\mathcal{R}_f(b,a,\theta)
&=\int\psi_{(\bs b,a,\theta)}(\xx)\,f(\xx)
\,\ud^2\xx \nonumber\\
&=\int \mathfrak{R}_f(\theta,t)\,a^{-1/2}\psi((t-b)/a)\,\ud t,
\end{align}
where  $\mathfrak{R}_f(\theta,t)$ represents the Radon transform of  $f$ defined by:
\begin{equation}
\mathfrak{R}_f(\theta,t)=\int\int f(x_1,x_2)\, \delta(x_1 \cos(\theta) + x_2
\sin(\theta)-t)\,\ud x_1 \ud x_2,
\end{equation}
with $\delta$ denoting the Dirac distribution.  The ridgelet transform
may be interpreted as a \nDim{1} wavelet transform of Radon slices
where the angle $\theta$ is constant and $t$ varies. Several
implementations and variations exist in order to overcome the issues
raised by the Radon transform discretization, such as the finite
ridgelet transform \cite{Do_M_2003_tip_fin_rtir}, the approximate
digital ridgelet transform \cite{Starck_J_2002_tip_cur_tid} or the
discrete analytical ridgelet transform
\cite{Helbert_D_2006_tip_3d_dart}.  Their multiscale implementation
\cite{Candes_E_1999_curves-surfaces_cur_senroe} is the basis for the
first generation curvelets described in
Sec.~\ref{sec:adaptedgeometry-curvelet}.  A ridgelet decomposition\footnote{BeamLab toolbox: \url{http://www-stat.stanford.edu/~beamlab/}.} \cite{Donoho_D_2003_incoll_dig_rtbtrf}
\imageHRP is given in Fig.~\ref{fig:fig_SzegedRidgelet}, with a
typical atom along with  a synthetic description of its implementation in
Fig.~\ref{fig:fig_SzegedRidgelet-projfreq}. 

\subsubsection{Curvelets}
\label{sec:adaptedgeometry-curvelet}

The curvelet representation, introduced by Cand\`es and Donoho
\cite{Candes_E_1999_curves-surfaces_cur_senroe,Candes_E_2003_j-comm-pure-appl-math_new_tfcoropc2s},
improves the approximation of cartoon images with $\Cdeux$ edges with
respect to wavelets. We review here the second generation of curvelets, as introduced in \cite{Candes_E_2003_j-comm-pure-appl-math_new_tfcoropc2s}.

\paragraph{Continuous Curvelet Transform}

A curvelet atom, with scale $s$, orientation $\th \in [0,\pi)$,
position $\yy \in [0,1]^2$ is defined as
\eql{\label{eq-curv-continuous} 
	\atom_{s,\yy,\th}(\xx) = \atom_{s}(
  R_{\th}^{-1}( \xx-\yy ) ) } 
where $\atom_{s}(\xx) \approx
s^{-3/4}\,\atom(s^{-1/2} x_1, s^{-1} x_2)$ is approximately a
parabolic stretch of a curvelet function $\atom$ with vanishing
moments in the vertical direction. At scale $s$, a curvelet atom is thus a needle oriented in the direction $\theta$ whose envelope is a specified ridge of effective length $s^{1/2}$ and width $s$, and which displays an oscillatory behavior transverse to the ridge.

\begin{figure}[ht!]
  \centering
  \includegraphics[width=.7\linewidth]{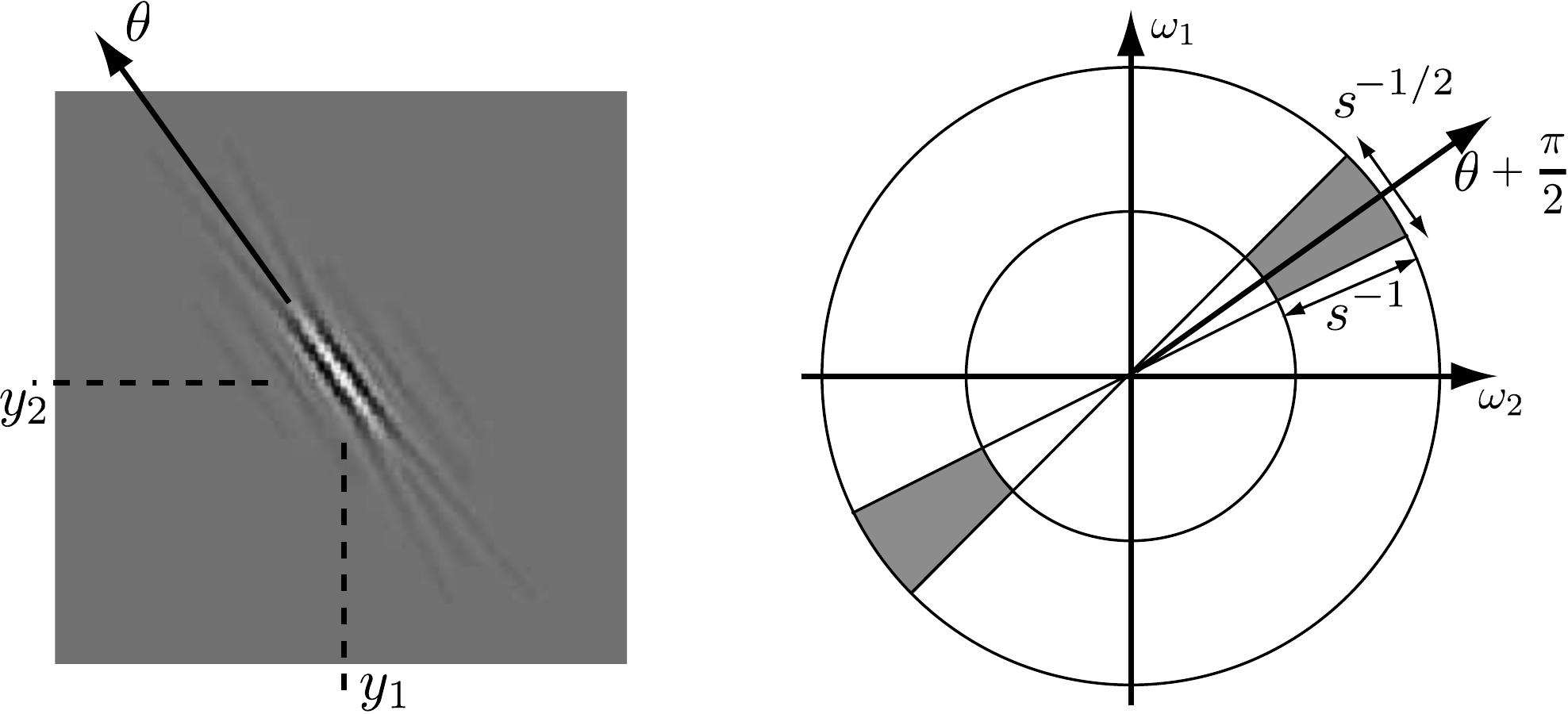}
  \caption{Left: Example of a curvelet $\atom(\xx, s, \yy,\th)$. Right: the
    frequency support of $\hat \atom(\om, s, \yy,\th)$ is a wedge. }
  \label{fig-curvelet-sample}
\end{figure}

A curvelet atom thus benefits from
a parabolic scaling property $width=length^2$ that is a major
departure from oriented wavelets.  Fig.~\ref{fig-curvelet-sample}
presents an example of a curvelet atom, together with its Fourier
transform, for the second generation of curvelets. The resulting curvelet Fourier tiling resembles that of the
Cortex transform \cite{Watson_A_1987_j-comput-vision-graph-image-process_cor_trcsni}.

The continuous curvelet transform computes the set of inner products
$\dotp{\atom_{s, \yy,\th}(\cdot)}{f}$ for all possible $(s,
\yy,\th)$. A careful design of $\atom_{s}$
\cite{Candes_E_2003_j-comm-pure-appl-math_new_tfcoropc2s} enables a conservation of energy and a simple reconstruction formula.
The decay of the curvelet transform as $s$ decreases allows one to
detect the position and orientation of contours \cite{Candes_E_2003_j-acha_con_ct1rws}.

\paragraph{Curvelet Frame}

The continuous curvelet representation is sampled in order to obtain a
curvelet frame $\Basis = \{\atom_{\m}\}_{\m}$,
\cite{Candes_E_2003_j-comm-pure-appl-math_new_tfcoropc2s}, see also
\cite{Candes_E_2003_j-acha_con_ct2df} for the description of a complex
curvelet tight frame.

A curvelet atom, with scale $2^j$, orientation $\th_\ell \in [0,\pi)$, position $\xx_n \in [0,1]^2$ is defined from the continuous atom \eqref{eq-curv-continuous}
\eq{
	\atom_{\m}(\xx) = \atom_{2^j, \th_\ell, \xx_n}(  \xx  )
	\qwhereq
	m=(j,n,\ell)
} 
where the sampling locations are 
\eq{
	\th_\ell = \ell \pi 2^{\lfloor j/2 \rfloor-1} \in [0,\pi)
	\qandq
	\xx_n = R_{\th_\ell} ( 2^{j/2} n_1, 2^{j} n_2 ) \in [0,1]^2.
}
The curvelet parameters are sampled using an increasing number of orientations at finer scales. This sampling is the key ingredient to ensure the tight frame property \cite{Candes_E_2003_j-comm-pure-appl-math_new_tfcoropc2s}, which provides a simple reconstruction formula. 

\begin{figure}[htbp]
  \centering
  \includegraphics[height=.6\linewidth]{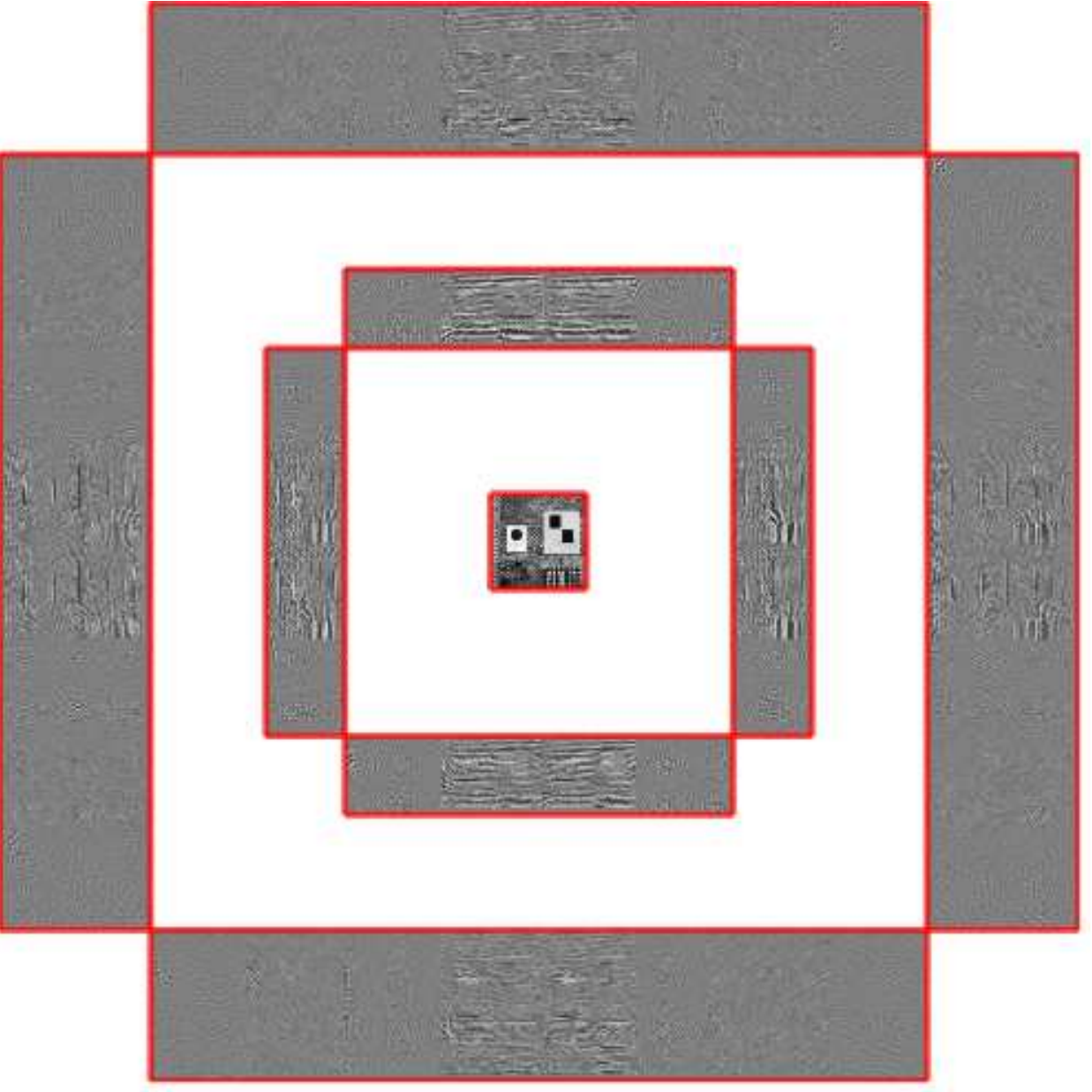}    \caption{Curvelet decomposition \imageHRP. The layout of the coefficients follows the frequency localization of curvelet atoms.}
  \label{fig:fig_SzegedCurvelet}
\end{figure}

A fast discrete curvelet transform computes the set of inner products $\{ \dotp{\atom_{\m}}{f} \}_{\m}$ in $O(N \log(N))$ operations for an image with $N$ pixels, see \cite{Candes_E_2006_siam-mms_fas_dct}.
The coronae and rotations of the continuous settings are replaced by their discrete Cartesian counterparts, i.e. concentric squares and shears. Figure \ref{fig:fig_SzegedCurvelet} shows an example of curvelets decomposition\footnote{The Curvelab toolbox has been used, see \url{http://www.curvelet.org/}.}.

Cand{\`e}s and Donoho prove \cite{Candes_E_2004_j-comm-pure-appl-math_new_tfcoropc2s} that the
curvelet non-linear approximation $f_M = \ThreshH_T(f, \Basis)$, where
$\ThreshH_T$ is defined in \eqref{eq-thresholding}, ensures an
approximation error decay $\norm{f-f_M}^2 = O(M^{-2} \log^3(M))$ for a
$\Cdeux$ regular image outside $\Cdeux$ regular edge curves. This is a
significant improvement over the $O(M^{-1})$ error decay of a wavelet
approximation described in Sec.~\ref{subsubsec-iso-wav}, and is
achieved with a fast $O(N \log(N))$ algorithm for discrete images. 
This asymptotic error decay is optimal (up to logarithmic factor) for the class of
images that are $\Cdeux$ regular outside $\Cdeux$ regular edge curves, see \cite{Candes_E_2004_j-comm-pure-appl-math_new_tfcoropc2s}. Monogenic curvelets are proposed in  \cite{Storath_M_2011_j-siam-j-imaging-sci_dir_mapdmct} to obtain additional advantages over monogenic wavelets, described in Section  \ref{sec:adaptedgeometry-complex}.

Shearlet atoms \cite{Guo_K_2007_j-siam-math-anal_opt_smrs,Kittipoom_P_2010_j-four-anal-appl_irr_sfgap} are built similarly to curvelets, but they replace, in their continuous formulation,  rotation and anisotropic stretch with anisotropic shears. The discrete shearlet transform~\cite{Kutyniok_G_2007_j-wavelet-theory-appl_con_risf,Lim_W_2010_j-ieee-tip_dis_stndtcssf} is thus implemented similarly to the discret curvelet transform~\cite{Candes_E_2006_siam-mms_fas_dct} using discrete shears\footnote{An implementation is available at \url{http://www.shearlab.org}}. It provides the same approximation properties as curvelets, albeit with a different directional sensitivity (\eg the number of orientations doubles at each scale). Recently a type-I ripplet transform \cite{Xu_J_2010_j-vcir_rip_ntip} has been proposed as an extension to curvelets with alternative scaling laws.

\subsubsection{Contourlets}
\begin{figure}[htbp]
  \centering
  \includegraphics[height=7cm]{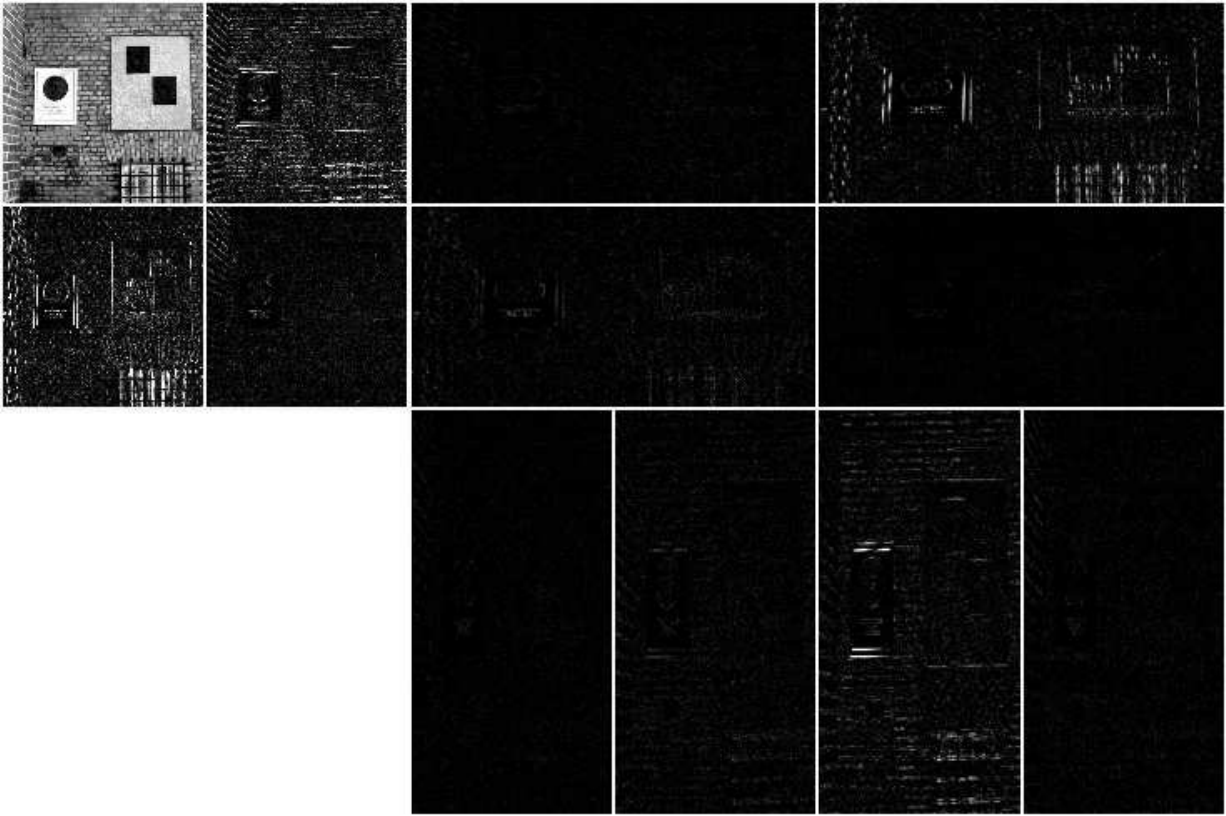}    \caption{Contourlet decomposition \imageHRP.}
  \label{fig:fig_SzegedContourlet}
\end{figure}

\begin{figure}[htb!]
  \centering
  \subfigure[\label{fig:fig_SzegedContourlet-proj1}]{
    {\includegraphics[height=4cm,keepaspectratio]{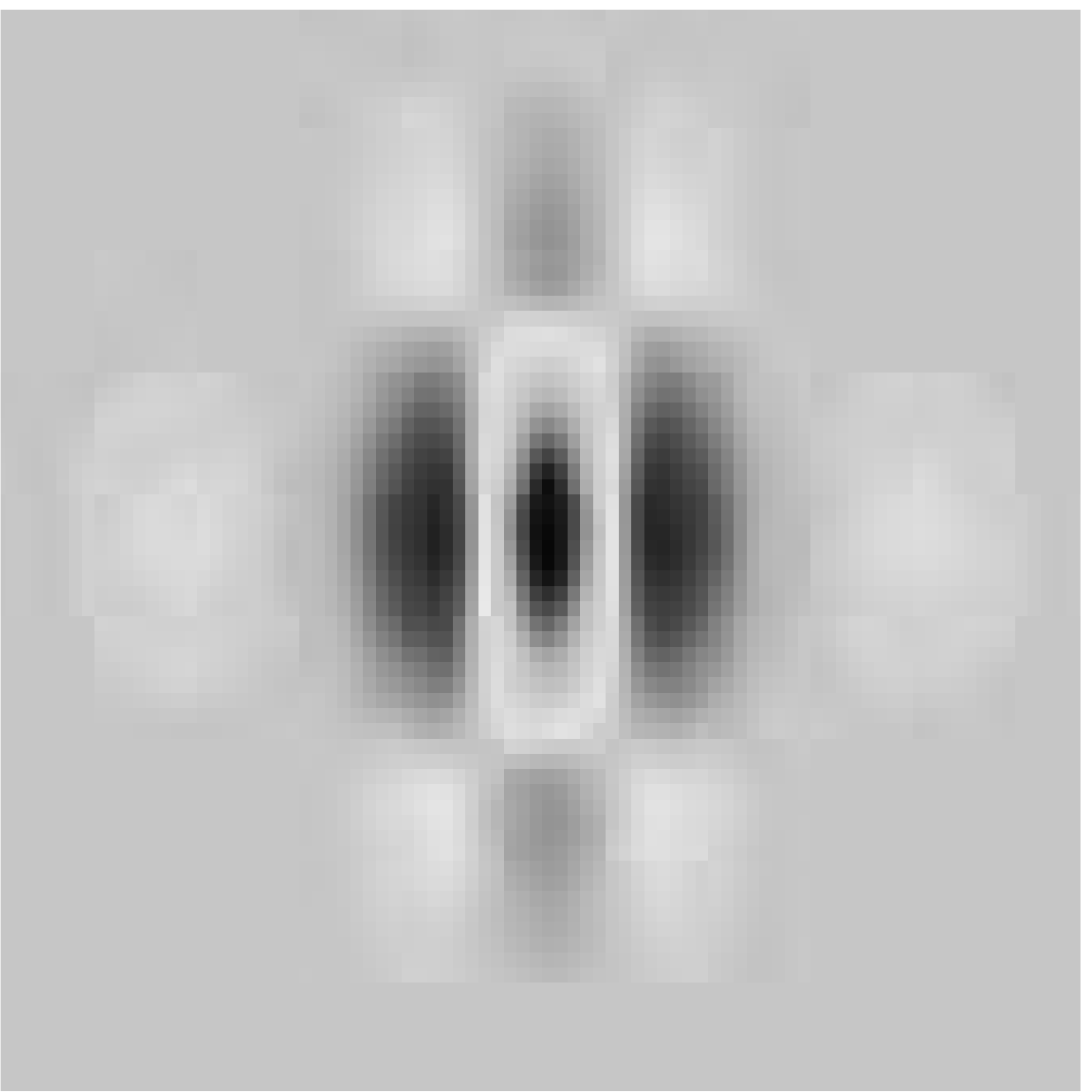}}}
  \subfigure[\label{fig:fig_SzegedContourlet-proj2}]{
    {\includegraphics[height=4cm,keepaspectratio]{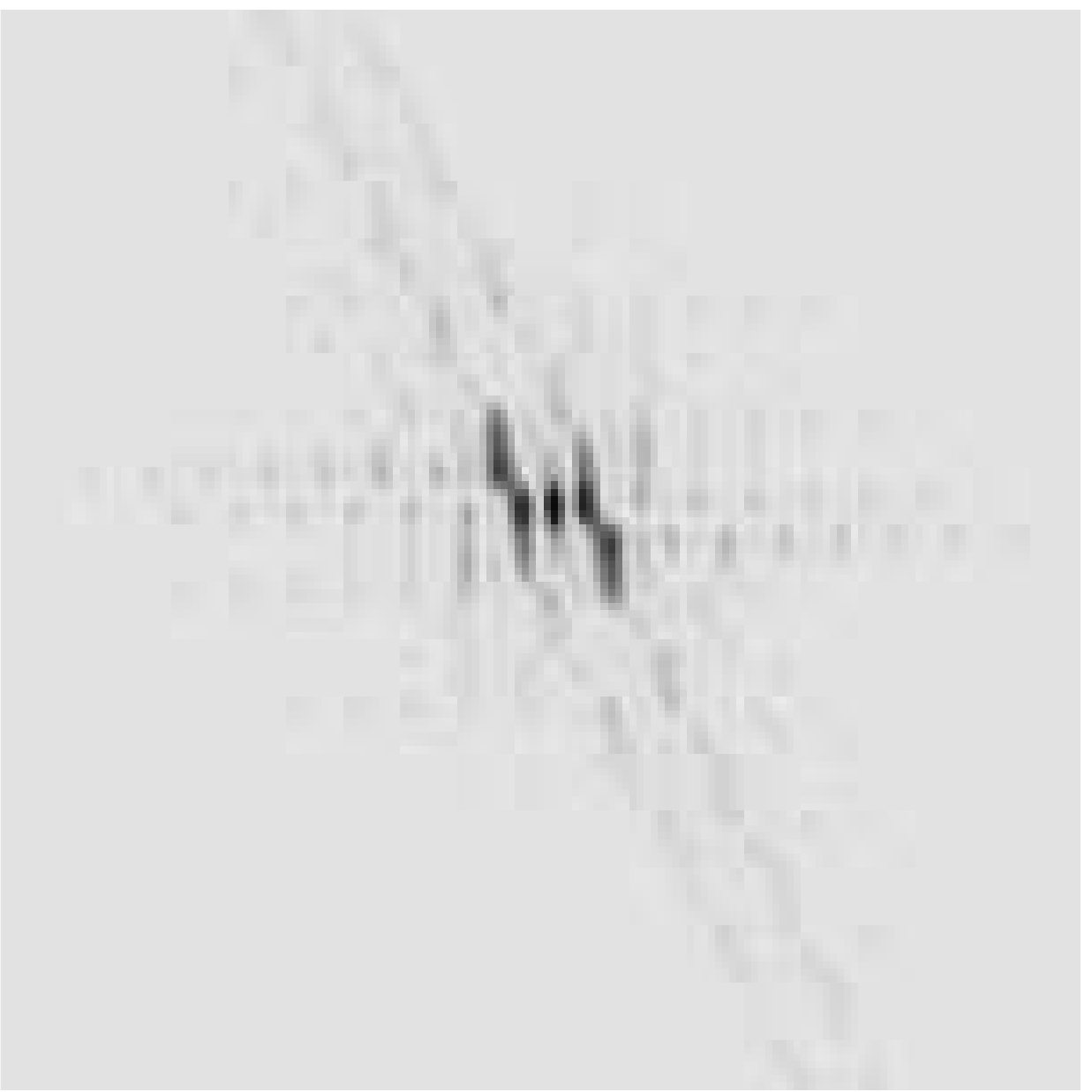}}}
  \subfigure[\label{fig:fig_FreqPartitionContourlet}]{
    \includegraphics[height=4cm,keepaspectratio]{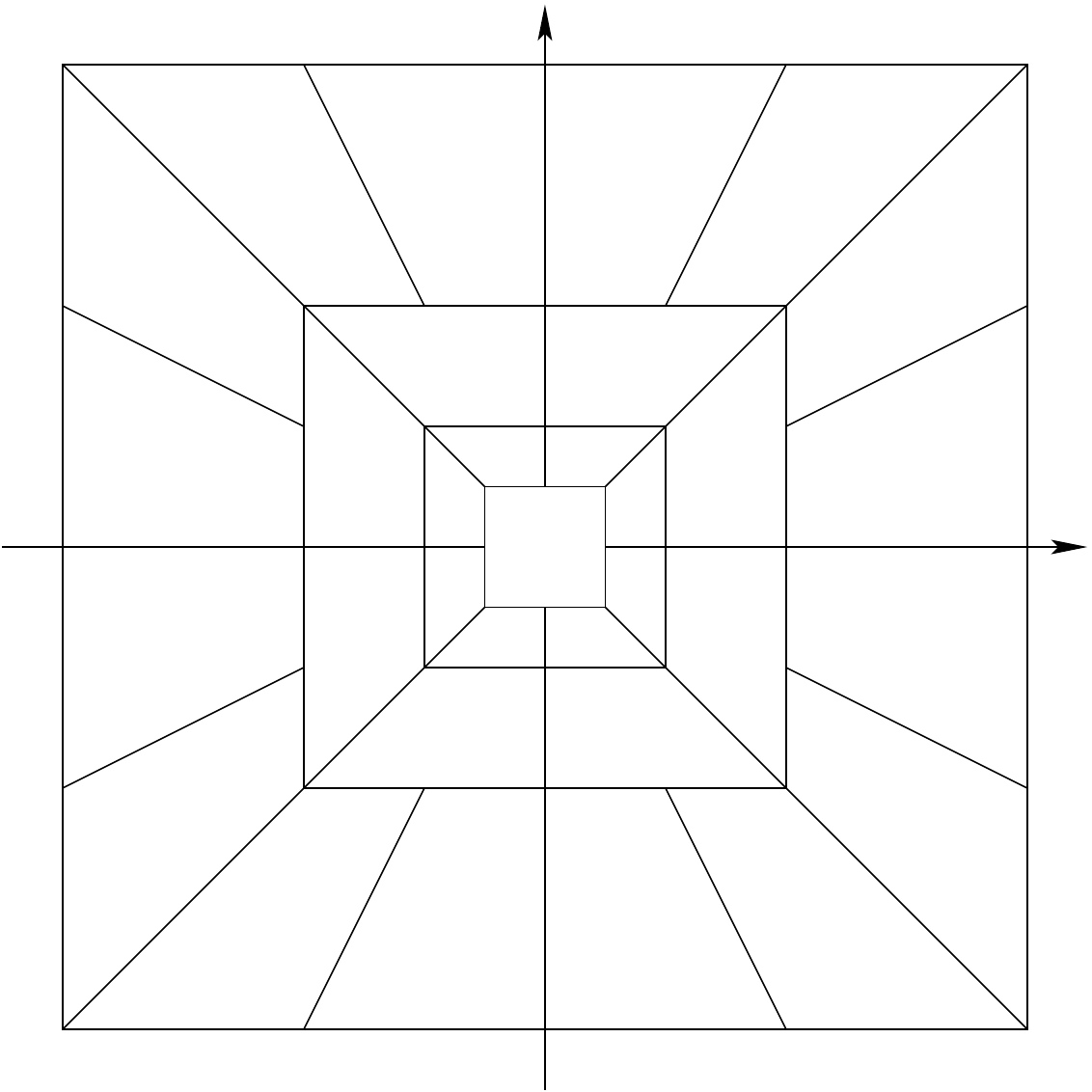}}
  \caption{The contourlet transform. (a)-(b) Two typical atoms; (c) Frequency tiling.}
  \label{fig:fig_BasisContourlet}
\end{figure}

Contourlets \cite{Do_M_2005_tip_con_tedmir} are
sometimes considered a low-redundancy discrete approximation of
curvelets. Actually, they are designed in the spatial domain (instead
of the frequency plane), aiming at a close-to-critical directional
representation.  Their construction is based on  
a Laplacian Pyramid \cite{Burt_P_1983_tcom_lap_pcic} (see
Fig.~\ref{fig:fig_SzegedPyramidLaplacian}).  The low-pass part
of the pyramid is further decomposed with a biorthogoal 9/7 DWT. Each difference image obtained from the pyramid is subject to 
directional filter bank (see Sec. \ref{sec:non-separ-direct}) (initially
from \cite{Bamberger_R_1992_j-ieee-tsp_fil_bdditd},
\cite{Do_M_2003_incoll_con} proposes a simpler implementation based
only on a quincunx structure). A contourlet decomposition is
illustrated\footnote{The contourlet toolbox has been used, see \url{http://www.ifp.illinois.edu/~minhdo/software/}.} in Fig.~\ref{fig:fig_SzegedContourlet}.  The resulting
frequency plane tiling is represented in
Fig.~\ref{fig:fig_FreqPartitionContourlet}. The contourlet inherits
its redundancy of $4/3$ from the pyramidal scheme. Its approximation rate is similar to that of curvelets (Sec. \ref{sec:adaptedgeometry-curvelet}). At one end of the redundancy spectrum, \cite{Lu_Y_2003_p-spie-wasip_cri_ccsdmir} proposes a critically sampled version. At the other end,  the constraints
thus laid on the basis functions 
(Figs.~\ref{fig:fig_SzegedContourlet-proj1}-\ref{fig:fig_SzegedContourlet-proj2}) are relaxed 
 by the design of a more redundant 
\cite{Cunha_A_2006_tip_non_cttda} version, based on non-subsampled (Sec. \ref{sec:adaptedgeometry-mband} )pyramid and directional filters.

\subsubsection{Frames for Oscillating Textures.}
While curvelets, contourlets and shearlets are optimized for the processing of edges, they are not tailored for the processing of oscillating textures, because of their poor frequency localization. 
Generic oscillating patterns can be captured using a local Fourier analysis on a regular segmentation of the image in squares. This corresponds to an expansion in a Gabor frame, see for instance \cite{Mallat_S_2009_book_wav_tspsw}. The spatial segmentation can be optimized using a decomposition in a best cosine packet dictionary as described in Section \ref{sec-tree-struc-dic}. 

Wavelet packets, detailed in Section \ref{sec-tree-struc-dic}, have been used to process and compress oscillating textures such as fingerprints. Brushlets \cite{Meyer_F_1997_j-acha_bru_tdiaic}, introduced by Meyer and Coifman, improve the frequency localization of wavelet packets.

Wave atoms \cite{Demanet_L_2007_j-acha_wav_asop} better capture geometric textures using an anisotropic scaling\footnote{See \url{http://www.waveatom.org}}. The wavelength of wave-atom oscillations is proportional to  the square of their diameter. This scaling allows a thresholding in a wave atom frame to optimally approximate textures obtained by a smooth warping of a sinusoidal profile, see~\cite{Demanet_L_2007_j-acha_wav_asop}.

\section{Redundancy and Adaptivity}
\label{sec:red_adapt}
Highly redundant representations allow us to improve the
representation of complicated images with edges and textures. However,
as described hereafter, computing efficient image representations in
such dictionaries 
  sometimes requires approximations.

\subsection{Pursuits in Redundant Dictionaries}
\label{sec:red_adapt-pursuit}

An approximation $f_M$ of an image $f$ with $M$ atoms from a highly
redundant dictionary $\Basis = \{\psi_{\m_j}:1\leq j\leq P\}$ is
written 
\eq{ f_M = \Psi a = \sum_{j}\ a_{j}\,\psi_{\m_j}, \qwithq
  \normz{a} = \#\enscond{j}{a_{j} \neq 0} \leq M.  
} 
Computing the
$M$-sparse coefficients $a$ that produce the smallest error
$\norm{f-f_M}$ in a generic dictionary is NP-hard
\cite{Natarajan_B_1995_j-siam-comp_spa_asls}. Furthermore, the
$M$-terms approximation $f_M = \ThreshH_T(f,\Basis)$ computed by
thresholding \eqref{eq-thresholding}  might be quite far from the best
$M$-terms approximation. One thus has to use approximate schemes in
order to compute an efficient approximation in a reasonable time.

\subsubsection{Matching Pursuits}
\label{sec:red_adapt-pursuit-matching}
Matching pursuit \cite{Mallat_S_1993_tsp_mat_ptfd} computes $f_M$ from
$f_{M-1}$ by choosing the atom $\psi_{\m}$ that minimizes the
correlation $|\dotp{\psi_{\m}}{f-f_{M-1}}|$. Orthogonal matching
pursuit \cite{Mallat_S_2009_book_wav_tspsw,Pati_Y_1993_p-asilomar_ort_mprfaawd}
further reduces the approximation error by projecting $f$ on the $M$
chosen atoms to compute $f_M$.

Under restrictive conditions on the dictionary $\Basis$, these greedy
algorithms compute an approximation $f_M$ that is close to the best
$M$-term approximation, see for instance
\cite{Tropp_J_2005_j-ieee-tit_gre_garsa,Donoho_D_2006_j-ieee-tit_sta_rsorpn}. These conditions typically
require the correlation $|\dotp{\psi_{\m}}{\psi_{\m'}}|$ to be small
for $\m \neq \m'$, which is not applicable to highly redundant
dictionaries typically used in image processing.

\subsubsection{Basis Pursuit}
\label{sec:red_adapt-pursuit-basis}

A sparse approximation is obtained by convexifying the $\lzero_N$ pseudo
norm, and solving the following basis pursuit denoising convex problem
\cite{Chen_S_1998_j-siam-sci-comp_ato_dbp}
\eql{\label{eq-bpdn}
	f_M = \Psi a = \sum_{j} a_{j} \psi_{\m_j} \qwhereq
	a \in \uargmin{\tilde a\,\in\, \RR^P} \tinv{2}\norm{f-\sum_{j} \tilde a_j \psi_{\m_j}}^2 + \mu\normu{\tilde a},
}
where $\mu>0$ is adapted so that $\normz{a}=M$.
This problem \eqref{eq-bpdn} is minimized, for instance, using iterative thresholding methods \cite{Daubechies_I_2010_j-comm-pure-appl-math_ite_rwlsmsr,Combettes_P_2005_siam-mms_sig_rpfbs}. Algorithmic  solutions to its  generalized form as sums  of convex functions (a common formulation to many data processing problems) may be solved  with great flexibility in the framework of proximity operators  \cite{Combettes_P_2010_incoll_pro_smsp}.

Similarly to matching pursuit algorithms, this $\lun_N$ approximation can be shown to be close to the best $M$-term approximation if the atoms of $\Basis$ are not too correlated, see for instance \cite{Tropp_J_2006_tit_jus_rcpmissn,Donoho_D_2006_j-ieee-tit_sta_rsorpn}.

\subsubsection{Pursuits in Parametric Dictionaries}
\label{sec:dictionaries}

Parametric dictionaries are obtained from basic operations 
(like rotation, translation, dilation, shearing, modulation, etc.)
applied to a continuous mother function.  Even if such dictionaries also
define  redundant bases similar to those introduced earlier, they
deserve a separate description since their parametric nature provides
them with some particular properties. They are generally created to provide a
very rich and dense family of functions built from the geometrical
features of the analyzed image. They have  applications in image
and video coding \cite{Vandergheynst_P_2006_incoll_ima_crd},
multi-modal signal analysis (\eg video plus audio)
\cite{Monaci_G_2006_j-sp_ana_msgvr}, and also for signal
decomposition on non-Euclidean spaces
\cite{Llonch_S_2010_j-patt-rec_3d_frssr}.

Formally, given a set of $S$ transformations $T^i_{m_i}$ for
$1\leq i\leq S$ parameterized by
$m_i\in\Lambda_i\subset\Rbb^{n_i}$, the parametric dictionary is
related to a certain discretization of
$\Lambda^{\ud}\subset\Lambda=\Lambda_1\times \cdots\times \Lambda_S$, \ie 
$$
\Basis\ =\ \{\psi_{\m}(\xx) =
[T^1_{m_1}\cdots T^S_{m_S}\psi](\xx)\in \LL^2(\Rbb^2):\ \m =(m_1,\cdots,m_S)\in\Lambda^{\ud}\}.
$$ 

The directional wavelets described in Sec.~\ref{sec:direct-wavel-fram}
and the subsequent frames built from them are actually an example of
parametric dictionaries with the translations
$T^1_{m_1}T^2_{m_2}=T^1_{b_1}T^2_{b_2}$, the rotation
$T^3_{m_3}=R_\theta$ and the dilation $T^4_{m_4}=D_a$ operations. For
these wavelets, the decomposition/reconstruction methods are
relatively easy to formulate, due to the continuous inversion formula
or using the frame condition.

However, checking the frame condition may sometimes become tedious. In
addition, more transformations of the mother function may be added in
order to enlarge the family of functions, further worsening the frame
bounds.

\begin{figure}[ht!]
  \centering
  \includegraphics[keepaspectratio,height=5cm]{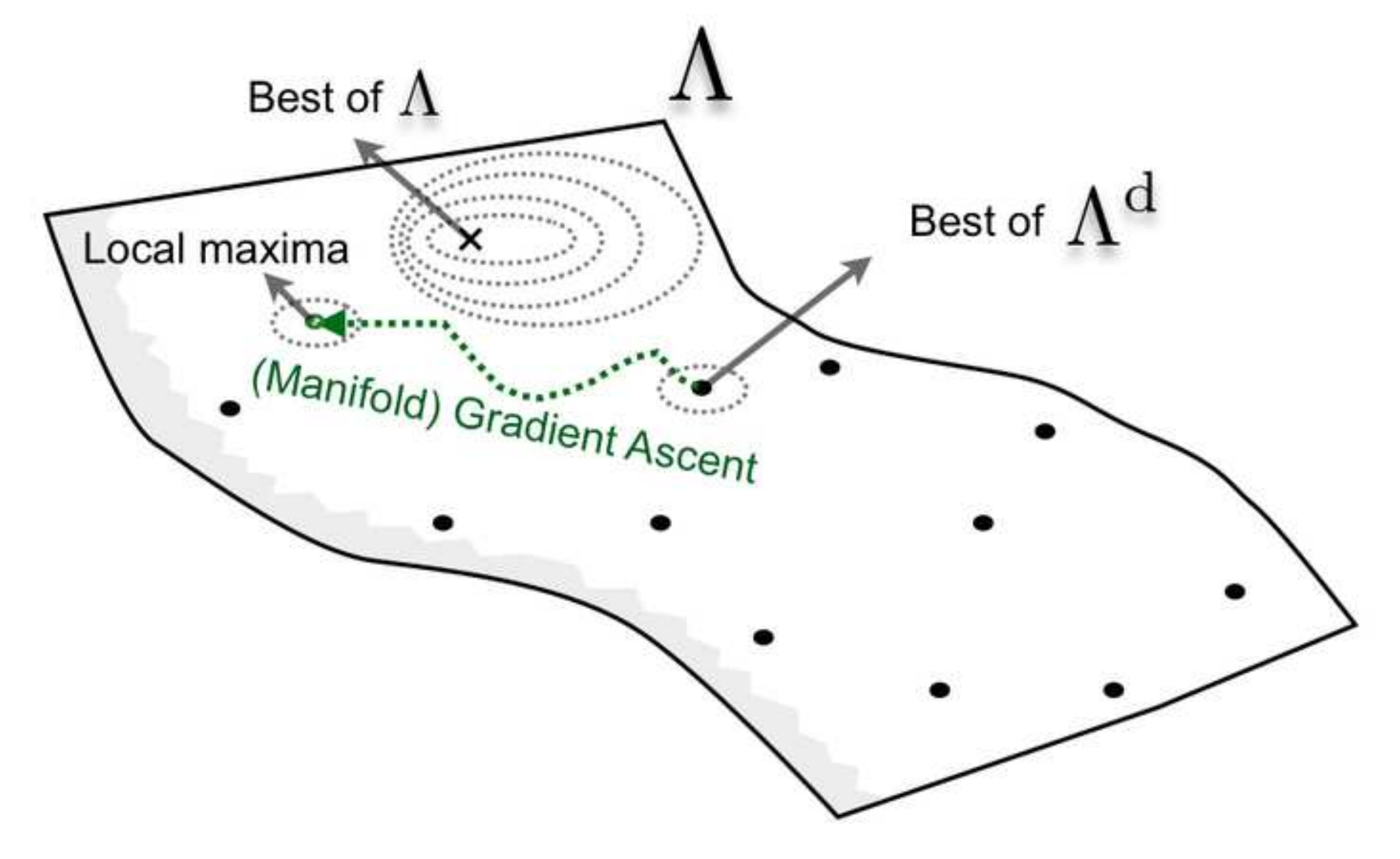}
  \caption{Explanation of the optimization in $\Lambda$ starting from
    a point in $\Lambda^\ud$.
    \label{fig:expl-convmp}}
\end{figure}

Fortunately, as described in Sec.~\ref{sec:red_adapt-pursuit} it is
still possible to find good description of images in very general
family of functions. Most of the time, since the Parametric
Dictionaries are much larger than other dictionaries of controlled
redundancy, the (Orthogonal) Matching Pursuit decomposition
(Sec.~\ref{sec:red_adapt-pursuit-matching}) is used to find a sparse
representation of signals. 

\begin{figure}
\centering
  \subfigure[\label{fig:barbara}]{
    \includegraphics[keepaspectratio,width=0.3\textwidth]{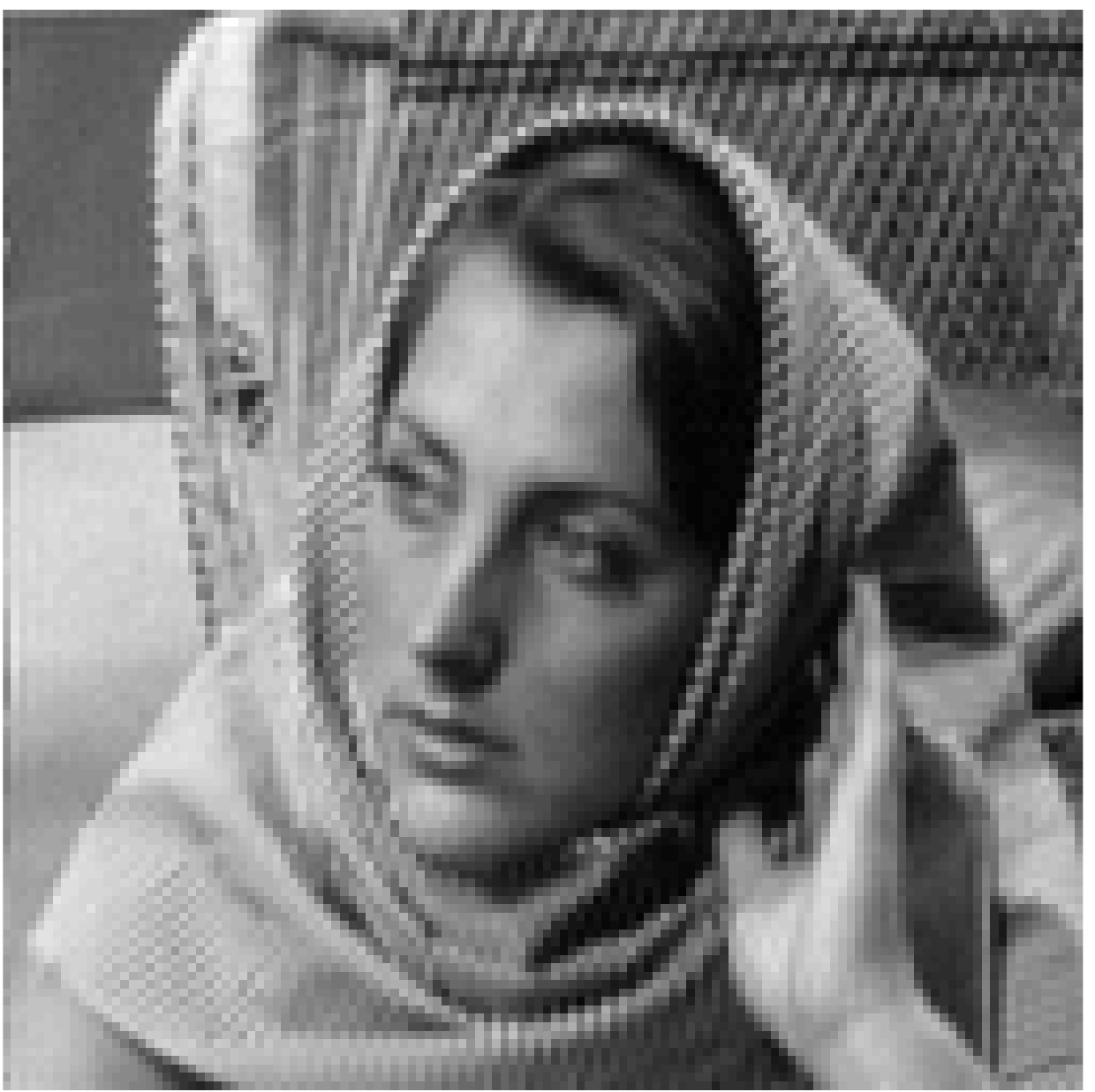}
  }
  \subfigure[\label{fig:barbara-nooptim}]{
    \includegraphics[keepaspectratio,width=0.3\textwidth]{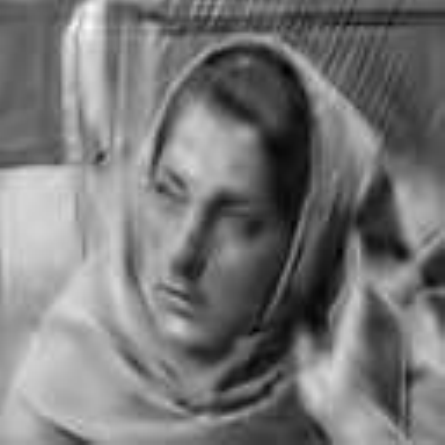}
  }
  \subfigure[\label{fig:barbara-optim}]{
    \includegraphics[keepaspectratio,width=0.3\textwidth]{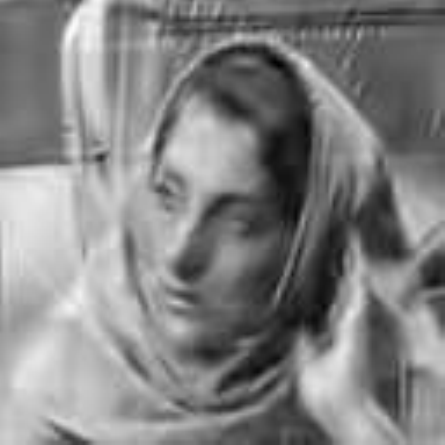}
  }
  \caption{(a) Original image. (b) Reconstruction with 300 atoms for a
    rich parametric dictionary containing $5\times 5$ anisotropic
    scales, 8 orientations, and $N$ translations. PSNR\,: $26.63$\,dB
    (CT: 4634s).  (c) Optimized Reconstruction at 300 atoms starting from a
    dictionary with only $3\times 3$ scales, 4 directions and $N$ translations, PSNR\,:
    $26.68$\,dB (CT: 949s).
    \label{fig:barbara-reconst-optimdic}
  }
\end{figure}

Interestingly, thanks to the parametric nature of $\Basis$, the
dictionary discretization can be refined during the Matching Pursuit
iterations. Indeed, since $\Basis$ is the discretization of the
continuous manifold $\mathcal{M}=\{\psi_{\m}:\m\in
\Lambda\}\subset \LL^2(\Rbb^2)$ generated by all the transformations
of $\psi$, at each iteration of MP in the decomposition of a signal
$f\in \LL^2(\Rbb^2)$ the refinement is performed as follows. As
illustrated on Fig.~\ref{fig:expl-convmp}, given the best atom
$\atom_{\bs m}$ found in $\Basis$, a gradient ascent respecting
the (Riemannian) geometry of $\mathcal{M}$ is run on $\Lambda$ to
maximize the correlation
$S(\m')=|\scp{\atom_{\m'}}{R^n_f}|$ between the
current MP residual $R^n_f = f - f^n$ at step $n$ and the atom
$\atom_{\m'}$. A new parameter $\m^*$ is then used
instead of $\m$ in the signal representation and the next
iteration is realized on the residual $R^{n+1}_f = R^{n}_f -
\scp{\atom_{\m^*}}{R^n_f}\,\atom_{\m^*}$
\cite{Jacques_L_2008_j-ieee-tip_geo_smpp} .

Fig.~\ref{fig:barbara-reconst-optimdic} presents the result of such
an improvement for two different decompositions of the {\tt Barbara}
image (with $N=128^2$ pixels) with similar qualities (expressed using
the Peak Signal-to-Noise Ratio - PSNR). The first one
(Fig.~\ref{fig:barbara-nooptim}) is obtained by a rich parametric
dictionary defined by anisotropic dilations, rotations, and
translations of a \nDim{2} second order directional derivative of a Gaussian. The
second decomposition uses a poorer dictionary with the same
parameterization and mother function but with a manifold optimization
on the atom parameters. The interest of the latter method is to
provide a similar quality for a smaller Computational Time (CT).

\subsubsection{Processing with Highly Redundant Dictionaries}
\label{sec:red_adapt-pursuit-highly_red}

\paragraph{Compression with Sparse Expansions}

Dictionaries with oriented atoms have proven to be successful for
improving the JPEG 2000 compression standard at low bit rates
\cite{Bergeaud_F_1996_j-comput-appl-math-birkhauser_mat_paris,FiguerasIVentura_R_2006_tip_low_rficrr}. The
approximation of the image is computed using the matching pursuit
algorithm.  Matching pursuit in Gabor dictionaries, \ie dictionaries
made of Gabor wavelets (Sec.~\ref{sec:direct-wavel-fram}), have been
used for coding
the motion residual  in video compression
schemes \cite{Neff_R_1997_j-ieee-tcsvt_ver_lbrvcbmp}.

\paragraph{Inverse Problem Regularization}

\newcommand{\nbrMeas}{S}

Data acquisition devices usually only acquire $\nbrMeas$ noisy low resolution
measurements $y = \Phi f_0 + w \in \RR^\nbrMeas$ of a high resolution image
$f_0 \in \RR^N$ of $N \gg \nbrMeas$ pixels. The linear operator $\Phi$ models
the acquisition and might include some blurring and sub-sampling of
the high resolution data.

Recovering a good approximation $f \in \RR^N$ of $f_0$ from these
measurements $y$ corresponds to solving a difficult ill-posed inverse
problem, that requires the use of efficient priors to model the
regularity of the image. Early priors include the Sobolev prior that
enforces smoothness of the image, and the non-linear total variation
\cite{Rudin_L_1992_j-phys-d_non_tvbnra} that can produce sharper
edges.

More recently, $\lun_N$ sparse priors in redundant dictionaries $\Basis$
have been proved to be efficient in order to solve several ill-posed
problems, see for instance \cite{Mallat_S_2009_book_wav_tspsw} and references
therein. In this setting, one computes the coefficients $a$ of $f =
\Psi a = \sum_j a_j \psi_{\m_j}$ in a frame $\Basis$ of $P$ atoms by solving a $\lun_N$
augmented Lagrangian form
\eql{\label{eq-l1-ip-recovery} 
	a \in \uargmin{\tilde a\in \RR^P} \tinv{2} \norm{ y - \Phi \Psi
    \tilde a }^2 + \mu \normu{\tilde a} \qwhereq \Psi \tilde a =
  \sum_j \tilde a_j \psi_{\m_j} 
} 
where $\mu$ should be adapted to the
noise level $\norm{w}$ that is supposed to be known.  This
minimization problem corresponds to computing the basis pursuit
approximation \eqref{eq-bpdn} of the measurements $y$ in the highly
redundant dictionary $\{ \Phi \psi_{\m_j}: 1\leq j\leq P\}$ of
$\RR^\nbrMeas$. It can thus be solved using the same algorithms.

\begin{figure}
\centering
  \subfigure[\label{fig-l1-deconv-a}]{
    \includegraphics[keepaspectratio,width=0.3\textwidth]{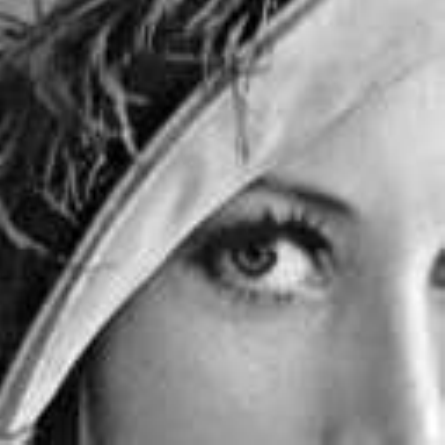}
  }
  \subfigure[\label{fig-l1-deconv-b}]{
    \includegraphics[keepaspectratio,width=0.3\textwidth]{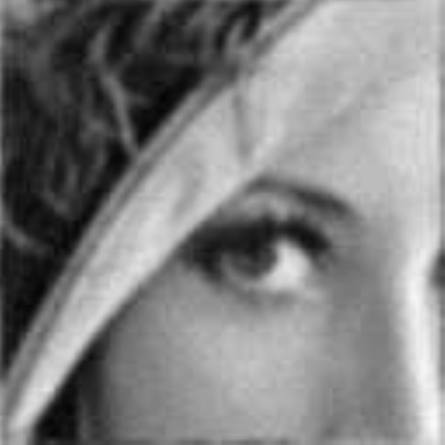}
  }
  \subfigure[\label{fig-l1-deconv-c}]{
    \includegraphics[keepaspectratio,width=0.3\textwidth]{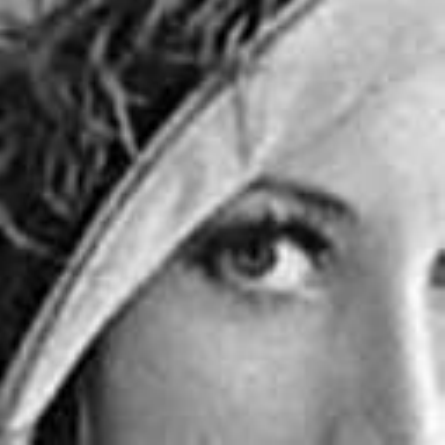}
  }
  \caption{Example of deconvolution using $\lun_N$ regularization in a frame of translation invariant wavelets. (a) Original $f_0$. (b) Observation $y=\Phi f_0 + w$. (c) Deconvolution $f$.
    \label{fig-l1-deconv}
  }
\end{figure}

Figure~\ref{fig-l1-deconv} shows the use of this sparse regularization
method when solving a deconvolution problem. In this application, the
operator is a convolution $\Phi f = f \star G_\sigma$ with a Gaussian
kernel $G_\sigma$ as defined in Sec. \ref{sec:notat-conv}.
The redundant dictionary $\Basis$ is a translation
invariant wavelet frame.

\subsubsection{Source Separation}
\label{sec:red_adapt-pursuit-ss}

\begin{figure}
\centering
  \subfigure[\label{fig-mca-decomposition-a}]{
    \includegraphics[keepaspectratio,width=0.3\textwidth]{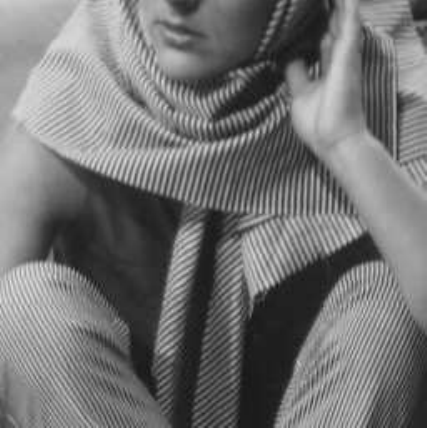}
  }
  \subfigure[\label{ffig-mca-decomposition-b}]{
    \includegraphics[keepaspectratio,width=0.3\textwidth]{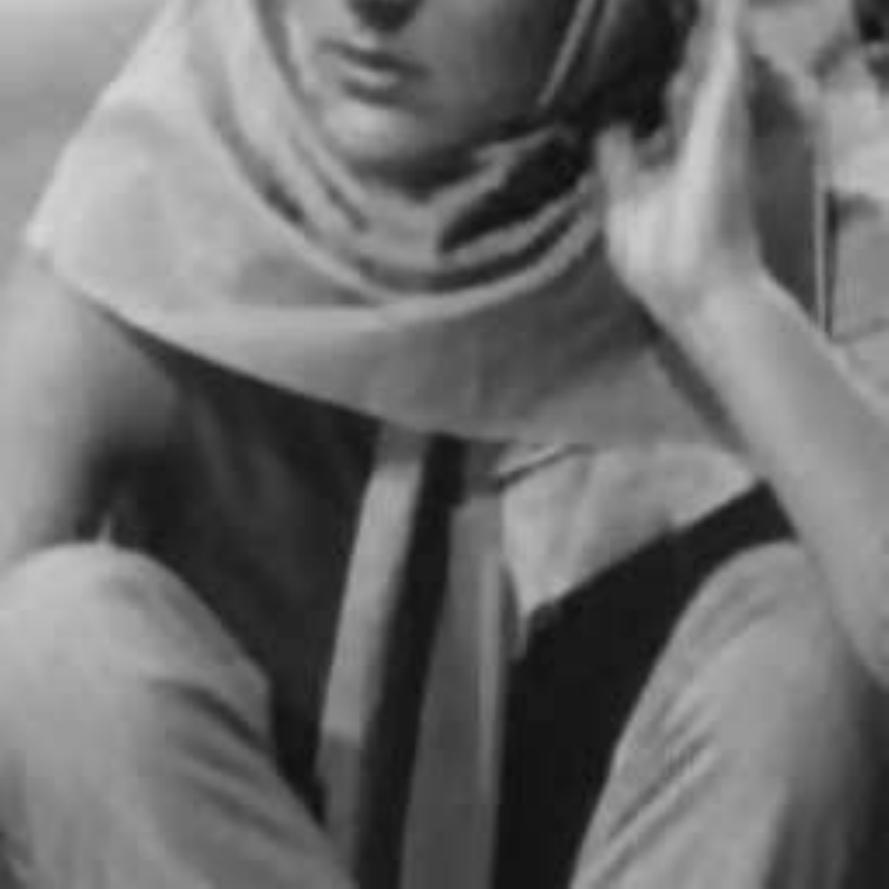}
  }
  \subfigure[\label{fig-mca-decomposition-c}]{
    \includegraphics[keepaspectratio,width=0.3\textwidth]{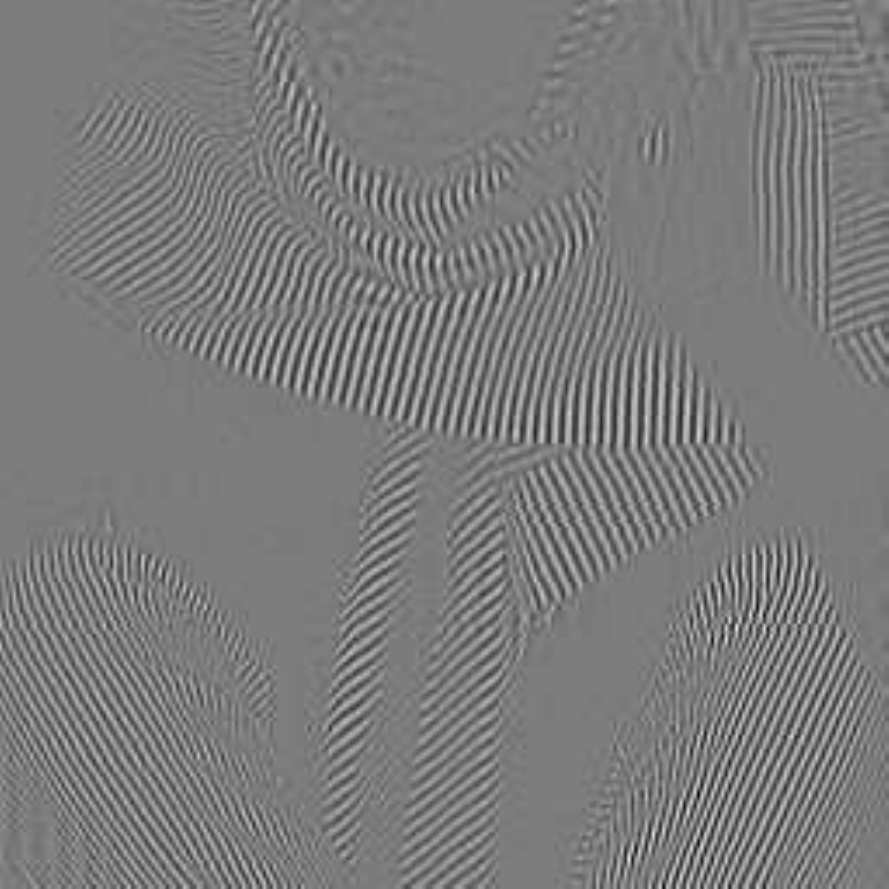}
  }
  \caption{Example of cartoon+texture decomposition using the MCA algorithm. (a) Original $y$. (b) Geometry layer $f_G$. (c) Texture layer $f_T$.\label{fig-mca-decomposition}
  }
\end{figure}

Sparse representations can be used to separate sources that are known
to be sparse in different dictionaries. This corresponds to the
morphological component analysis (MCA) of Starck \etal
\cite{Starck_J_2004_j-adv-imag-electron-phys_red_tamca}. In its simplest setting, it can be used to separate
a single noisy image $y$ into a sum $y = f_G+f_T+w$ of a cartoon-like component
$f_G$ (or \emph{geometric} component), a texture component $f_T$ and
residual noise $w$. One can use a dictionary $\Basis = \Basis_G \cup
\Basis_T$ union of wavelets ($\Basis_G$) and local cosine
($\Basis_T$), and compute a sparse approximation $f$ of $y$ 
\eql{\label{eq-mca} 
	f = \Psi a = \Psi_G a_G + \Psi_T a_T 
} 
where $a = [a_G; a_T]$ is the solution
of the $\lun_N$ basis pursuit \eqref{eq-bpdn} applied to $y$.  
The separation, obtained using $f_G = \Psi_G a_G$ and $f_T = \Psi_T
a_T$, is illustrated in Fig.~\ref{fig-mca-decomposition}.

The modeling of natural images as a sum of a cartoon layer and an oscillating texture layer has been initiated by Y. Meyer in his book \cite{Meyer_Y_2001_incoll_osc_pipnee}. Beside sparsity-based approaches such as~\eqref{eq-mca}, other variational methods have been proposed, see for instance the work of J.-F. Aujol \etal~\cite{Aujol_J_2005_j-math-imaging-vis_ima_dbvcoc}.

\subsection{Tree-structured Best Basis Representations}
\label{sec-tree-struc-dic}

Pursuit algorithms are quite slow and face difficulties in order to
compute provably efficient approximations when the dictionary is too
redundant. In order to avoid these bottlenecks, one needs to consider more
structured representations, that allow one to use fast and provably
efficient approximation strategies. The structuring of the
representation can be implemented by computing an adapted basis
$\basis{\la}$ parameterized by a geometric parameter $\la$ that
captures the local direction of edges or textures. This section
details best basis schemes: they introduce the desired adaptivity
together with fast algorithms employing the hierarchical structure of
parameters $\la$.

\subsubsection{Quadtree-based Dictionaries}

A dictionary of orthonormal bases is a set $\Dd_\La = \{ \basis{\la}
\}_{\la \in \La}$ of orthonormal bases $\basis{\la} =
\{\atom_{\m}^\la\}_{\m}$ of $\RR^N$, where $N$ is the number of pixels
in the image.  Instead of using an \emph{a priori} fixed basis such as
the wavelet or Fourier basis, one chooses a parameter $\la^\star \in
\La$ adapted to the structure of the image to process and then uses
the optimized basis $\basis{\la^\star}$.

\begin{figure}
\centering
    \includegraphics[keepaspectratio,width=0.7\textwidth]{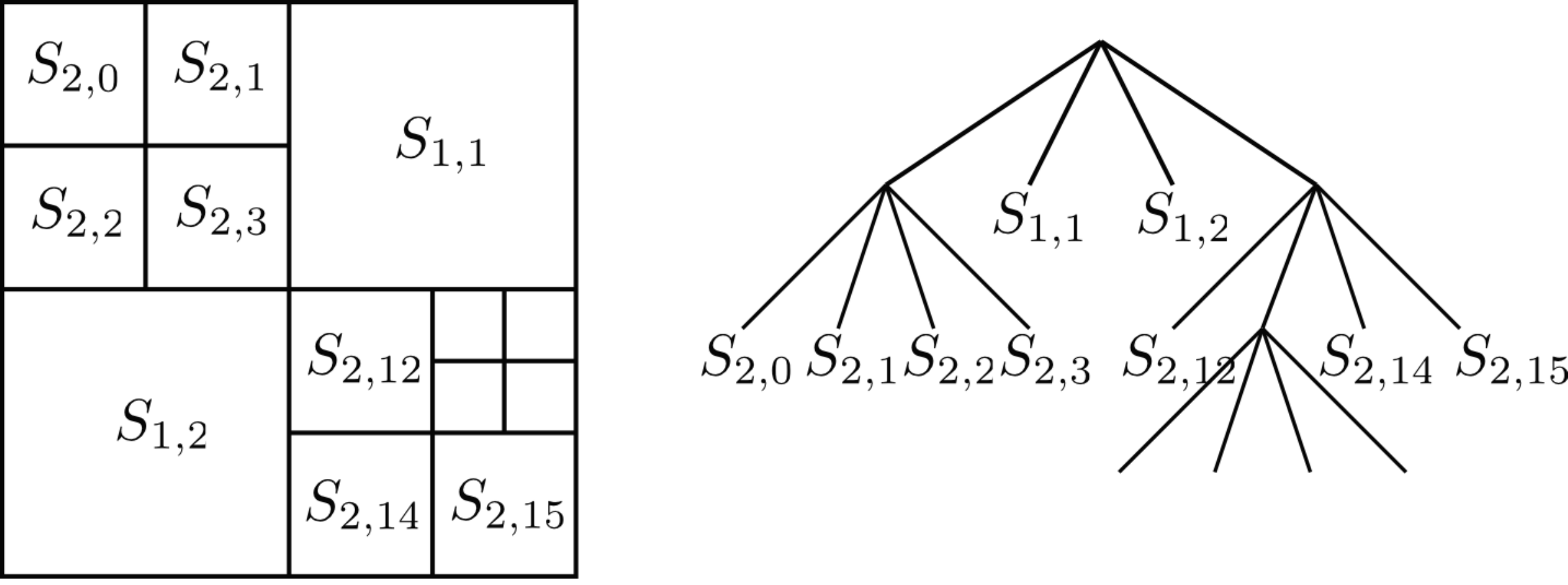}
  \caption{Left: example of dyadic subdivision of $[0,1]^2$ in squares $\sqij$ ; right: corresponding quad-tree $\la$.\label{fig-subdivision-quadtree}
  }
\end{figure}

In order to enable the fast optimization of a parameter $\la^\star$
adapted to a given signal or image $f$ to process, each $\la \in \La$
is constrained to be a quadtree. The quadtree $\la$ that parameterizes
a basis $\basis{\la}$ defines a dyadic segmentation of the square
$[0,1]^2 = \bigcup_{(j,i) \in L(\la)} \sqij$, where $L(\la)$ are the
leaves of the trees, as shown on Fig.~\ref{fig-subdivision-quadtree}.
Each square $\sqij$ is recursively split into four sub-squares
$S_{j+1,4i+k}$ for $k=0,\,\cdots,\,3$. In order to enrich the
representation parameterized by a quadtree, we attach to each leave of
the tree a geometric token, and denote as $\tau$ the number of
tokens. A token indicates the direction of the image geometry
in a square of the segmentation.

\subsubsection{Best Basis Selection}

Given a number $M$ of coefficients, the best basis $\basis{\la^\star}
\in \Dd_\La$ adapted to $f \in \RR^N$ minimizes the best $M$-terms
approximation error. This can be equivalently obtained by minimizing a
penalized Lagrangian that weights the approximation error with the
number of coefficients 
\eql{\label{eq-lagrangian}
	\la^\star  \in \uargmin{\la \in \La}  \lagrangian(f,\basis{\la},T) = 
	\norm{ f-f_M^\la }^2 + M^\la T^2,
}
where $f_M^\la$ is the best $M^\la$-term approximation in $\basis{\la}$ computed by thresholding at $T>0$
\eql{	f_M^\la = \ThreshH_T(f,\basis{\la},T) = \sum_{|\dotp{\atom_{\m}^\la}{f}|>T} \dotp{\atom_{\m}^\la}{f} \atom_{\m}^\la
	\qandq
	M^\la = \#\enscond{\m}{|\dotp{\atom_{\m}^\la}{f}|>T},
}
since $\basis{\la}$ is orthonormal.
This Lagrangian can be re-written as a sum over each coefficient in the basis
\eql{\label{eq-lagr-split-bases}
	\lagrangian(f,\basis{\la},T) = \sum_{\m} \max( |\dotp{ \atom_{\m}^{\la} }{f}|^2, T^2 ).
}
This kind of Lagrangian can be efficiently optimized using a dynamic search algorithm, originally presented by Coifman \etal \cite{Coifman_R_1992_tit_ent_babbs}, which is a particular instance of the Classification and Regression Tree (CART) algorithm of Breiman \etal \cite{Breiman_L_1984_book_cla_rt} as explained by Donoho \cite{Donoho_D_1997_j-annals-statistics-car_bobc}. 
It is possible to consider other criteria for best basis selection, such as for instance the entropy of the coefficients. This  leads different Lagrangians that can be minimized with the same method~\cite{Coifman_R_1992_tit_ent_babbs}. 

The complexity of the algorithm is proportional to the complexity of
computing the whole set of inner products
$\enscond{\dotp{\atom_{\m}^\la}{f}}{\la \in \La}$ in the
dictionary. For several dictionaries, such as those considered in this
section, a fast algorithm performs this computation in $O(P)$
operations, where $P$ is the total number of atoms in $\Dd_\La$. For
tree-structured dictionaries, this complexity is thus $O(\tau N
\log_2(N))$, where $\tau$ is the number of tokens associated to each
leaf of the tree. This is much smaller than the total number of basis
$\basis{\la}$ in $\Dd_\La$, that grows exponentially with $N$.

\subsubsection{Wavelet and Cosine Packets}
\label{sec-tree-wavcospackets}

A basis $\basis{\la}$ with oscillating atoms is defined using a
separable cosine basis over each square of the dyadic segmentation. In
this case no geometry is used, the oscillation of the atoms does not
follow the geometry of the image, and $\tau=1$. An approximation in an
adapted cosine basis $\basis{\la}$ allows one to capture the spatial
variations of a texture \cite{Mallat_S_2009_book_wav_tspsw}.

A wavelet packet basis $\basis{\la}$ defines a dyadic subdivision of
the \nDim{2} frequency domain
\cite{Wickerhauser_M_1991_misc_lec_nwp}. The projection of an image on
the atoms of $\basis{\la}$ is computed through a pyramidal
decomposition that generalizes the orthogonal wavelet transform,
adding flexibility to overcome its dyadic frequency decomposition.
Uniform dyadic wavelet packet decompositions generate a subset of
$M$-band wavelets with equal-span frequency subbands obtained from $J$
decomposition levels, with $M=2^J$. In order to adapt to the specific
frequency content of the image, the resulting tree is parsed through a
best basis selection procedure \cite{Coifman_R_1992_tit_ent_babbs},
reminiscent of the subdivision in Fig.~\ref{fig-subdivision-quadtree}.

This construction is generalized by considering non-stationary wavelet
packets \cite{Cohen_A_1998_incoll_non_ssmawp}, that apply different quadrature
mirror filters at each scale of the tree.  A dynamic programming
algorithm detailed in \cite{Ouarti_N_2009_p-icip_bes_bsnswpd} computes an
adapted non-stationary basis.

\subsubsection{Adaptive Approximation}

\paragraph{Wedgelets}

A geometric approximation is obtained by considering for each node of
the dyadic segmentation a collection of $\tau$ different low-dimensional discontinuous approximation spaces
\cite{Donoho_D_1999_j-annals-statistics_wed_nmee}. For each node of
the quadtree, a token indicates the local direction and position of
the edge. The low-dimensional approximation spaces are piecewise polynomials
over each of the two wedges.

The wedgelets introduced by Donoho
\cite{Donoho_D_1999_j-annals-statistics_wed_nmee} rely on piecewise
constant approximation. This scheme is efficient when approximating a
piecewise constant image $f$ whose edges are $\Cdeux$ curves. For such
cartoon images, the approximation error decays like $\norm{f-f_M}^2 =
M^{-2}$, see
\cite{Donoho_D_1999_j-annals-statistics_wed_nmee,Fuhr_H_2006_incoll_bey_wnirp}. It is also possible to consider approximation spaces with higher-order
polynomials in order to capture arbitrary cartoon images
\cite{Shukla_R_2005_tip_rat_dotscappi}, see also
\cite{Kassim_A_2009_j-ieee-tip_hie_sbichqbt} for a related
construction. The computation of the low-dimensional projection can be
significantly accelerated, see \cite{Friedrich_F_2007_j-siam-sci-comp_eff_mcpdarwa}.

The piecewise constant model for images being relatively simplistic,
wedgelets have been upgraded to platelets
\cite{Willett_R_2003_tmi_pla_maresplmi} and surflets \cite{Chandrasekaran_V_2009_j-ieee-tit_rep_cmpfs}. They aim at improving
the management of smooth intensity variations, since they rely on
planar or even smoother approximation on dyadic square or wedge based grids.

\paragraph{Bandlets}

For coding, orthogonal expansions are preferred over low-dimensional approximations as considered by wedgelets. Switching to
non-linear approximation in bases also better handles directional
textures that do not correspond to a fixed low-dimensional space
parameterized by a wedge.

\begin{figure}[htb!]
  \centering
  \subfigure[\label{fig-bandlets-a}]{
	{\includegraphics[width=.3\linewidth,keepaspectratio]{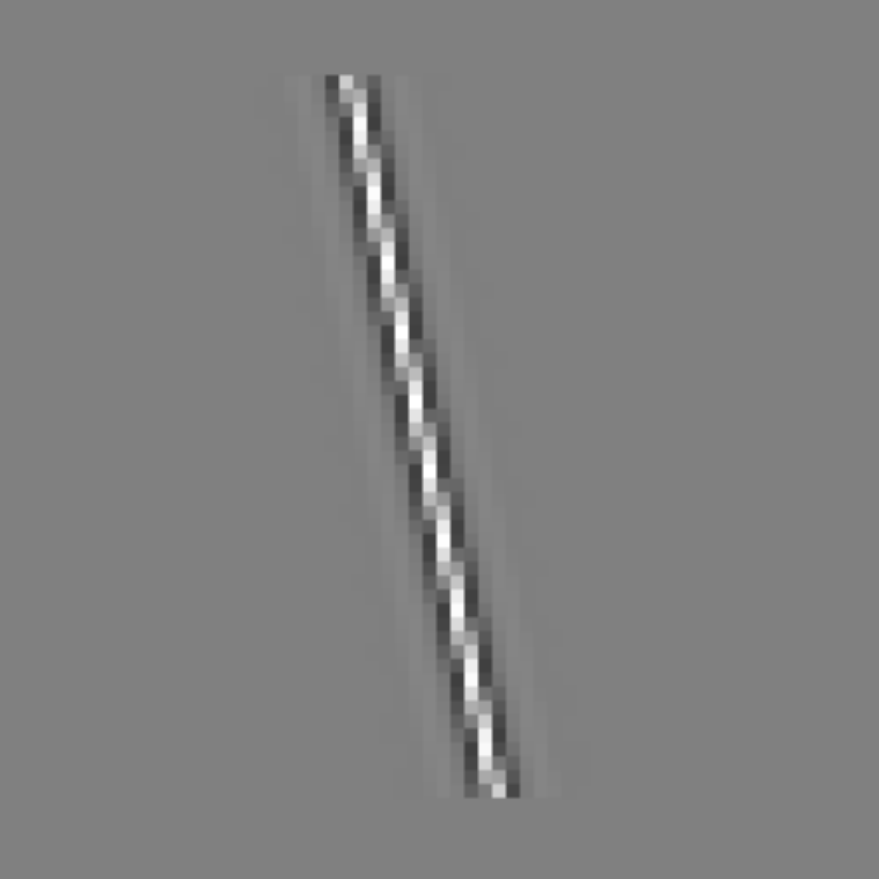}}}  \subfigure[\label{fig-bandlets-b}]{
	{\includegraphics[width=.3\linewidth,keepaspectratio]{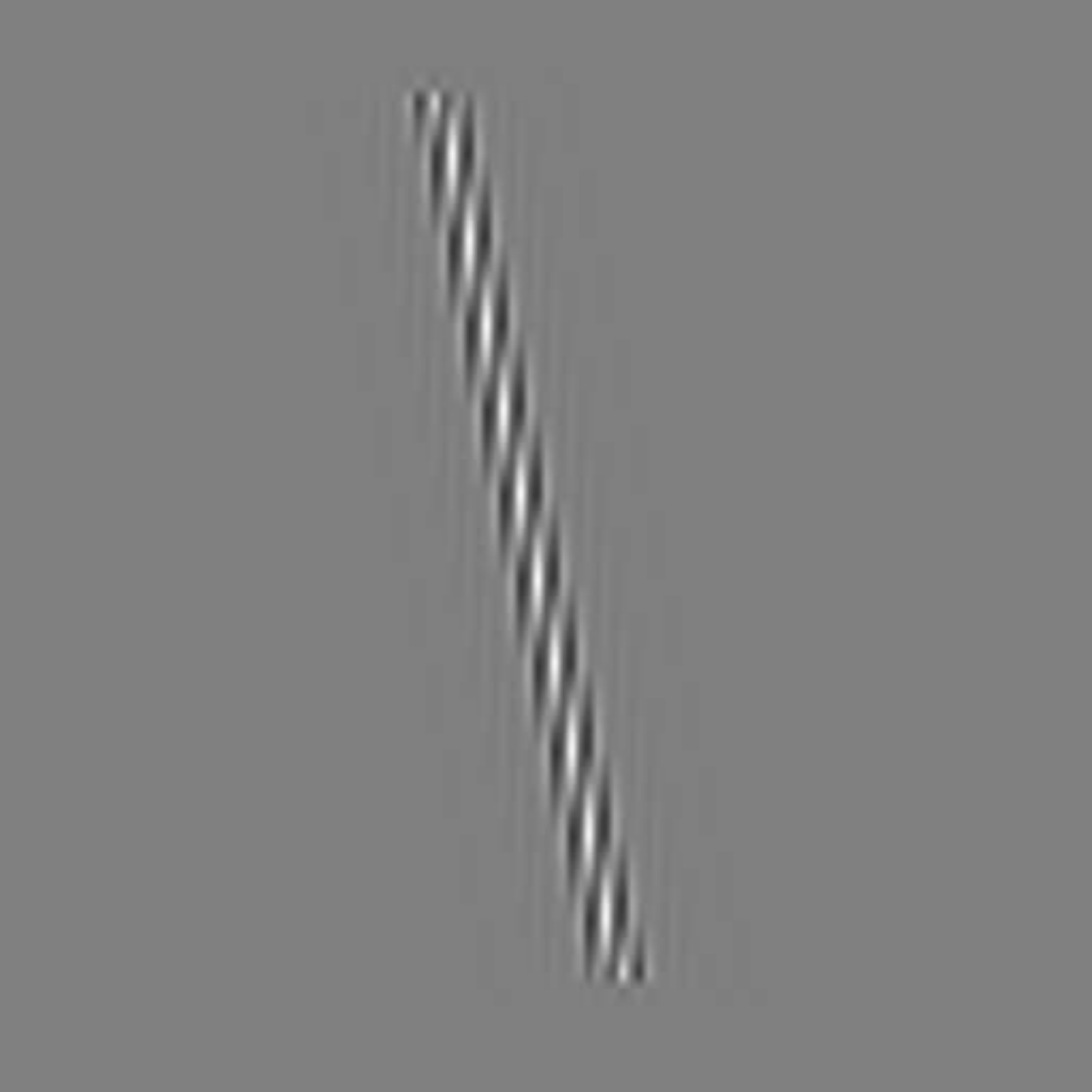}}}
  \caption{Example of a bandlet atom. (a) Atom in the spatial domain. (b) Wavelet-bandelet atom.}
  \label{fig:fig_SzegedBandelet-projfreq}
\end{figure}

The bandlet bases dictionary is introduced by Le~Pennec and Mallat
\cite{LePennec_E_2005_j-siam-mms_ban_ac}. Bandlets perform an efficient adaptive
approximation of images with geometric singularities. An anisotropic
basis with a preferred orientation is defined over each square of the
dyadic segmentation. Fig.~\ref{fig:fig_SzegedBandelet-projfreq} (a)
shows an example of bandlet atom. The orientation is parameterized
with the token stored in the leaf of tree.  Keeping only a few
bandlet coefficients and setting the others to zero performs an
approximation of the original image that follows the local orientation
indicated by the token. 

\paragraph{Adaptive Approximation over the Wavelet Domain}

Applying such an adaptive geometric approximation directly on the
image leads to unpleasant visual artifacts. In order to overcome this
issue, one applies a tree-structured approximation or a best basis
computation on the discrete set of wavelet coefficients. The
wedgeprint of Wakin \etal \cite{Wakin_M_2006_j-ieee-tip_wav_dacpsi} uses a vector
quantization to extend the wedgelet scheme to the wavelet domain. The
second generation bandlets of Peyr\'e and Mallat
\cite{Peyre_G_2008_j-comm-pure-appl-math_ort_bbgia} use an adaptive
bandlet basis for each scale of the wavelet transform.  All these
methods benefit from the same approximation error decay as their
single scale predecessors, but work better in practice.

Fig.~\ref{fig:fig_SzegedBandelet-projfreq} shows how a bandlet atom
(a) is mapped to a wavelet-bandlet atom (b).  Decomposing an image
over a bandlet basis composed of atoms of type (b) is equivalent to
applying first a wavelet transform, and then decomposing the wavelet
coefficients over atoms of type (a).

Another adaptive approximation relying on the processing of the wavelet
domain is the easy path wavelet transform (EPWT)
\cite{Plonka_G_2009_j-siam-mms_eas_pwt_nawtsrtdd}. It provides a hybrid
and adaptive approach exploiting the local correlations of images
along path vectors through index subsets in the Wavelet domain.

\subsubsection{Adaptive Tree-structured Processing}

For compression and denoising applications, one computes the best
basis $\basis{\la^\star}$ adapted to the image $f$ to compress or
denoise by minimizing the corresponding Lagrangian
\eqref{eq-lagr-split-bases}. The coefficients
$\dotp{\atom_{\m}^\la}{f}$ are then binary coded (for compression) or
thresholded (for denoising). The resulting improvement of the best
basis approximation error over wavelets translates into improvement in
the rate distortion (for compression) or average risk (for denoising)
of the best basis method, see for instance
\cite{Wakin_M_2006_j-ieee-tip_wav_dacpsi,Peyre_G_2008_j-comm-pure-appl-math_ort_bbgia}.

One can also use best bases to recover an image from noisy low-dimensional measurements $y=\Phi f + w$ where $\Phi$ is an ill-conditioned
linear mapping. For some problems such as inpainting, small missing
regions or 
light blur removal, 
the best basis $\la$ can be
estimated directly from the observation $y$. 

\begin{figure}
\centering
    \includegraphics[keepaspectratio,width=0.9\textwidth]{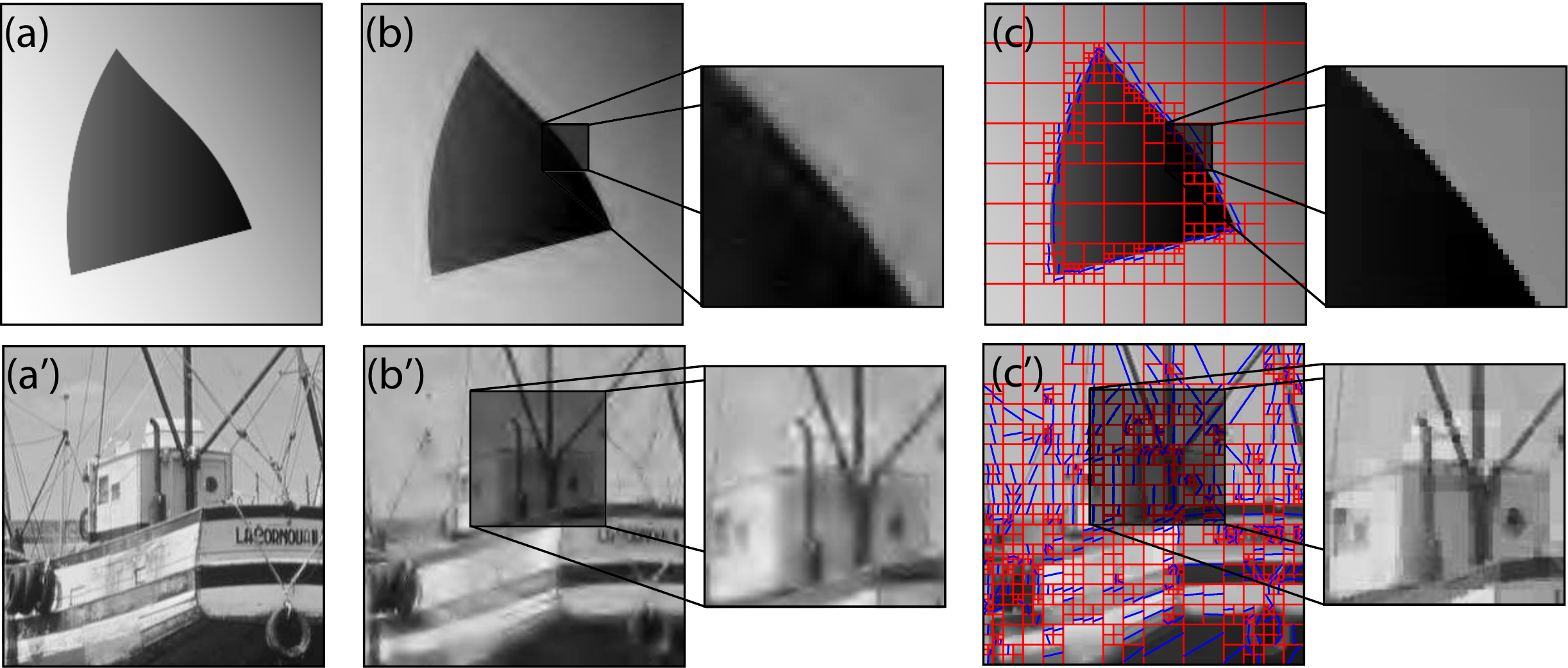}
  \caption{(a,a') original image ; (b) compressed sensing reconstruction
using a translation invariant wavelet frame (PSNR=37.1dB) ;
(c) reconstruction using a best bandlet basis (PSNR=39.3dB).
(b') wavelet frame, PSNR=22.1dB, (c') bandlet basis, PSNR=23.9dB.\label{fig-image-triangle-boat}
  }
\end{figure}

An example of inverse problem where sparsity in a best basis
significantly improves over sparsity in a fixed basis is compressed
sensing. Compressed sensing is a new data sampling strategy, where the
measurement operator $\Phi$ of size $P \times N$ is generally the
realization of some random matrix ensemble.  The sampling operations
$y = \Phi f + w \in \RR^P$ allows one to acquire a high resolution
signal $f \in \RR^N$ directly in a compressed format of $P<N$
measurements. Compressed sensing theory ensures that if the number of
measurements $P$ is large enough with respect to the sparsity $K$ of
the signal $f$ in a basis $\Bb$, typically, $P=O(K \log N/K)$ for
Gaussian random matrix $\Phi$, one recovers a good approximation of
the signal using a $\lun_N$ sparse regularization as
in~\eqref{eq-l1-ip-recovery}. It can be shown that the quality of the
reconstruction depends both on the sensing noise power $\|w\|$ and on
the ``compressibility'' of $f$, that is, its deviation from the
strictly sparse case. We refer to the review paper of
Cand\`es~\cite{Candes_E_2006_p-int-congress-math_com_s} and the references therein for more
details. Fig.~\ref{fig-image-triangle-boat} shows a comparison of
compressed sensing recovery from $P=N/6$ measurements using a
redundant frame $\Bb$ of translation invariant wavelets, and a best
bandlet basis. In this last result, it is necessary to use an
iterative algorithm that progressively improves the quality of the
estimated geometry, see~\cite{Peyre_G_2010_j-ieee-tsp_bes_bcs}. As explained in this last
reference, the same technique can be used for \emph{inpainting} large
holes in images.

\subsubsection{Adaptive Segmentations and Triangulations}

In order to enhance the quality of the representation, it is possible
to consider tree-structured segmentations $[0,1]^2 = \bigcup_{\be \in
  \la} \be$ of the image where the boundaries of the sub-domains $\be
\in \la$ are not restricted to be axis-aligned. The advantage is that
such an adaptive segmentation defines regions $\be \in \la$ with
arbitrary complicated boundaries. Unfortunately, the combinatorial
explosion of the set of all possible $\la$ forbids the search for an
optimal segmentation with a fast algorithm. One has thus to use a
greedy scheme that selects at each step a split to reduce the
approximation error.

\paragraph{Recursive Splitting and Approximation Spaces}

A greedy scheme computes an embedded segmentation $\la = \{ \la_j
\}_{j}$, where $\la_{j+1} \subset \la_j$ is obtained by splitting a
region $\be \in \la_j$. The full segmentation $\la$ can thus be
represented and coded using a binary tree. This defines
multiresolution spaces $V_{\la_{j+1}} \subset V_{\la_j}$ where
$V_{\la_j}$ is composed, for instance, of piecewise polynomial
functions on each region $\be \in \la_j$.

It is possible to compute a single-scale orthogonal projection $f_M =
P_{V_{\la_j}}(f)$ of an image $f$ on a fixed resolution space
$V_{\la_j}$ in order to perform image approximation or compression. It is also
possible to define a detail space $V_{\la_{j+1}} = V_{\la_j} \oplus
W_{\la_j}$. A wavelet basis $\basis{\la}$ can be built by considering
a basis for each $W_{\la_j}$. A non-linear thresholding approximation
$f_M = \ThreshH_T(f,\basis{\la},T)$ provides an additional degree of
adaptivity and reduces the approximation error $\norm{f-f_M}$. Wavelet
bases on adaptive segmentations also enable a progressive coding of
the coefficients by decaying $T$, which is important for image
compression applications.

\paragraph{Adaptive Segmentation}

A popular splitting rule is the binary space tiling, that splits
a region $\be \in \la_j$ according to a straight line, see for
instance \cite{Dekel_S_2005_j-siam-j-numer-anal_ada_mabspgw}.

Other popular approaches restrict the regions $\be \in \la_j$ to  triangles, so that $\la_j$ is a triangulation of the domain $[0,1]^2$. It is possible to refine the triangulation by adding new vertices, or on the contrary to remove vertices to go from $\la_{j+1}$ to $\la_j$. These vertex-based schemes do not satisfy $\la_{j+1} \subset \la_j$, so one cannot build a wavelet basis using such triangulations. These vertex refinement methods generate a single scale approximation $P_{V_{\la_j}}(f)$ and lead to efficient image coders, see for instance \cite{Demaret_L_2006_j-sp_ima_clsat}.

To generate embedded approximation spaces $\la_{j+1} \subset \la_j$,
one needs to split the triangles $\be \in \la_j$. Regular split of
orthogonal triangles leads to isotropic adaptive triangulations
\cite{Distasi_R_1997_j-ieee-tcom_ima_cbttc}. Splitting triangles
according to a well chosen median leads to anisotropic triangulations
that exhibit optimal aspect ratio for smooth images, see
\cite{Cohen_A_2011_PREPRINT_ada_mabat}.  More complicated,
non-linear coding schemes are possible, for instance using normal
meshes \cite{Jansen_M_2005_j-acha_mul_apstdfntm}, that treat an image
as an height field.

\subsection{Lifting Representations}
\label{sec-lifting}

To enhance the wavelet representation, the wavelet filters can be
adapted to the image content. The lifting scheme,  popularized by Sweldens
\cite{Sweldens_W_1997_j-siam-math-anal_lif_scsgw} and latent in earlier works \cite{Dyn_N_1987_j-comput-aided-geomet-desfou_pisccd,Bruekers_F_1992_j-ieee-sel-areas-com,Hampson_F_1998_j-ieee-tip_m_bnsdpr}, is an unifying framework to design
adaptive biorthogonal wavelets, through the use of spatially varying
local interpolations. While it can typically reduce the computation of
the wavelet transform by a factor of about two in \nDim{1}, it also guarantees 
perfect reconstruction for arbitrary filters, and can be used (Sec.~\ref{sec:lift-scheme-wavel}) on
non-translation invariant grids to build  wavelets on surfaces, see
Sec.~\ref{sec:noneuclidian}.

\subsubsection{Lifting Scheme}
\label{sec-lifting-scheme}

\begin{figure}[t!]
  \centering
\subfigure[\label{fig:lifting}]{
  \raisebox{5mm}{\includegraphics[width=7cm]{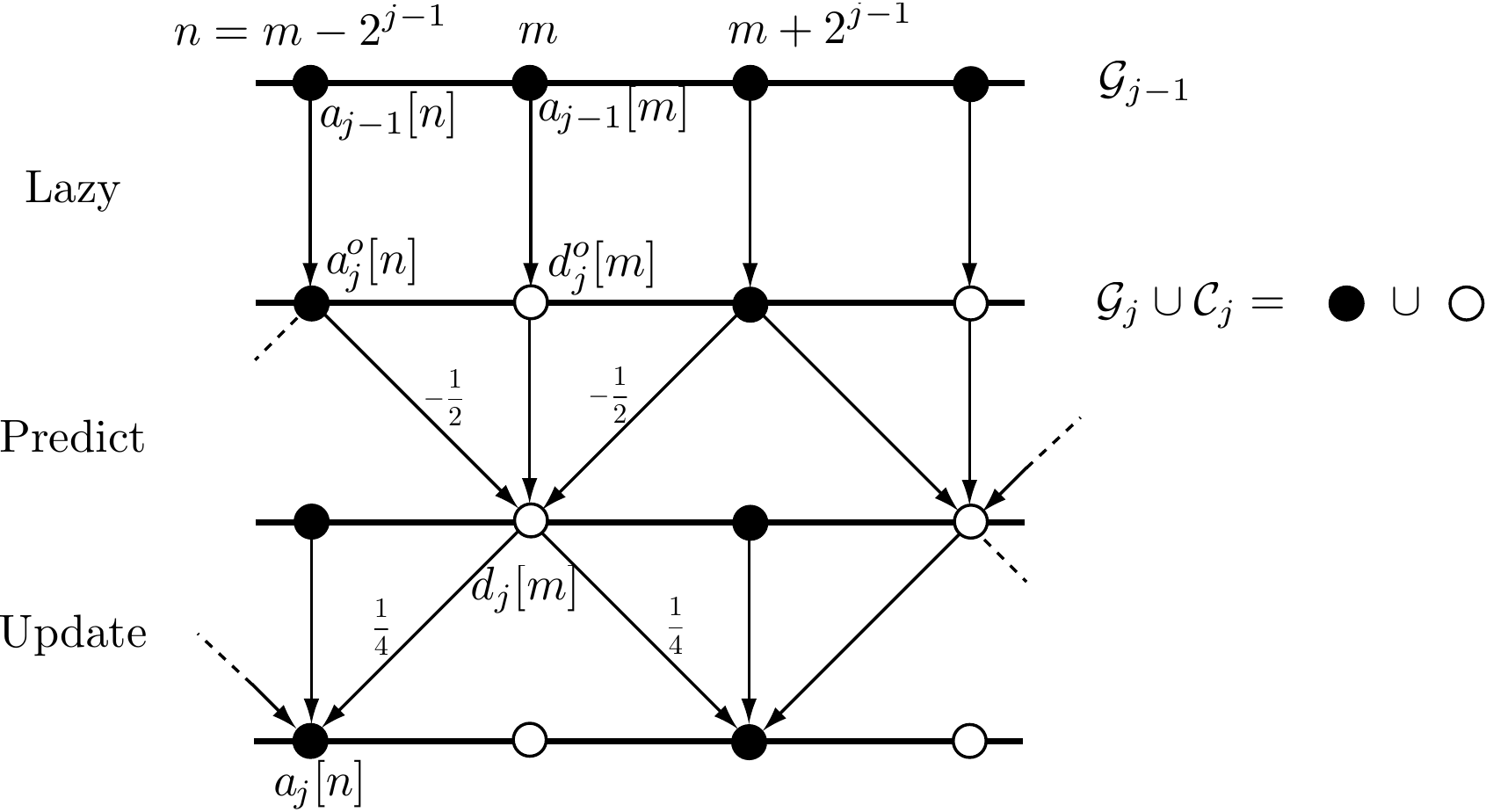}}}\hspace{15mm} 
\subfigure[\label{fig:fig_SzegedLISQ_Lifting}]{
  \includegraphics[height=6cm]{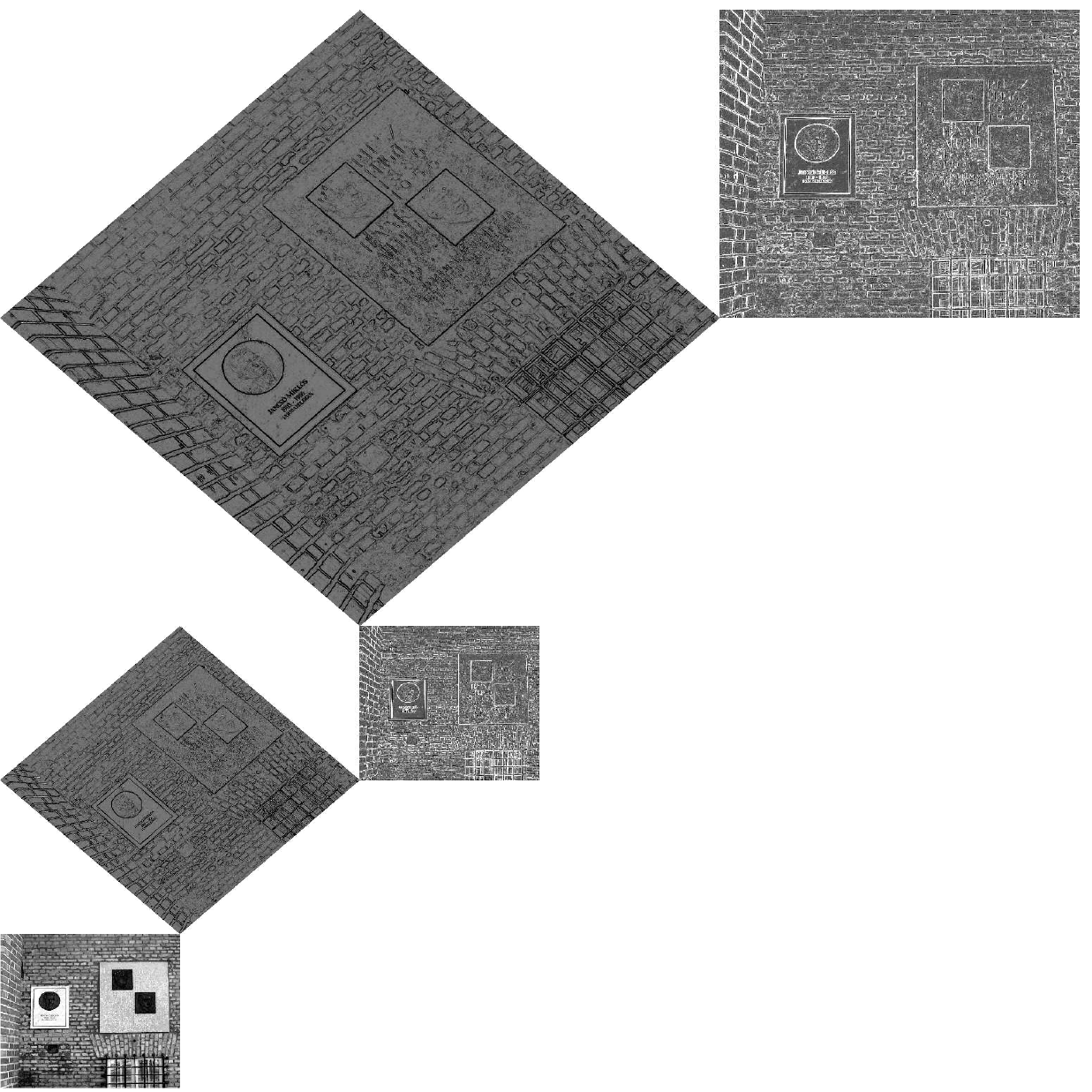}
}
  \caption{(a) Predict and update lifting steps (b) MaxMin lifting \imageHRP.}
\end{figure}

At each scale $j$, the scaling  coefficients $a_{j-1}$ are
evenly split  into two groups $a_j^o$ and $d_j^o$. The wavelet
coefficients $d_j$ and the coarse scale coefficients $a_j$ are
obtained by applying linear operators $P_j^{\la_j}$ and $U_j^{\la_j}$
parameterized by ${\la_j}$ 
\eql{\label{lifting-step} d_j = d_j^o -
  P_j^{\la_j} a_j^o \qandq a_j = a_j^o + U_j^{\la_j} d_j.  
}  
The resulting lifted wavelet coefficients $\{d_j\}_j$ are thresholded or quantized to achieve denoising or compression. These two lifting or ladder steps are easily inverted by reverting the order of the operations. The predictor $P_j^{\la_j}$ interpolates the sub-sampled values $a_j^o$ in order to reduce the amplitude of the wavelet coefficients $d_j$, while the update mapping $U_j^{\la_j}$ stabilizes the transform by maintaining certain quantities such as the mean of the scaling coefficients. By applying sequentially several predict and one update operators, one can recover arbitrary biorthogonal wavelets on uniform \nDim{1} grid \cite{Daubechies_I_1998_j-four-anal-appl_fac_wtls}, speeding up the wavelet decomposition algorithm by a factor of about two in \nDim{1}. 
The lifting structure in Fig. \ref{fig:lifting} corresponds to the 5/3 lifted wavelet. Such structures may furthermore adapt to non-linear filters and  morphological operations 
\cite{Taubman_D_1999_p-icip_ada_nsltic,Egger_1995_j-proc-ieee_hig_cicamsd}. An example\footnote{LISQ toolbox: \url{http://www.mathworks.com/matlabcentral/fileexchange/13507}.}  of lifting based quincunx  scheme example from \cite{Goutsias_J_2000_j-ieee-tip_non_msds1mp,Heijmans_H_2000_j-ieee-tip_non_msds2mw} is displayed  in Fig. \ref{fig:fig_SzegedLISQ_Lifting}.

\subsubsection{Adaptive Predictions}

It is possible to design the set of parameter $\la = \{\la_j\}_j$ to adapt the transform to the geometry of the image. We call $\la_j$ an association field, since it typically links a coefficient of $a_j^o$ to a few neighboring coefficients in $d_j^o$. Each association is optimized to reduce, as much as possible, the magnitude of  wavelet coefficients $d_j$, and should thus follow the geometric structures in the image. One can compute these associations to reduce the length of the wavelet filter near the edges, using the information from the coarser scale \cite{Claypoole_R_2003_tip_non_wticl}.  Locally adaptive schemes have proven efficient in stereo and video coding \cite{Gouze_A_2004_j-ieee-tip_des_samlslc,Kaaniche_M_2009_j-ieee-tip_vec_lssic,Quellec_G_2010_j-ieee-tip_ada_nwtlacbir,Kaaniche_M_2011_PREPRINT_non_slsausssic}.

 Such schemes are related to adaptive non-linear subdivision \cite{Cohen_A_2002_inbook_non-ssaip}. To further reduce the distortion of geometric images, the orientations of the association fields $\{ \la_j \}_j$ can be optimized though the scales.  Because of the lack of structure of the set of bases $\basis{\la}$, computing the field $\la_j$ that produces the best non-linear approximation is intractable. These flows are thus usually computed using heuristics to detect the local orientation of edges, see for instance \cite{Gerek_O_2000_j-ieee-tip_ada_psdsic,Yin_B_2008_j-spic_dir_lbwtmdic,Chappelier_V_2006_tip_ori_wticd,Heijmans_H_2005_j-acha_bui_nawul}.
These adaptive lifting schemes are extended to perform adaptive video transforms where the lifting steps operate in time by following the optical flow $\la_j$, see for instance \cite{PesquetPopescu_B_2001_p-icassp_thr_dlsmcvc,Secker_A_2003_j-ieee-tip_lif_bimatlimatfhsvc}.

\subsubsection{Grouplets}

A difficulty with lifted transforms is that they do not guarantee the
orthogonality of the resulting wavelet frame. The stability of the
transform thus tends to degrade for complicated association fields
$\{\la_j\}_j$. The grouplet transform, introduced by Mallat
\cite{Mallat_S_2009_acha_geo_g}, also makes use of association fields,
but it replaces the lifting computation of wavelet coefficients by an
extended Haar transform, where coefficients in $d_j^o$ are processed
in sequential order to maintain orthogonality.

\begin{figure}[htb!]
  \centering
  \subfigure[\label{fig-grouplet-synth-a}]{
	{\includegraphics[height=3.5cm,keepaspectratio]{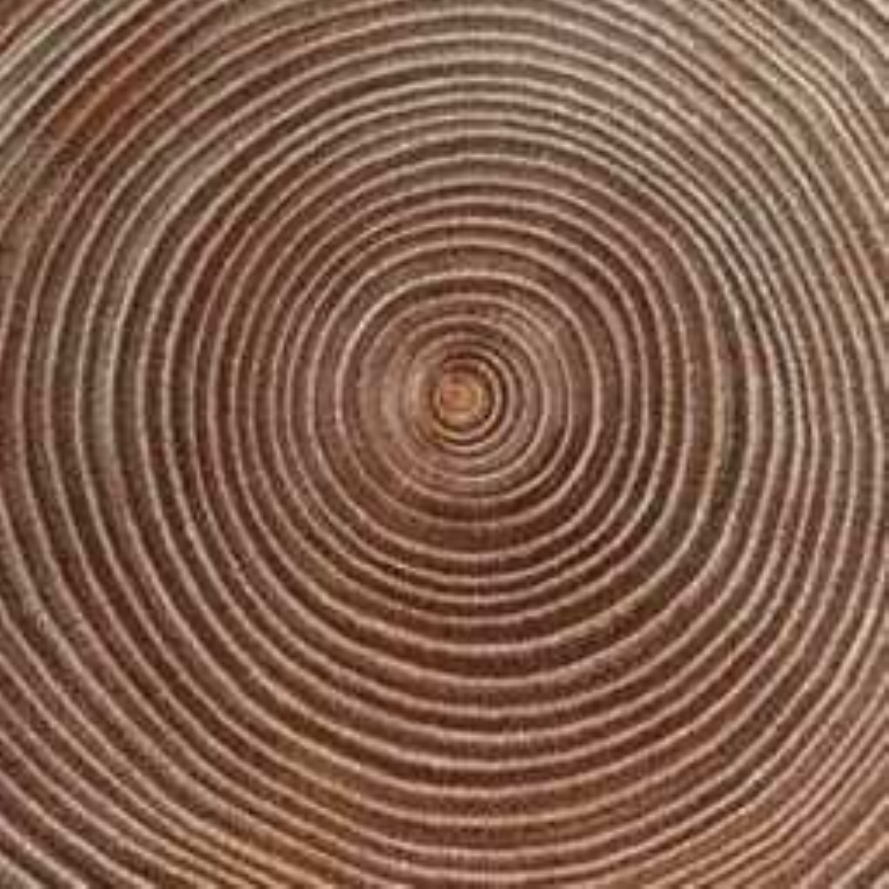}}}
  \subfigure[\label{fig-grouplet-synth-b}]{
    \includegraphics[height=3.5cm,keepaspectratio]{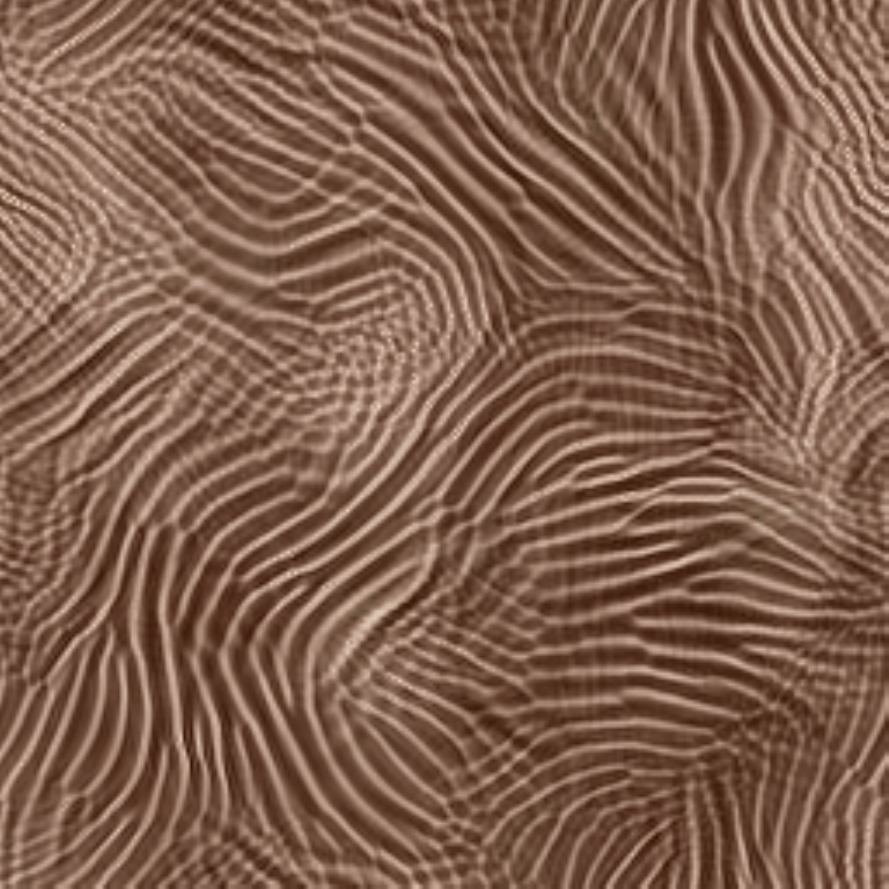}}
  \subfigure[\label{fig-grouplet-synth-c}]{
	{\includegraphics[height=3.5cm,keepaspectratio]{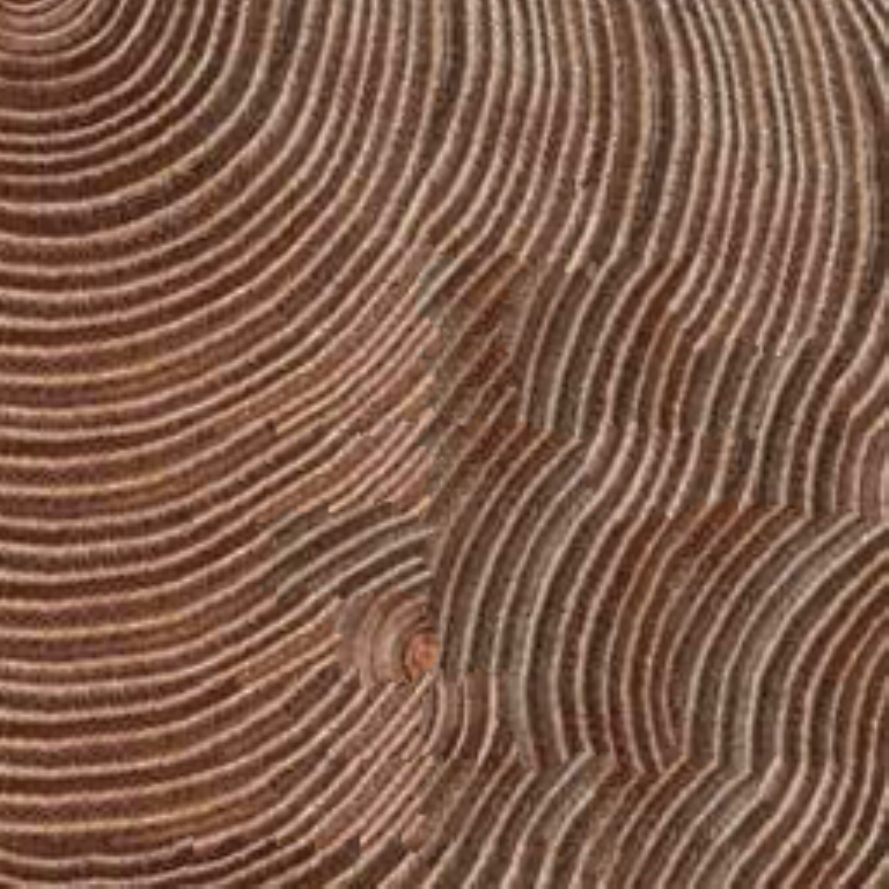}}}
  \subfigure[\label{fig-grouplet-synth-d}]{
    \includegraphics[height=3.5cm,keepaspectratio]{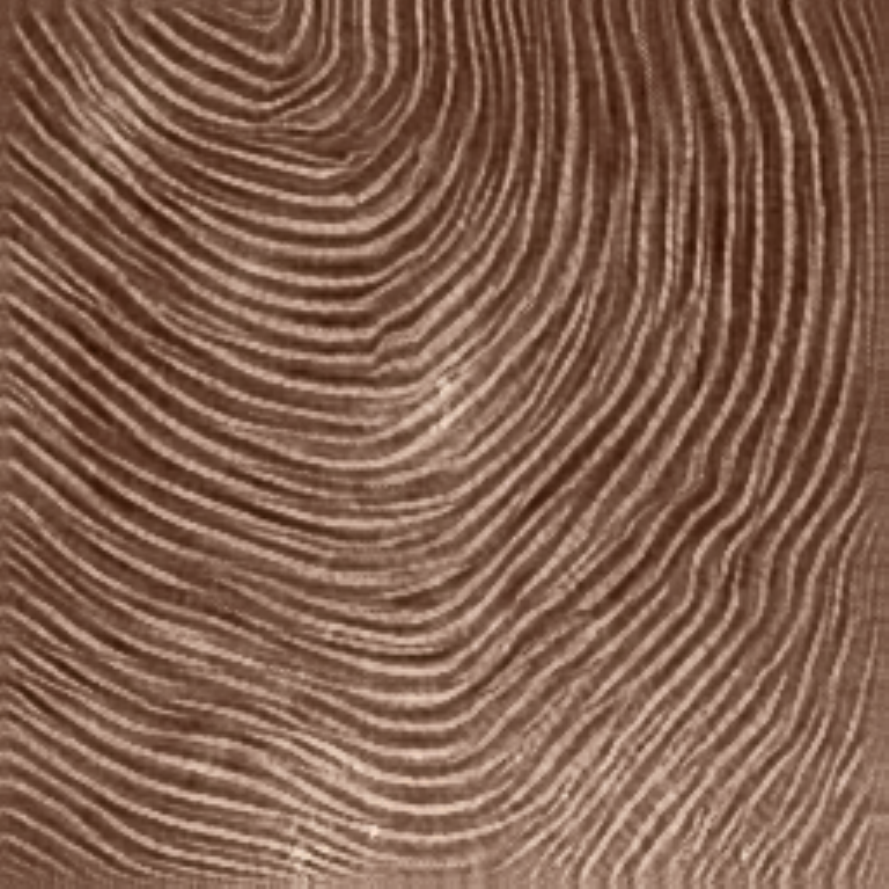}}
  \caption{Example of texture synthesis by statistical modeling of grouplet coefficients. (a) Exemplar. (b) Wavelet \cite{Portilla_J_2000_j-ijcv_par_tmbjscwc}. (c) Quilting \cite{Efros_A_2001_p-acm-siggraph_ima_qtst}. (d) Grouplets \cite{Peyre_G_2009_j-ieee-tpami_tex_pg}.}
  \label{fig-grouplet-synth}
\end{figure}

Grouplets defined over each scale of the wavelet transform have been
used to perform image denoising, super-resolution
\cite{Mallat_S_2009_acha_geo_g} and inpainting~\cite{Peyre_G_2009_j-ieee-tpami_tex_pg}
by solving a $\lun_N$ regularization similar to
\eqref{eq-l1-ip-recovery}. Grouplets can also be used to solve
computer graphics problems such as texture synthesis. Classical
approaches to texture synthesis use statistical models over a fixed
representation such as a wavelet basis, see for
instance~\cite{Heeger_D_1995_p-acm-siggraph_pyr_btas,Portilla_J_2000_j-ijcv_par_tmbjscwc}. Building
similar statistical models over a grouplet
basis~\cite{Peyre_G_2009_j-ieee-tpami_tex_pg} allows one to better synthesize the
geometry of some textures, and gives results similar to state of the
art computer graphics approaches such as texture
quilting~\cite{Efros_A_2001_p-acm-siggraph_ima_qtst}. Furthermore, the
explicit parameterization of the geometry though the association
fields $\la$ allows the user to modify this geometry and synthesize
dynamic textures. A comparison of these different approaches on one
texture synthesis example is given in Fig.~\ref{fig-grouplet-synth}.

\section{Transformations on Non-Euclidean Geometries}
\label{sec:noneuclidian}
	
In this section we describe how the concepts of \emph{frequency},
\emph{scale} and even \emph{directionality} have been extended to the
processing of data on non-euclidean geometries like the sphere and
other manifolds.

\subsection{Data Processing on the Sphere}
\label{sec:processing-sphere}

The unit sphere $S^2=\{\bs x\in\Rbb^3: \|\bs x\|=1\}\subset \Rbb^3$ is one of the
most natural non-Euclidean spaces. Very early, possibly due to
 influences for astronomy and geosciences, many data processing techniques
have been developed for this surface. Many filtering, multiscale,
directional and hierarchical methods have been designed, either in the
spherical frequency domain induced by the spherical harmonics basis ---
often following  the spirit of some Euclidean techniques exposed in
the previous sections --- or on the sphere itself thanks to some
geometrical tools such as the stereographic dilation or the lifting schemes
for wavelet analysis.

\subsubsection{Filtering}
As for the plane, filtering operations may be defined on $S^2$. Given
the common two-angle spherical parameterization $\bs
\xi=(\theta,\varphi)\in S^2$ with the co-latitude
$\theta\in[0,\pi]$ and longitude $\varphi\in[0,2\pi)$, this operation
is realized through spherical convolution evaluated on ${\rm SO}(3)$ (the
group of rotations in $\Rbb^3$). For a function $f\in \LL^2(S^2)=\{g:
\|g\|^2_2=\int_{S^2} |g|^2 <\infty\}$ and a
filter $h\in \LL^2(S^2)$, the convolution is
$$
(f \star h)(\bs\rho)\ =\ \int_{S^2}\ h(\bs\rho\,\bs\xi) f(\bs\xi)\
\ud\mu(\bs\xi),
$$
where $\bs\rho\in {\rm SO}(3)$ is a rotation (driven by three angles)
applied to the point $\bs\xi\in S^2$ and
$\ud\mu(\bs\xi)=\sin\theta\ud\theta\ud\varphi$. For an axisymmetric
filter, \ie  if $h(\bs \xi)=h(\theta)$, the convolution reduces to $ (f
\ast h)(\bs\xi')\ =\ \int_{S^2}\ h(\bs\xi'\cdot\bs\xi)
f(\bs\xi)\ \ud\mu(\bs\xi)$, where $\bs\xi'\cdot\bs\xi$ is
the common 3-D scalar product between $\bs\xi'$ and $\bs\xi$
seen as unit vectors.

\subsubsection{Fourier Transform} 

The Fourier transform of a function $f\in \LL^2(S^2)$ is defined by
$$
\hat{f}_{\ell m}=\scp{Y_{\ell m}}{f}=\int_{S^2}Y_{\ell
  m}^*(\bs\xi)\,f(\bs\xi)\ \ud\mu(\bs\xi), \quad f(\bs \xi)=\sum_{\ell,m} \hat{f}_{\ell m}\,Y_{\ell
  m}(\bs \xi)
$$ 
with respect to orthonormal basis of \emph{spherical harmonics}
$\mathcal{Y}=\{Y_{\ell m}(\bs\xi):\ell\geq 0, |m|\leq\ell\}$, \ie the
eigenvectors of the spherical Laplacian
\cite{Healy_D_2003_j-four-anal-appl_fft_2siv}. 

The frequency content of $f$ is thus represented by the value of $\hat
f_{\ell m}$ on the order $\ell\in\Nbb$, which basically counts the
number of oscillations on the latitudes, and the moment
$m\in\{-\ell,\cdots,\ell\}$ counting longitude
oscillations. Numerically, only certain discretizations of the sphere
can provide perfect quadrature formulae to compute the Fourier
coefficients of band-limited functions on the sphere, sometimes with
very efficient algorithms
\cite{Healy_D_2003_j-four-anal-appl_fft_2siv,Driscoll_J_1994_j-adv-appl-math_com_ftc2s}.

\subsubsection{Spherical Scale-Space} Similarly to what happened for
signals or images, the first notion of ``scale'' on the sphere was
imported from the Heat Dynamic that is also known on this space. In
that framework, if a spherical function $f\in \LL^2(S^2)$ is considered
 the initial heat configuration, the spherical heat dynamics smooth
it with time $\tau>0$, conferring a scaling notion on  this parameter.

Interestingly, as for Euclidean spaces, the solution at time $\tau>0$
of the heat equation initialized to some function $f\in\LL^2(S^2)$ is
simply $f(\bs \xi,\tau) = \sum_{\ell,m} \hat f_{\ell m}(\tau)
Y_{\ell m}(\bs \xi)$, with $\hat f_{\ell m}(\tau) = \hat f_{\ell m}
\,e^{-\ell(\ell+1)\,\tau}$ and $f(\bs
\xi,0)=f(\bs\xi)$. Alternatively, since for an axisymmetric filter
$h$ we have the spherical convolution theorem
$$
\widehat{(f\ast h)}_{\ell m}\ =\ \textstyle\sqrt{\tfrac{4\pi}{2\ell+1}}\,\hat{f}_{\ell
  m}\,\hat{h}_{l0},
$$
the solution of the Heat Equation can also be obtained by a
convolution by a specific kernel $G^\circ_\tau(\bs \xi)$, coined
spherical Gaussian of width ${\sqrt \tau}$. It is defined in frequency
by $\widehat{(G^\circ_\tau)}_{\ell m} = \sqrt{{(2\ell+1)}/{4\pi}}
\,e^{-\ell(\ell+1)\,\tau}$.

The link between the heat dynamics and the spherical convolution with
the axisymmetric filter $G^\circ_\tau$ has been exploited by
 B\"ulow \cite{Bulow_T_2002_p-dagm_mul_ips} to develop several specific spherical filters for feature
detection, such as the Laplacian of Gaussian or the directional derivative
of Gaussian.

\subsubsection{Spectral Wavelets}
\label{sec:spectral-wavelets}

Freeden \etal
\cite{Freeden_W_1996_j-adv-appl-math_sph_wtd,Freeden_W_2003_j-rev-mat-complutense_sur_wmga}
have fully exploited the connection between convolution and frequency
filtering on the sphere to develop a continuous wavelet transform on
the sphere. This is done by introducing a family of axisymmetric
functions $\psi_a(\bs\xi)$, coined spherical wavelet, continuously
indexed by $a>0$, and such that $\int_{\Rbb_+} |\widehat{(\psi_a)}_{\ell 0}|^2\ \ud a/a = 1$,
$\widehat{(\psi_a)}_{00}=0$, plus additional regularity
conditions. The wavelet coefficients of a function $f\in \LL^2(S^2)$
are then defined as $W_f(a,\bs\xi)=(f\ast \psi_a)(\bs\xi)$. The
reconstruction is possible (almost everywhere) by 
$$
f(\bs\xi')\ =\ \langle f\rangle\ +\ \int_{\Rbb_+}\int_{S^2}\
W_f(a,\bs\xi)\,\psi_{a}(\bs\xi'\cdot\bs\xi)\ \tfrac{\ud a}{a}\,\ud\bs\xi,
$$ 
with $\langle f\rangle=\tinv{4\pi}\int_{S^2}f(\bs\xi)\,\ud\mu(\bs\xi)$.

In
\cite{Freeden_W_1996_j-adv-appl-math_sph_wtd,Freeden_W_2003_j-rev-mat-complutense_sur_wmga},
an MRA  on the sphere is also built by
defining Quadrature Mirror Filters in the frequency domain. A spatial
sub-sampling of the different subspaces of the MRA can also decrease the
redundancy of the basis hence created.

Following a similar approach,  (isotropic) needlet  frames
introduced in
\cite{Narcowich_F_2006_j-siam-math-anal_loc_tfs,Kerkyacharian_G_2007_j-elec-j-stat_nee_aeip,Guilloux_F_2009_j-acha_pra_wds}
represent another example of spectral wavelets, \ie  wavelets shaped in
the Fourier domain. Needlets additionaly offer relationships with 
quadrature formulae used to turn integrals of bandlimited functions
into discrete summations.

\subsubsection{Stereographic Wavelets}
\label{sec:ster-wavel-fram}

In the previous sections, the notion of scale in the processing of
spherical data was always defined in the frequency domain, \ie  by
dilating the frequency domain by a parameter, preventing a fine
control of the spatial support of the filter.  

\begin{figure}
  \centering
  \parbox[]{5cm}{\subfigure[]{\includegraphics[height=5cm,keepaspectratio]{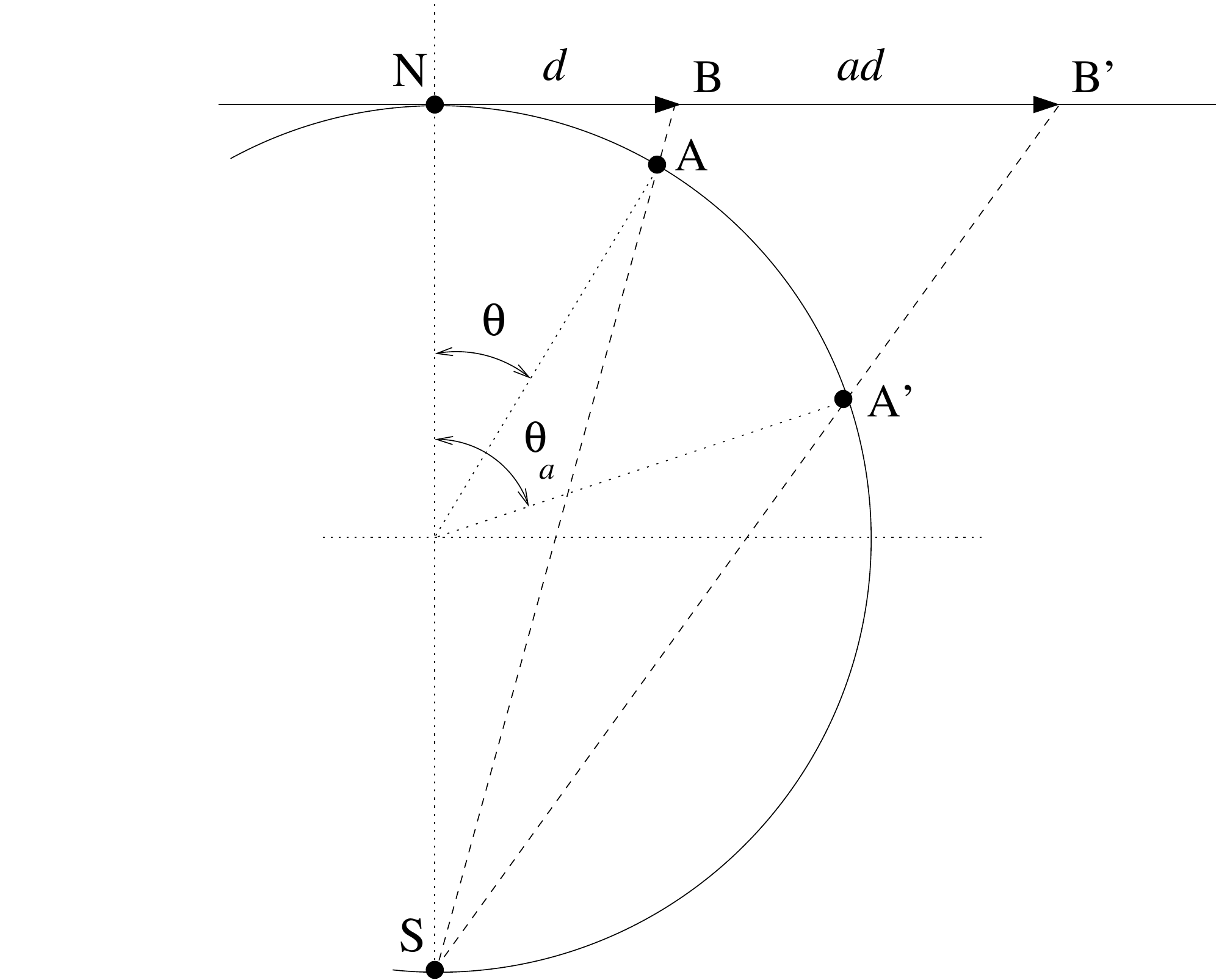}}}\hspace{2cm}
  \parbox[]{6cm}{
  \subfigure[]{\includegraphics[height=2.5cm,keepaspectratio]{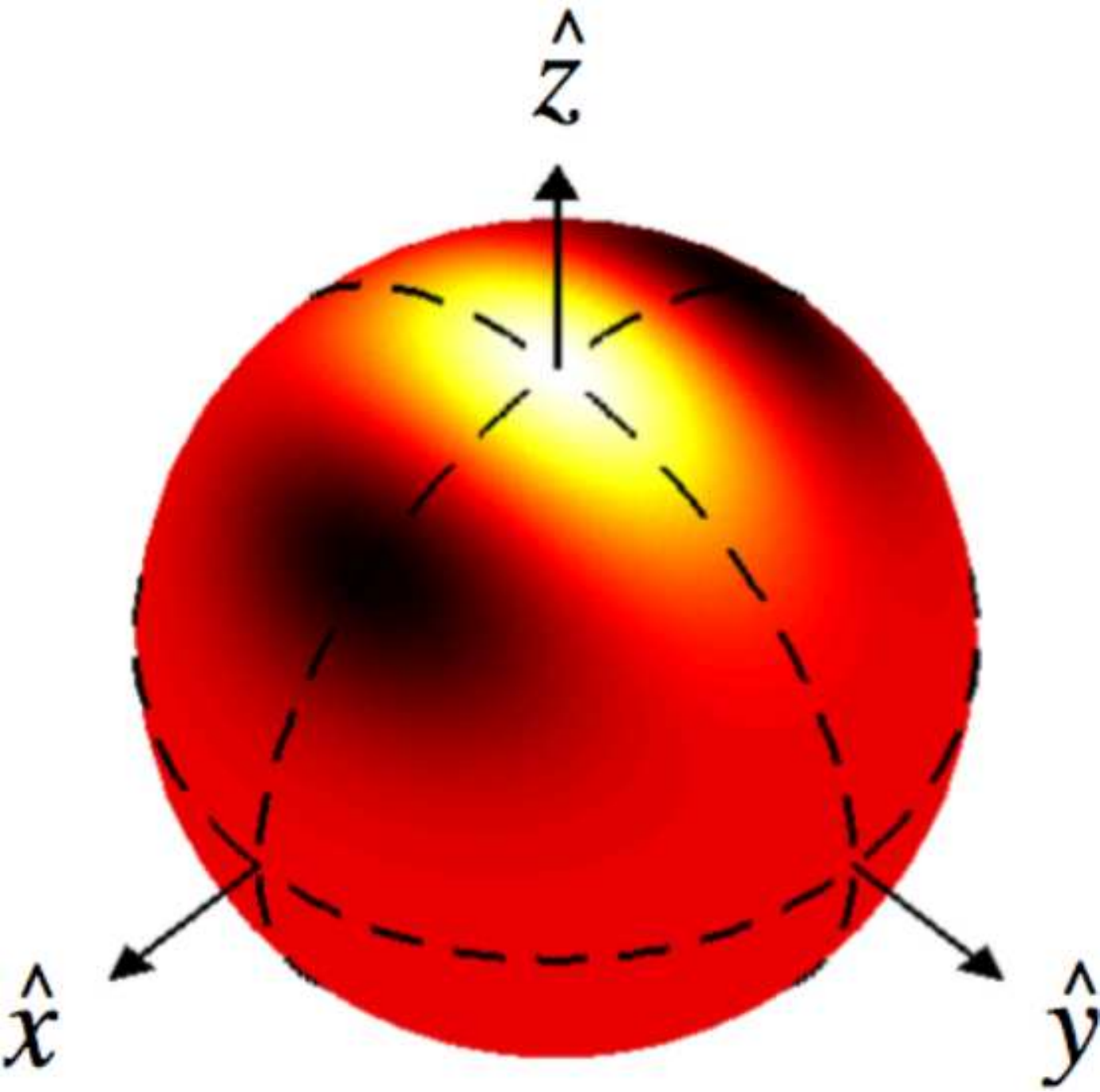}}\hspace{5mm} 
  \subfigure[]{\includegraphics[height=2.5cm,keepaspectratio]{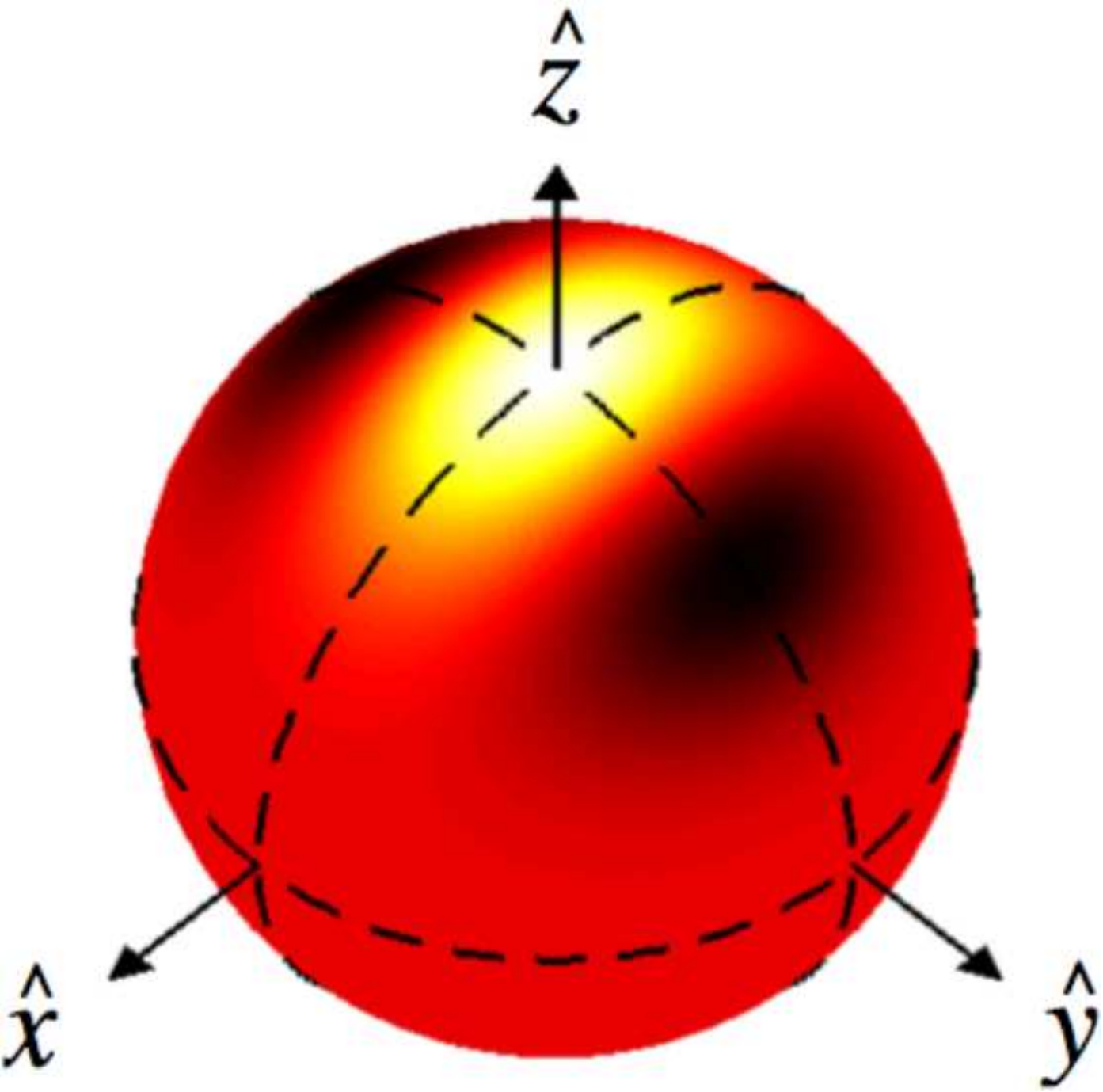}}\\
  \subfigure[]{\includegraphics[height=2.5cm,keepaspectratio]{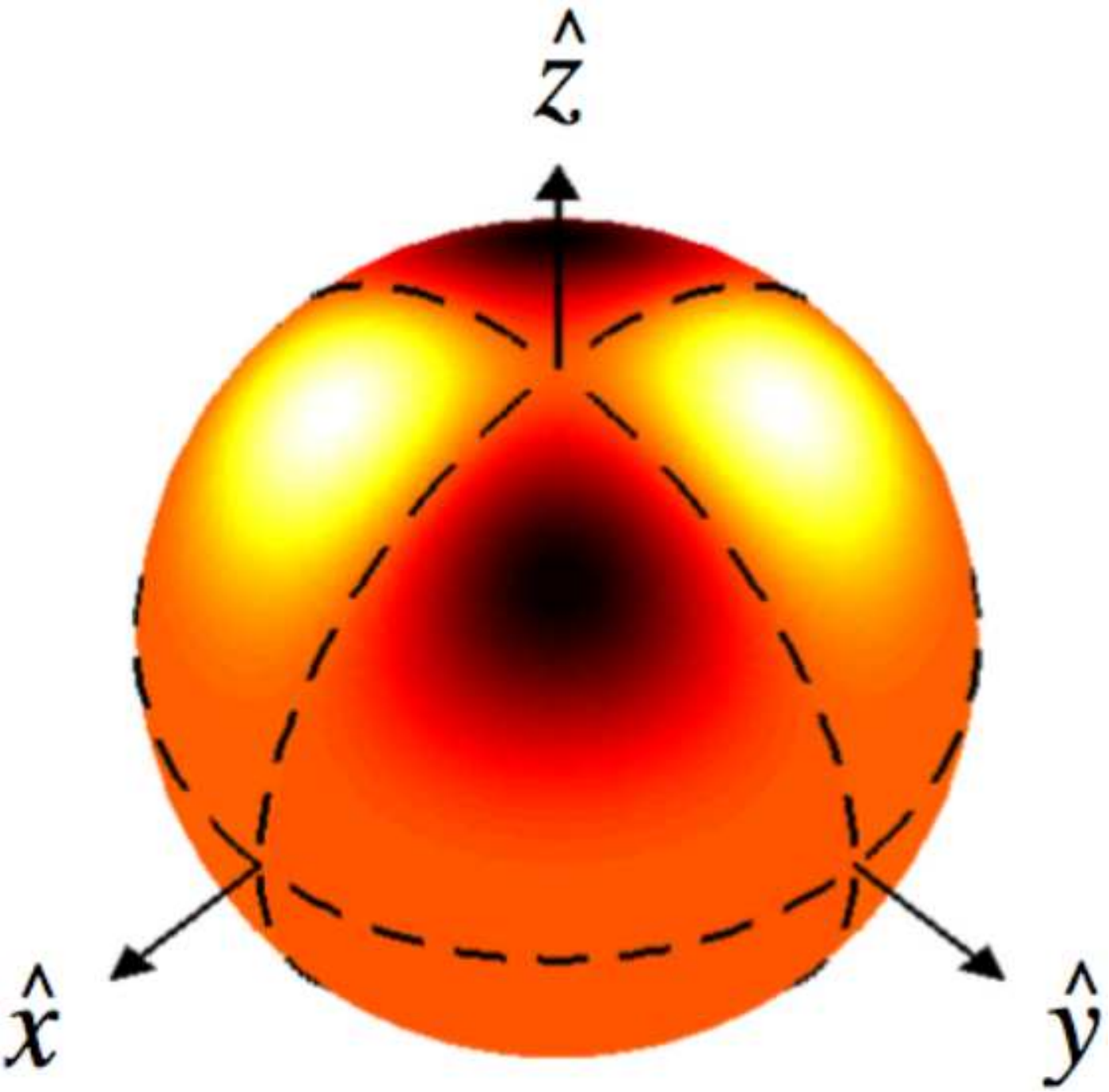}}\hspace{5mm} 
  \subfigure[]{\includegraphics[height=2.5cm,keepaspectratio]{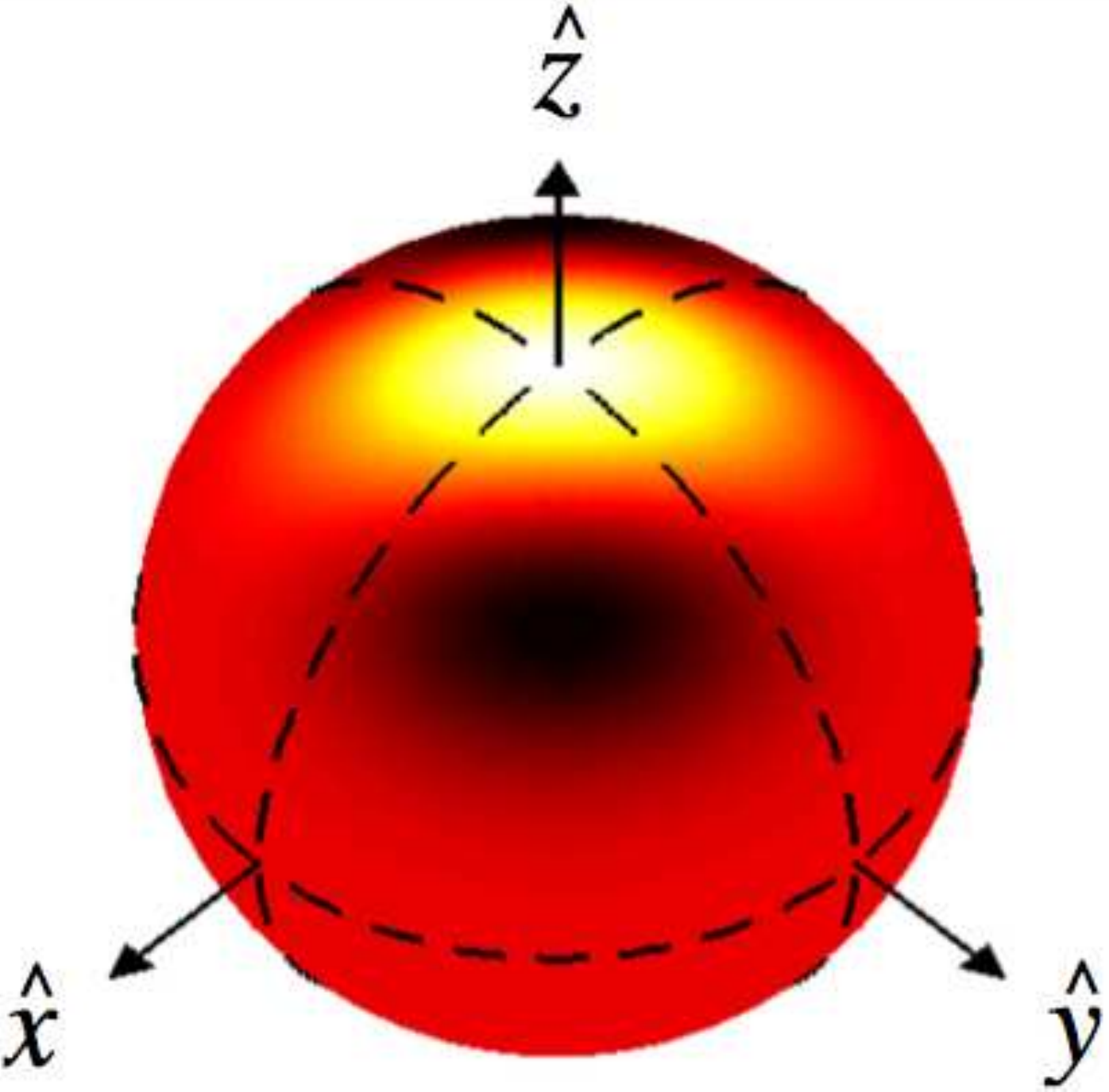}}
 }
  \caption{(a) Stereographic dilation on $S^2$. On the right, the
  (steerable) second
    directional derivative of Gaussian. The three images (b)-(d) are
    the basis elements, while the fourth in (e) is a
    linear combination of the  first three yielding a rotation of $\pi/4$ around the North pole.
  \label{fig:stereo_dil_expl}
}
\end{figure}

An alternative approach introduced by Antoine and Vandergheynst
\cite{Antoine_J_1999_j-acha_wav_2sgta,Antoine_J_2002_j-acha_wav_sia}
defines the dilation directly in the spatial domain. The compactness
of $S^2$ is respected, by introducing a \emph{stereographic} dilation.
As illustrated on Fig.~\ref{fig:stereo_dil_expl}-(a) for point dilation,
the stereographic dilation $D_a$ of a function $g \in \LL^2(S^2)$
amounts to projecting $g$ on the plane tangent at the North Pole by
the \emph{stereographic projection} $\Pi$, to applying there a
Euclidean dilation $d_a$ by a scale $a>0$, and to lifting the
resulting function back to the sphere by $\Pi^{-1}$
\cite{Wiaux_Y_2005_j-astrophys-j_cor_pbsew}. Mathematically, $[D_a
g](\theta,\varphi) = \lambda(a,\theta)\,g(\theta_{1/a},\varphi)$, with
$\tan \theta_{\alpha}/2 = \alpha \tan\theta/2$ and where $\lambda$ is
a normalizing function such that $\|D_ag\|_2=\|g\|$.

Given a mother wavelet $\psi\in \LL^2(S^2)$ centered on the North
pole, the proposed approach considers the joint action of
\emph{translations}, \ie  rotation operators $R_{\bs \rho}$ in ${\rm SO}(3)$, and
of the dilations $D_a$ on $\psi$. The wavelet transform of $f$ is therefore: 
 
$$
W_{f}(\bs \rho,a)\ =\ \scp{\psi_{(\bs \rho,a)}}{f},\quad \bs \rho\in {\rm SO}(3),\ a>0,
$$
with $\psi_{(\bs \rho,a)}=R_{\bs \rho}D_a\psi$. If the wavelet is
\emph{admissible}, which is nearly equivalent to impose
$\int_{S^2}\,\ud\mu(\theta,\varphi)\
\tfrac{\psi(\theta,\varphi)}{1+\cos\theta} = 0$, the reconstruction of
$f$ is possible through
$$
f(\bs \xi)\ =\ \langle f\rangle\ +\ \int_{\Rbb_{+}^*}\int_{{\rm SO}(3)} \tfrac{\ud a
  \ud\nu(\bs \rho)}{a^3}\ W_f(\bs \rho,
a)\,[R_{\bs \rho}L_{\psi}^{-1}D_a\psi](\bs \xi),
$$
where $\nu$ is the Lebesgue measure on ${\rm SO}(3)$ and $L_{\psi}$ is a
multiplicative operator function of $\psi$ only and expressed in the
Fourier domain \cite{Antoine_J_1999_j-acha_wav_2sgta}. For
axisymmetric wavelets, this result simplifies by the fact that the action
of $R_{\bs \rho}$ on $\psi$ is controlled by  two angles only.

Many wavelets may be defined on the sphere since it has been proved in
\cite{Wiaux_Y_2005_j-astrophys-j_cor_pbsew} that any admissible wavelet
on the plane $\LL^2(\Rbb^2)$ can be imported by inverse stereographic
projection $\Pi^{-1}$. A Laplacian of Gaussian (LoG), difference of
Gaussians (DoG), Morlet Wavelet, and many other are generally used
\cite{Antoine_J_1999_j-acha_wav_2sgta,Bogdanova_I_2005_j-acha_ste_wfs,Demanet_L_2003_p-spie-wasip_gab_ws}. Numerically, this spherical
CWT is obtained thanks to the convolution theorem mentionned previously. This transform has been for instance
intensively used in the analysis of the Cosmic Microwave Background
(CMB), an astronomical signal remnant of some specific evolution phase
of the Big Bang \cite{Cayon_L_2000_j-mon-not-roy-astron-soc_iso_wptepscmbm,Abrial_P_2007_j-four-anal-appl_mor_caisapa,Wiaux_Y_2008_j-mon-not-roy-astron-soc_non_galmmwmapd}.

Wavelet frames can be developed in this theory by discretizing the
scaling parameter $a$
\cite{Bogdanova_I_2005_j-acha_ste_wfs}. These frames, that do
not subsample the spherical positions, have successfully served for the
construction of invertible filter banks on the 2-Sphere
\cite{Yeo_B_2008_tip_con_ifb2s} even if the stereographic dilation is
not really compatible with the frequency description of the wavelets.

\subsubsection{Haar Transform on the Sphere}

The constructions of spherical wavelets described in the previous section make use of the Fourier decomposition on the sphere. It is possible to define wavelets directly over the spherical domain without Fourier analysis, using for instance the lifting scheme method \cite{Schroder_P_1995_p-acm-siggraph_sph_werfs}, see Sec.~\ref{sec:lift-scheme-wavel}. This allows one to define spherical wavelets with a compact support, although the stability of the resulting transform is more difficult to control than over the planar domain.

Inspired by this lifting scheme \cite{Schroder_P_1995_p-acm-siggraph_sph_werfs}, one can easily define a Haar basis on the sphere by considering a family $\{ \Mm^j \}_{J \leq j \leq 0}$ of embedded spherical triangulations that approximate a sphere \cite{Lessig_C_2008_j-acm-tog_soh_oshws}. These triangulations are obtained by a
regular 1:4 refinement rule starting from an initial regular polyhedron $\Mm^0$, and the edges are projected on the sphere to define spherical triangles. 

\begin{figure}[htb!]
  \centering
  \begin{tabular}{@{}c@{\hspace{1mm}}c@{\hspace{1mm}}c@{\hspace{1mm}}c@{}}
		\includegraphics[width=.24\linewidth]{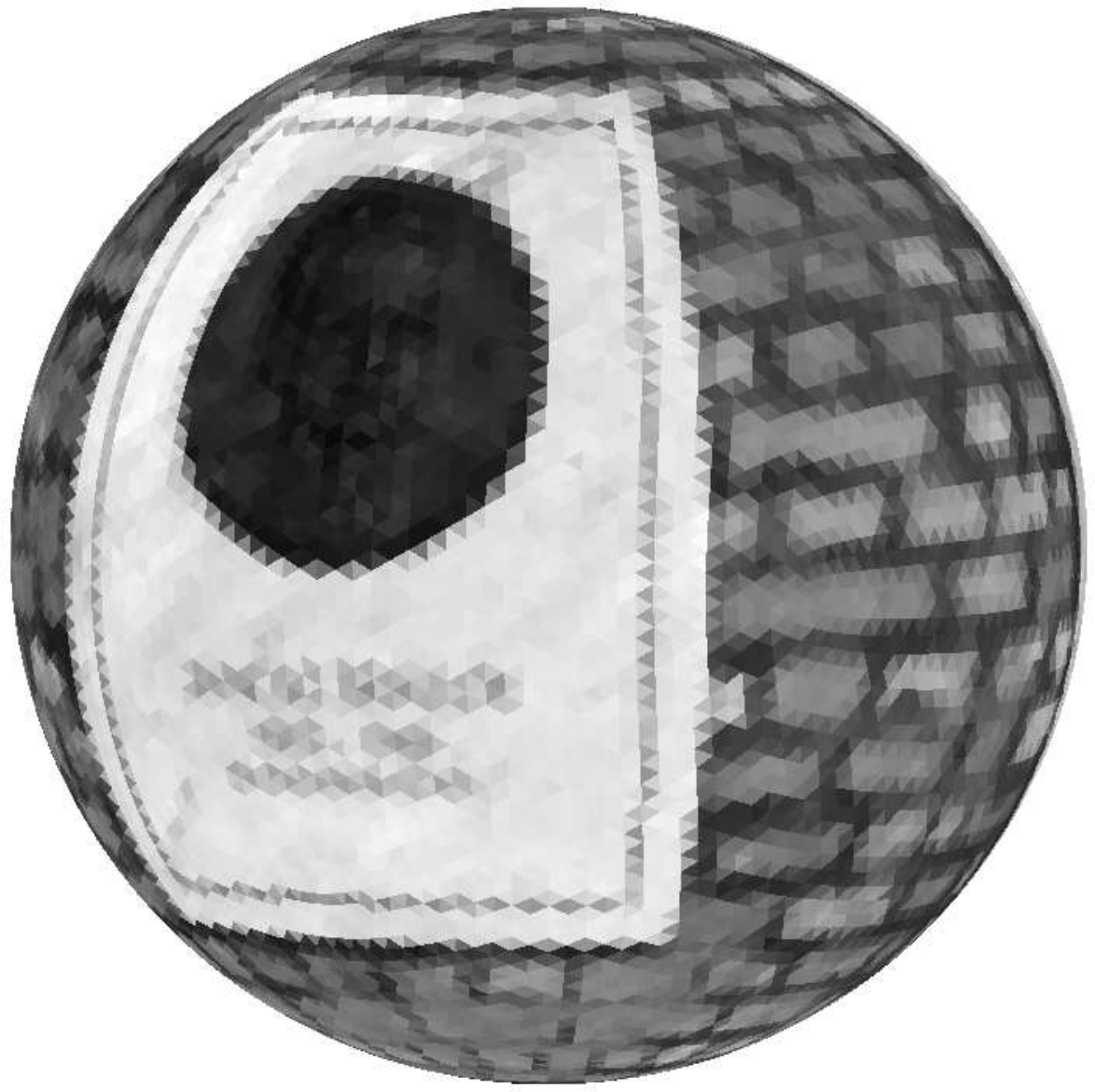}&		\includegraphics[width=.24\linewidth]{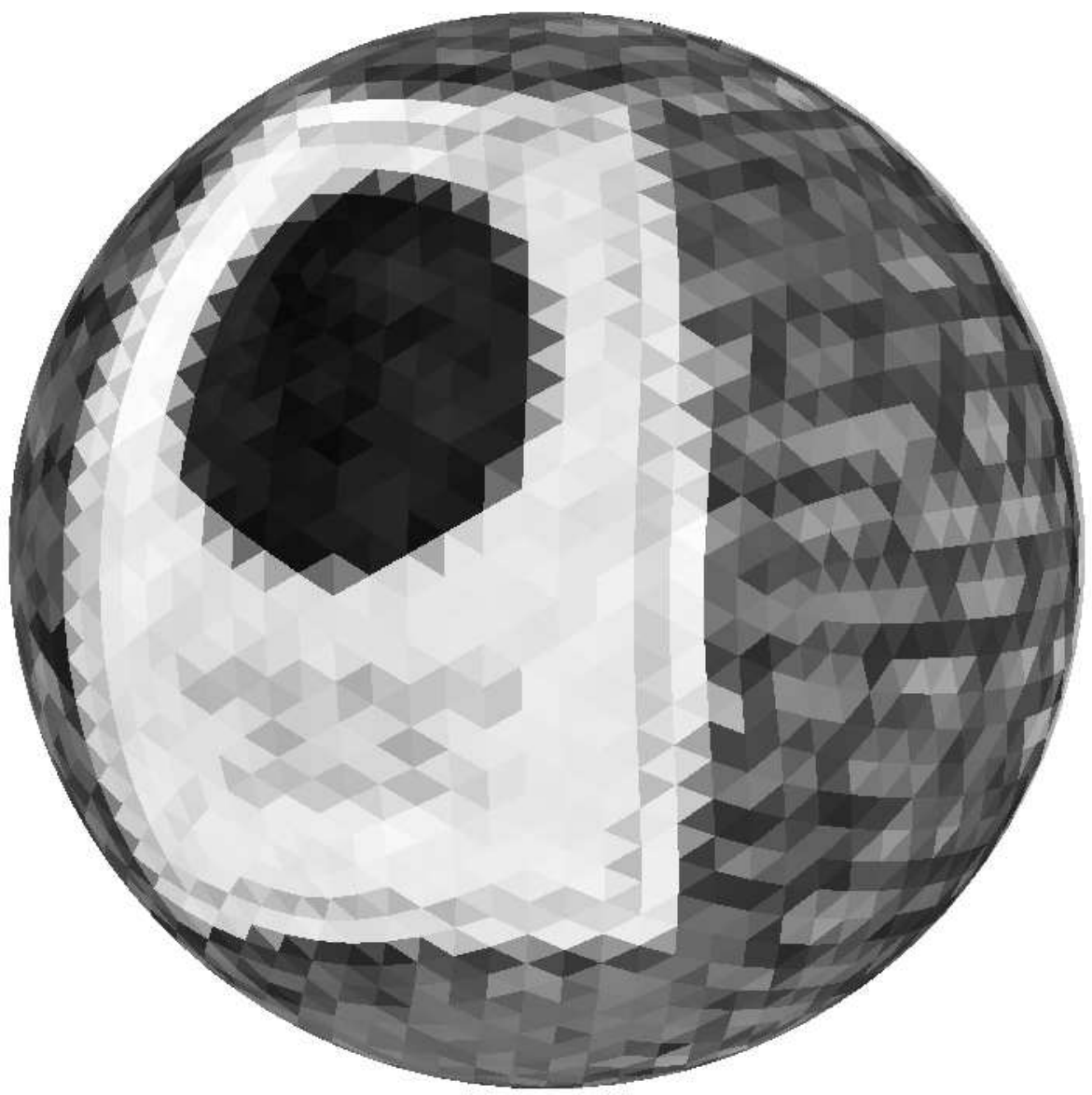}&		\includegraphics[width=.24\linewidth]{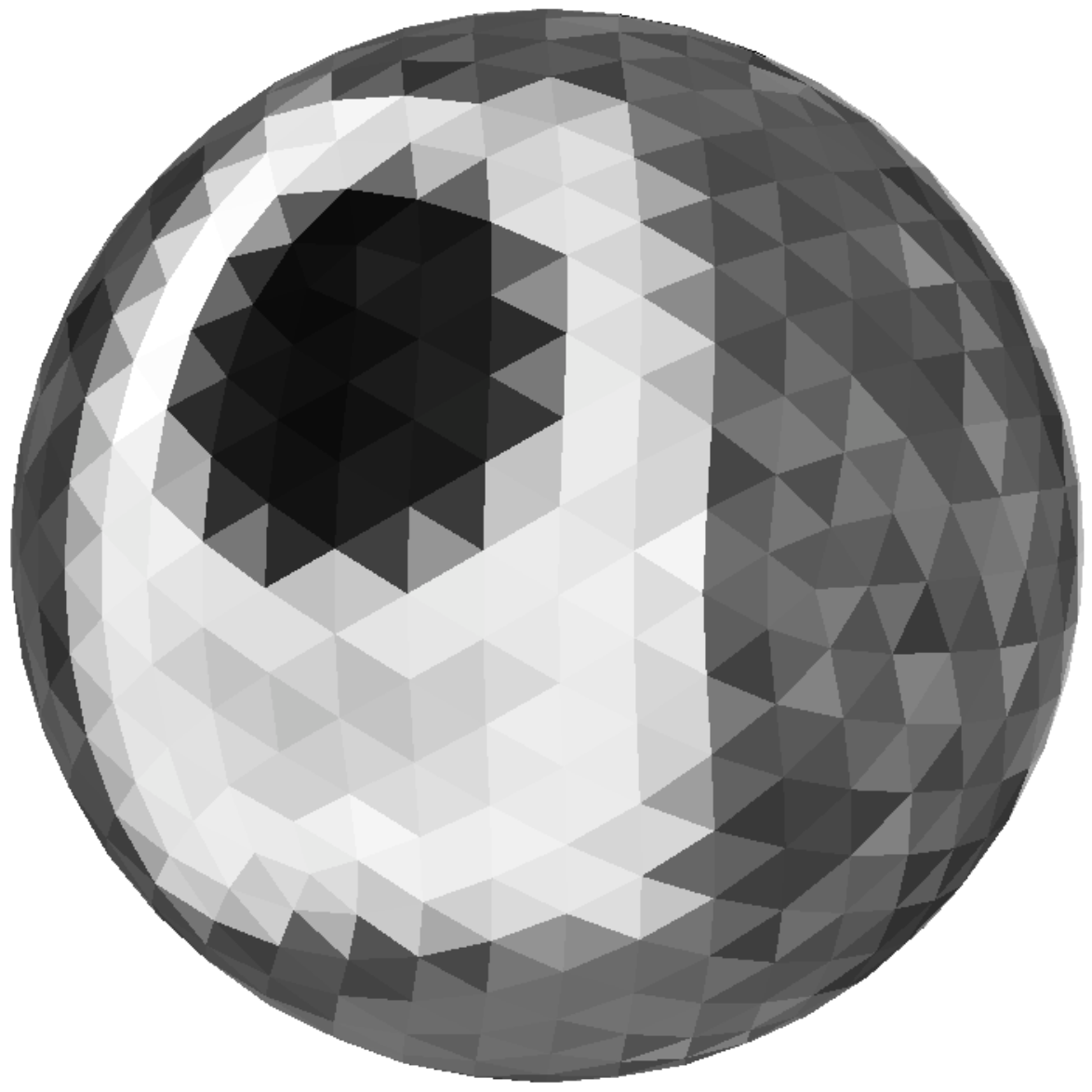}&		\includegraphics[width=.24\linewidth]{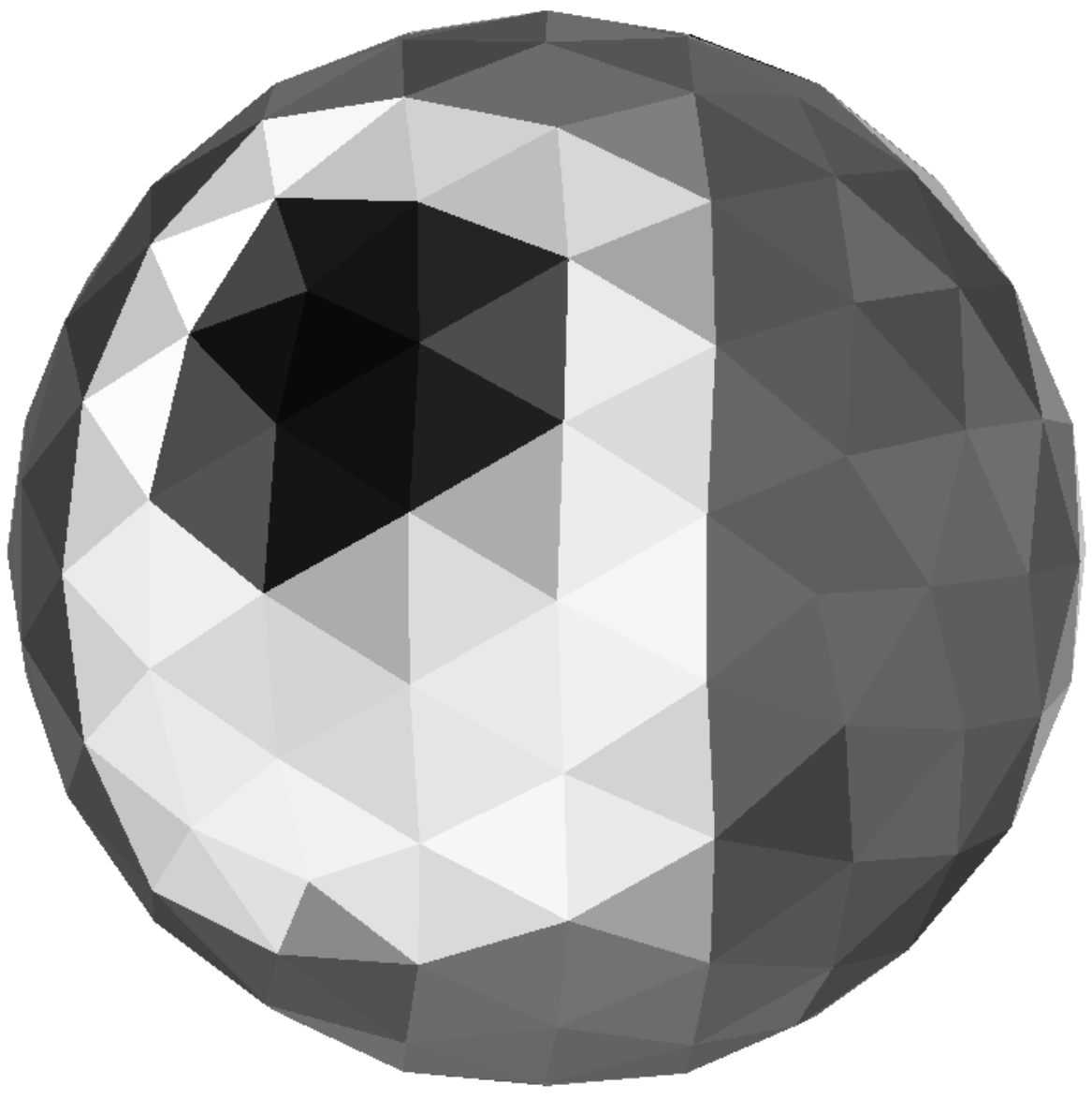}\\
		$j=5$ & $j=4$ & $j=3$ & $j=2$
	\end{tabular}
	\caption{Projection on the spherical Haar multiresolution.\label{fig-haar-sphere-linear}}
\end{figure}

The corresponding spherical multiresolution defines $V_j \subset \LL^2(S^2)$ as the set of functions that are constant on each triangle of  $\Mm^j$. Figure \ref{fig-haar-sphere-linear} shows the linear projection of a spherical function on some of these multiresolution spaces.

\begin{figure}[htb!]
  \centering
  \begin{tabular}{@{}c@{\hspace{1mm}}c@{\hspace{1mm}}c@{\hspace{1mm}}c@{}}
		\includegraphics[width=.24\linewidth]{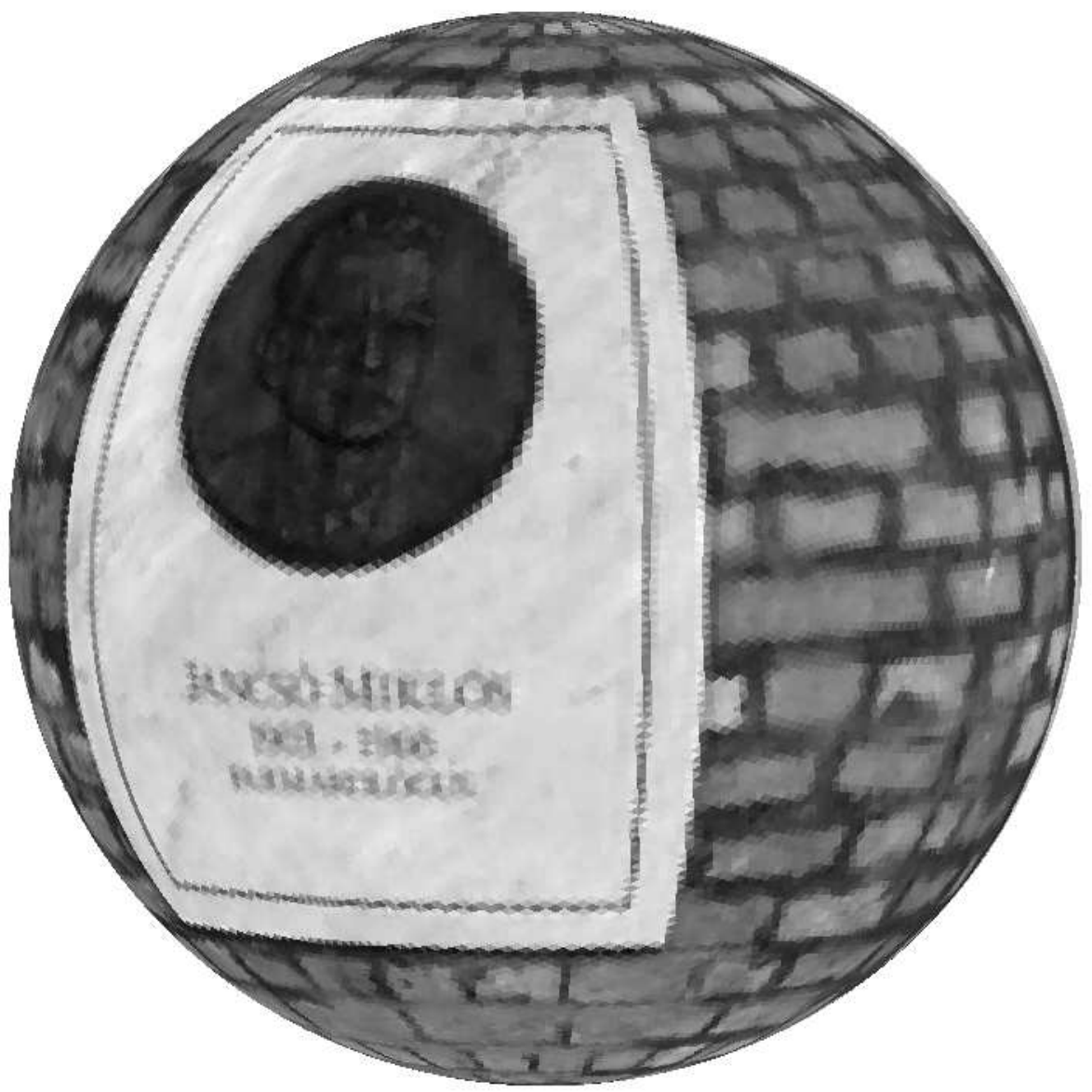}&		\includegraphics[width=.24\linewidth]{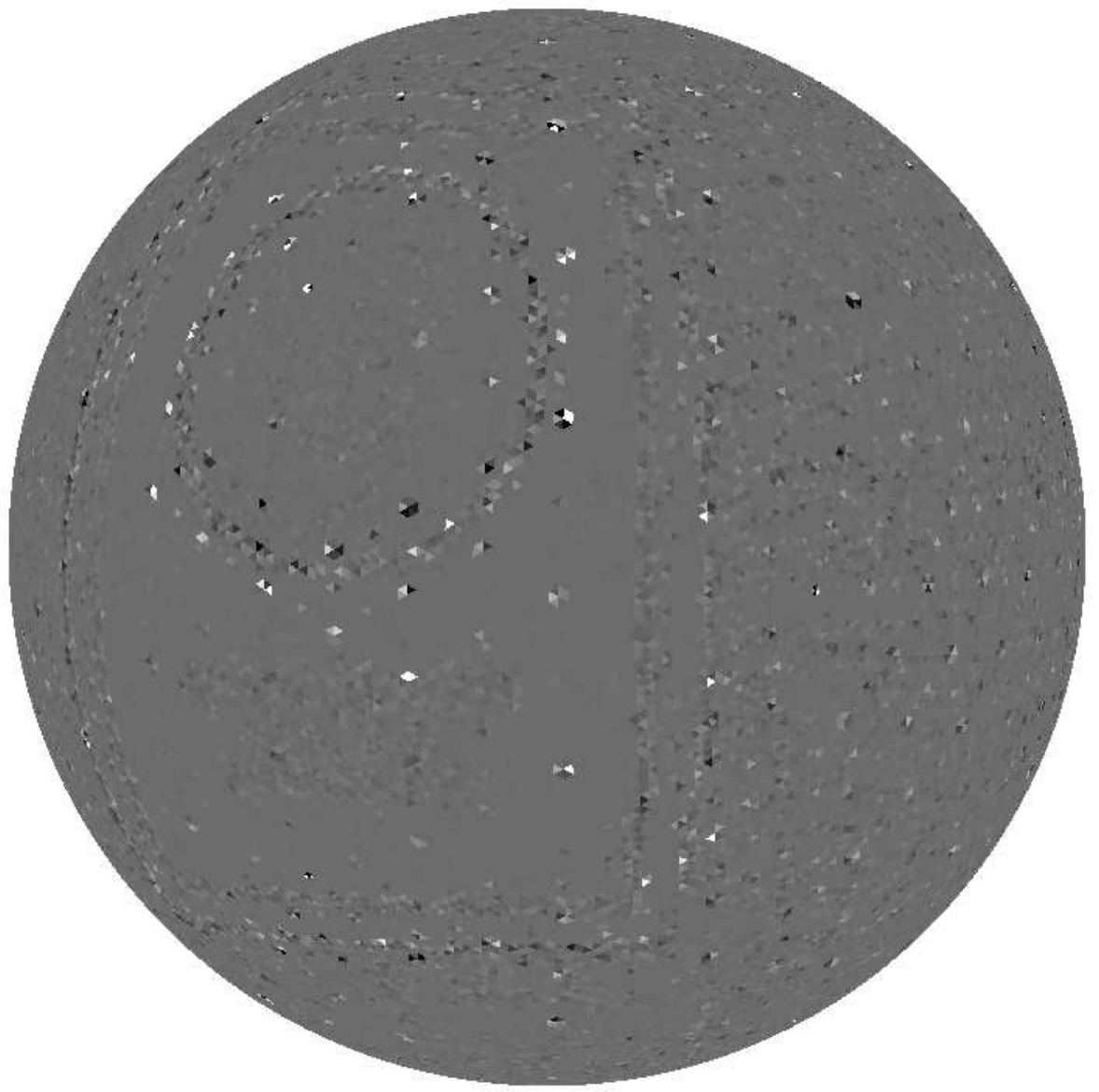}&
		\includegraphics[width=.24\linewidth]{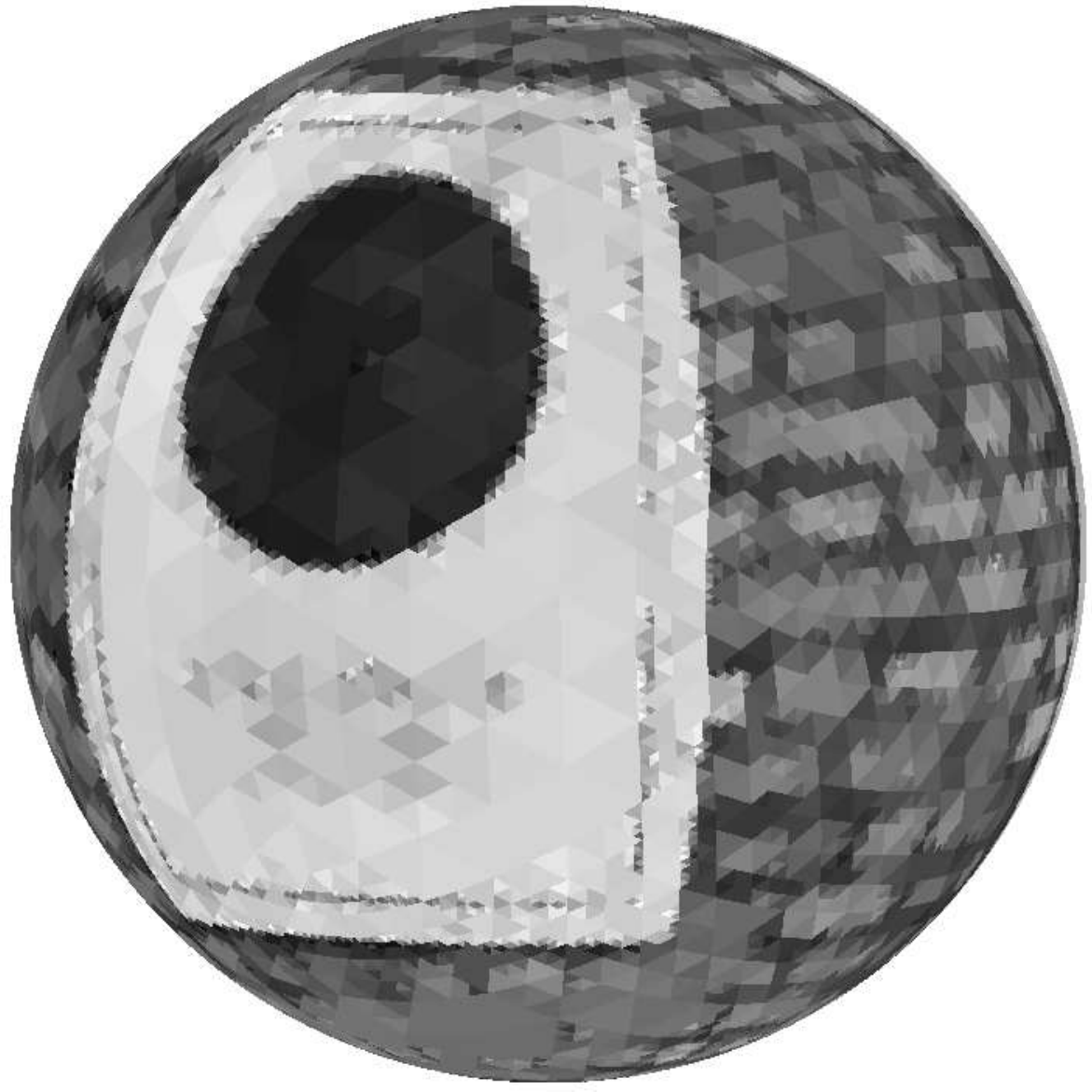}&		\includegraphics[width=.24\linewidth]{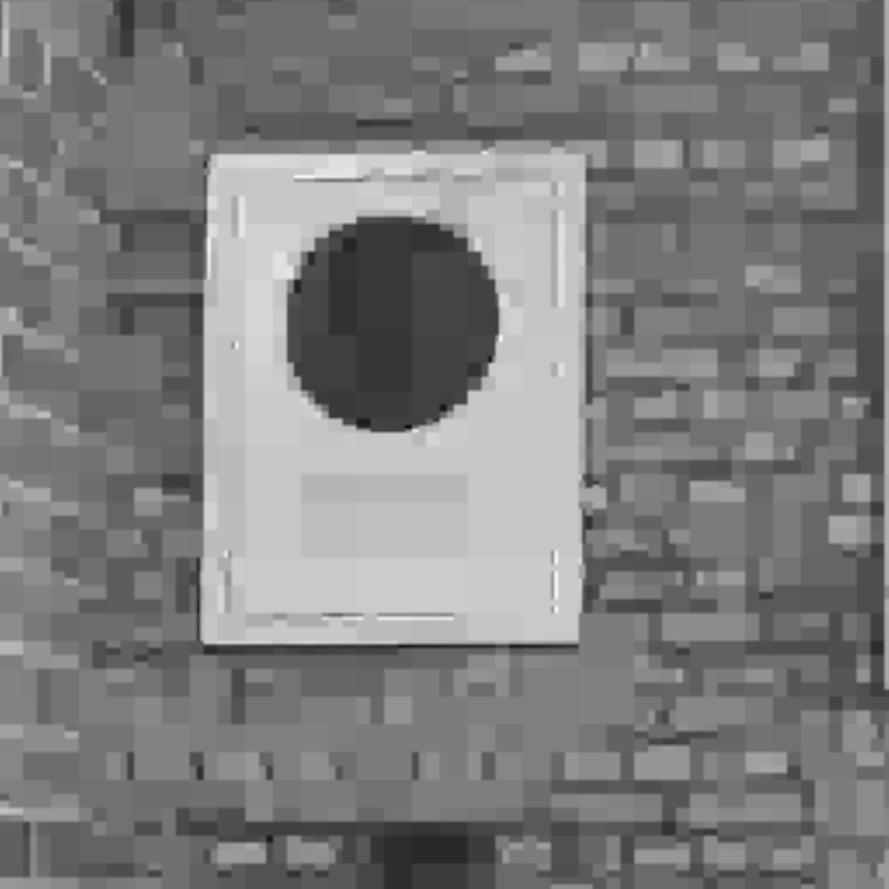}\\
		$f$ & $\{\dotp{\psi_{j,n}}{f}\}_{j,n}$ & $\ThreshH_T(f)$ (spherical) & $\ThreshH_T(f)$ (planar)
	\end{tabular}
	\caption{Comparison of spherical and planar Haar approximations. The threshold $T$ is 
		adjusted so that $\ThreshH_T(f)$ is an approximation with a number of coefficients equal to 
		$5\%$ of the number of pixels in 
		the high resolution planar image.\label{fig-haar-sphere-nonlinear}}
\end{figure}

Following the usual definition (Sec. \ref{sec:introtools-discss-mra}),
a Haar wavelet basis $\{\psi_{j,n}\}_n$ is an orthogonal basis of the
detail space $W_j$ such that $V_{j+1} = V_j \oplus W_j$. The wavelet
coefficients $\dotp{\psi_{j,n}}{f}$ are computed using a pyramid
algorithm that mimics the usual Haar transform, except that for each
triangle, one gathers three detail coefficients and one coarse scale
coefficient. Figure \ref{fig-haar-sphere-nonlinear} shows these Haar
coefficients together with a comparison between spherical and planar
non-linear approximations $\ThreshH_T(f)$.

\subsubsection{Steerable Wavelets on the Sphere}
\label{sec:steer-wavel-sphere}

Finally, the sphere is compatible with the definition of steerable
filters similarly to those defined in Sec.~\ref{sec:steerability}
for the plane. In particular, using the stereographic projection $\Pi$
introduced in the previous section, steerability on the sphere is also
imported from the plane. This fact has been used in
\cite{Wiaux_Y_2005_j-astrophys-j_cor_pbsew,Wiaux_Y_2006_j-astrophys-j_fas_dcssf,Vandergheynst_P_2010_incoll_wav_s}
to define differential and steerable filters useful to detect
directional features in the Cosmic Microwave Background. An example of
a steerable wavelet is given in Fig.~\ref{fig:stereo_dil_expl}(b-e).
Spherical steerability may also be directly studied in the frequency
domain with spectral dilation
\cite{Wiaux_Y_2008_j-mon-not-roy-astron-soc_exa_rdws}.

\subsubsection{Other Constructions}

It is impossible to cite the vast literature on multiscale
decomposition on the sphere. Let us just quote some of them. Wavelets,
ridgelets and curvelets have been translated on the sphere by Starck
\etal \cite{Starck_J_2006_j-astron-astrophys_wav_rcs} by using a particular
spherical sampling, called HEALPix, locally similar to a square
discretization. Locally supported biorthogonal wavelet bases have been also realized thanks to
some radial projections of the planar faces of a cube on $S^2$ in \cite{Rosca_D_2007_j-four-anal-appl_wav_bsorp}.
 
\subsection{Wavelets on General 2-Manifolds}
\label{sec:wavelets-general-2}

Given a two-dimensional manifold $\mathcal{M}$, \ie  locally
isomorphic to $\Rbb^2$, authors in
\cite{Antoine_J_2010_j-acha_wav_tmona} describe how to define a
Continuous Wavelet Transform (CWT) for function $f:\mathcal{M}\to \Cbb$.

Similarly to the way the stereographic dilation is defined for
the sphere, the local dilation of a function $\psi$ around the point
$\bs \xi\in\mathcal{M}$ relies on the knowledge of a local and invertible
projection $\Pi_{\bs \xi}$ between $\mathcal{M}$ and its tangent plane
$T_{\bs \xi}\mathcal{M}$ on $\bs \xi$. The desired dilation of scale $a>0$
therefore factorizes  as $D_{(\bs\xi,a)}=\Pi_{\bs \xi}^{-1}d_a\Pi_{\bs \xi}$ with $d_a$ the
common Euclidean dilation of function in
$T_{\bs \xi}\mathcal{M}\simeq\Rbb^2$. 

Given the Hilbert space $\mathcal{H}=\LL^2(\mathcal{M},\ud\mu)$ of square integrable
function on $\mathcal{M}$, for a proper measure $\ud\mu$, the CWT of a
function $f$ on $\mathcal{M}$ is then formally defined by correlating
$f$ with a set of prototype wavelets $\psi_{(\bs \xi)} \in \mathcal{H}$
localized around any $\bs \xi\in\mathcal{M}$, \ie  
$$
W_f(\bs \xi,a)\ =\
\scp{\psi_{(\bs \xi,a)}}{f}_{\mathcal{H}}\ \triangleq\ \int_{\mathcal{M}}\,\ud\mu(\bs \xi')\
f(\bs \xi')\, \psi_{(\bs \xi,a)}(\bs \xi'),\qquad \psi_{(\bs
\xi,a)}=D_{(\bs\xi,a)
}\psi_{(\bs \xi)}.
$$
The theoretical invertibility of this transform has however to be
studied specifically in each case, \ie  given $\mathcal{M}$ and
$\Pi_{\bs \xi}$. Results exist for instance for the two-sheeted hyperboloid
and the paraboloid in $\Rbb^3$ \cite{Antoine_J_2008_j-ijwmip_con_wtcs}.

\subsection{Lifting Scheme Wavelets on Meshed Surfaces}
\label{sec:lift-scheme-wavel}

The lifting scheme of Sweldens \cite{Sweldens_W_1996_j-acha_lif_scdcbw}, described in
Sec.~\ref{sec-lifting}, can be used to define wavelets on
non-translation invariant geometries, including surfaces with
complicated topologies. Lifted wavelets on surfaces are usually built
on a semi-regular mesh grid, was first considered by Lounsbery \etal
\cite{Lounsbery_M_1997_j-acm-tog_mul_asatt}, and then refined within the lifting
framework by Schr{\"{o}}der and Sweldens
\cite{Schroder_P_1995_p-acm-siggraph_sph_werfs}.

\begin{figure}[htb!]
  \centering
  \subfigure[\label{fig-semiregular-4}]{
	{\includegraphics[height=5cm,keepaspectratio]{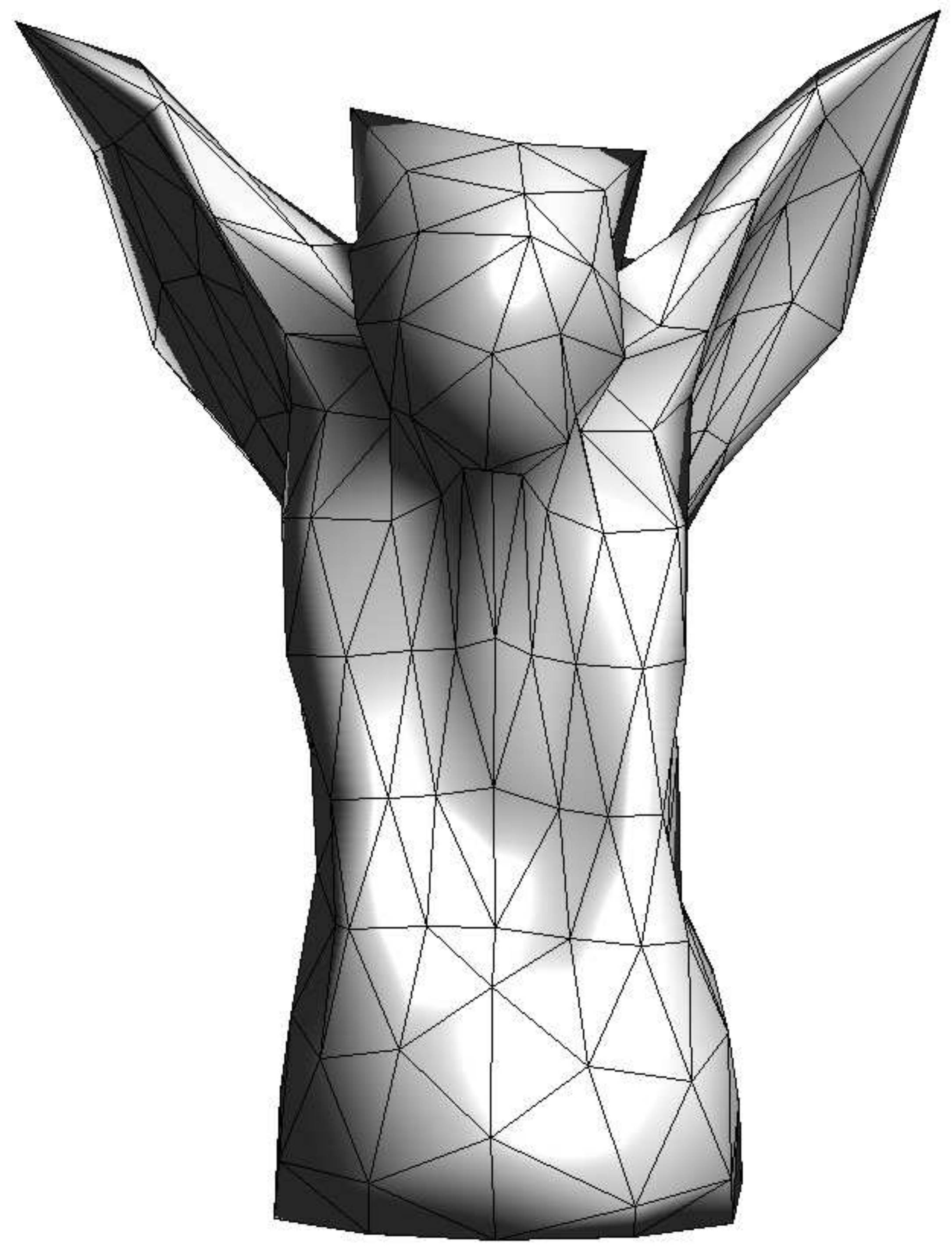}}}
  \subfigure[\label{fig-semiregular-5}]{
	{\includegraphics[height=5cm,keepaspectratio]{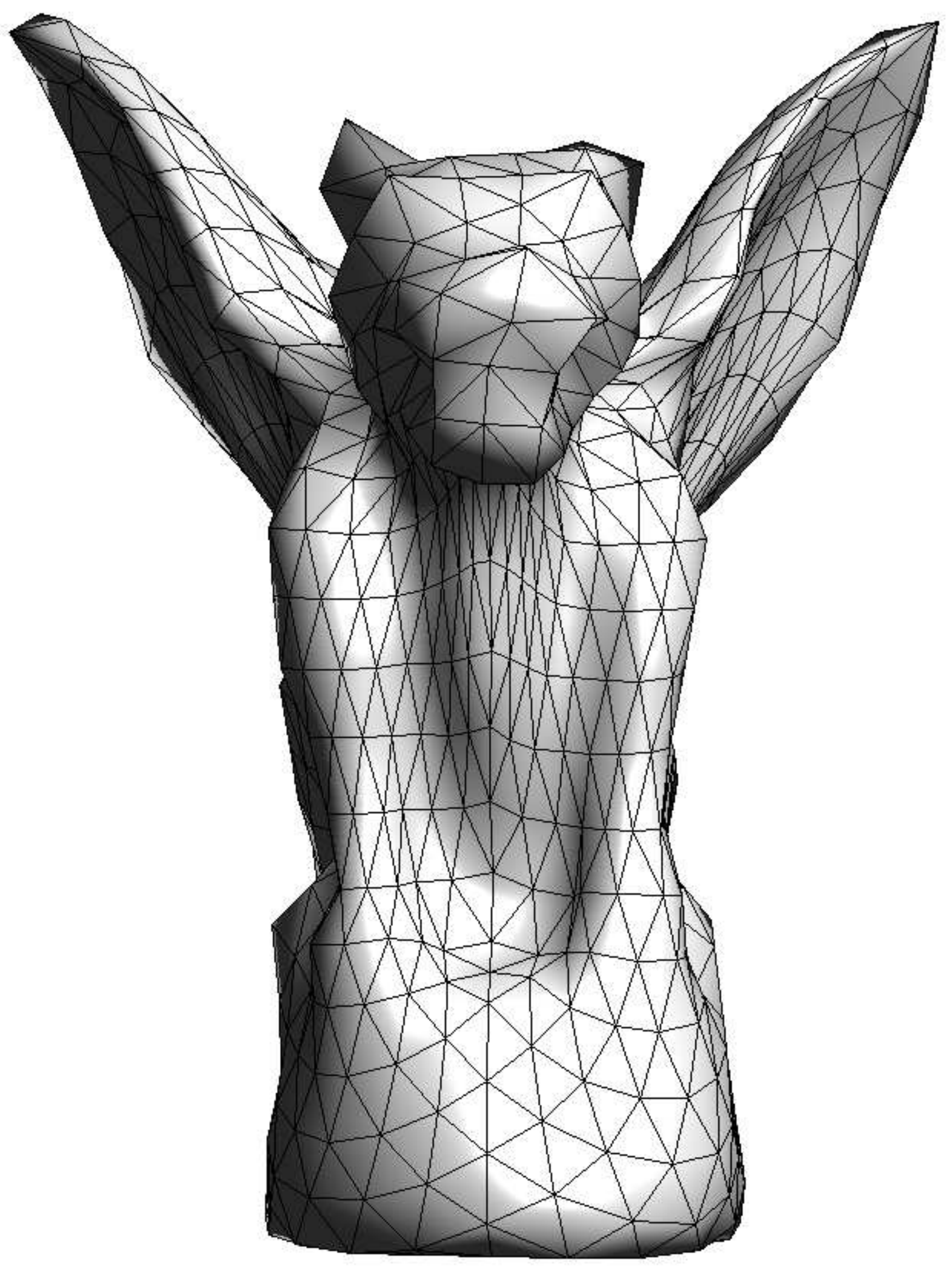}}}
  \subfigure[\label{fig-semiregular-6}]{
	{\includegraphics[height=5cm,keepaspectratio]{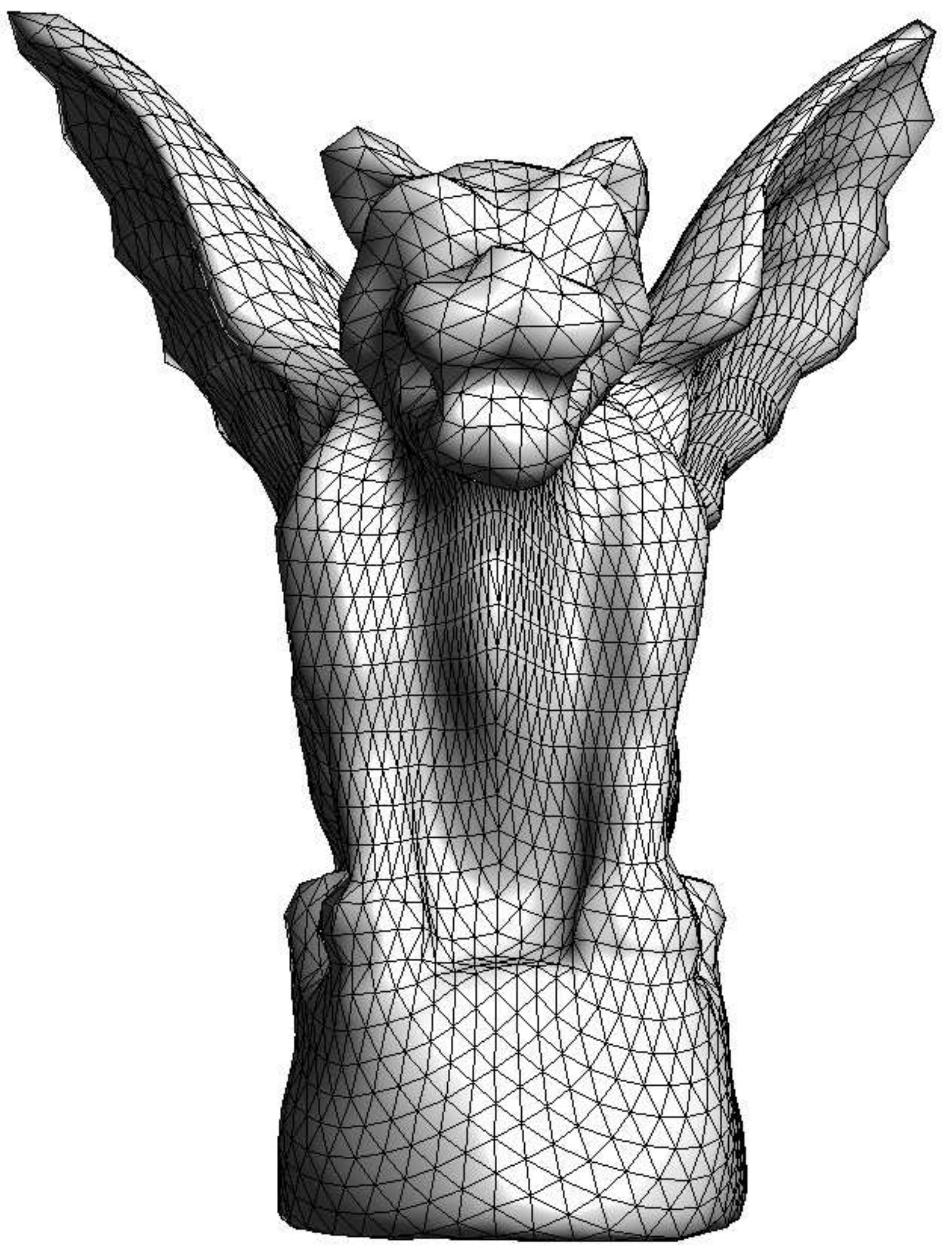}}}
  \subfigure[\label{fig-semiregular-7}]{
	{\includegraphics[height=5cm,keepaspectratio]{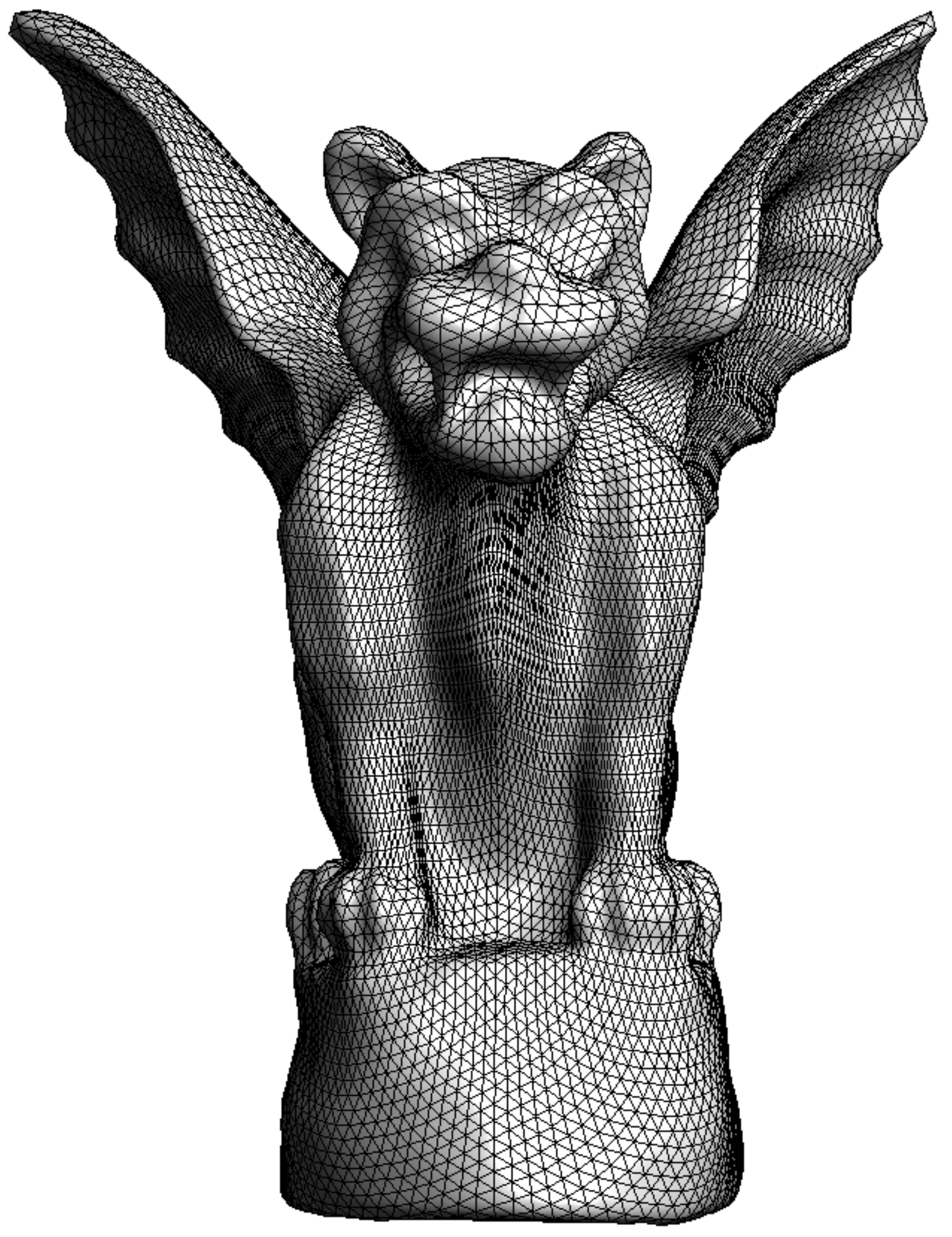}}}
  \subfigure[\label{fig-comp-4}]{
    \includegraphics[height=5cm,keepaspectratio]{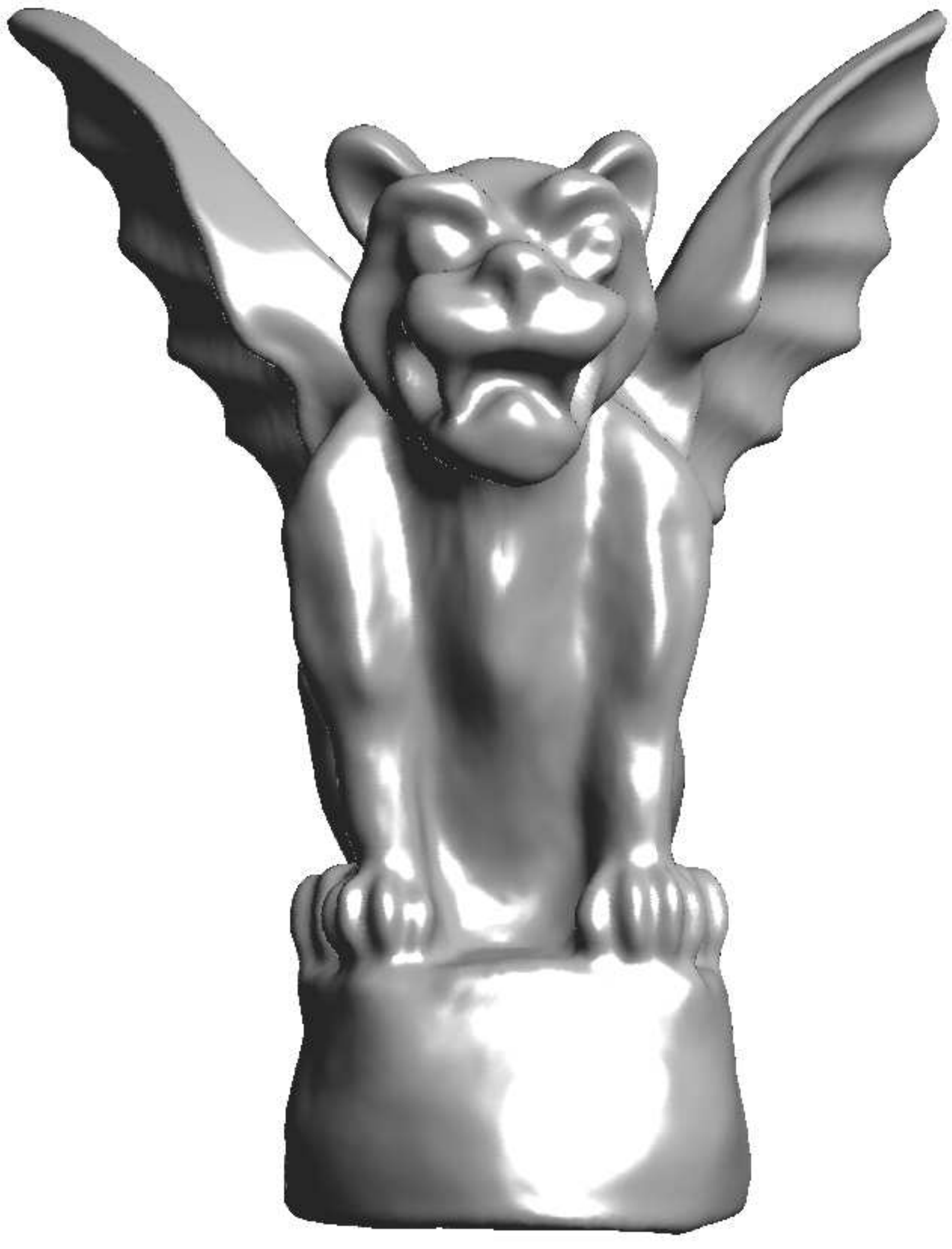}}
  \subfigure[\label{fig-comp-3}]{
    \includegraphics[height=5cm,keepaspectratio]{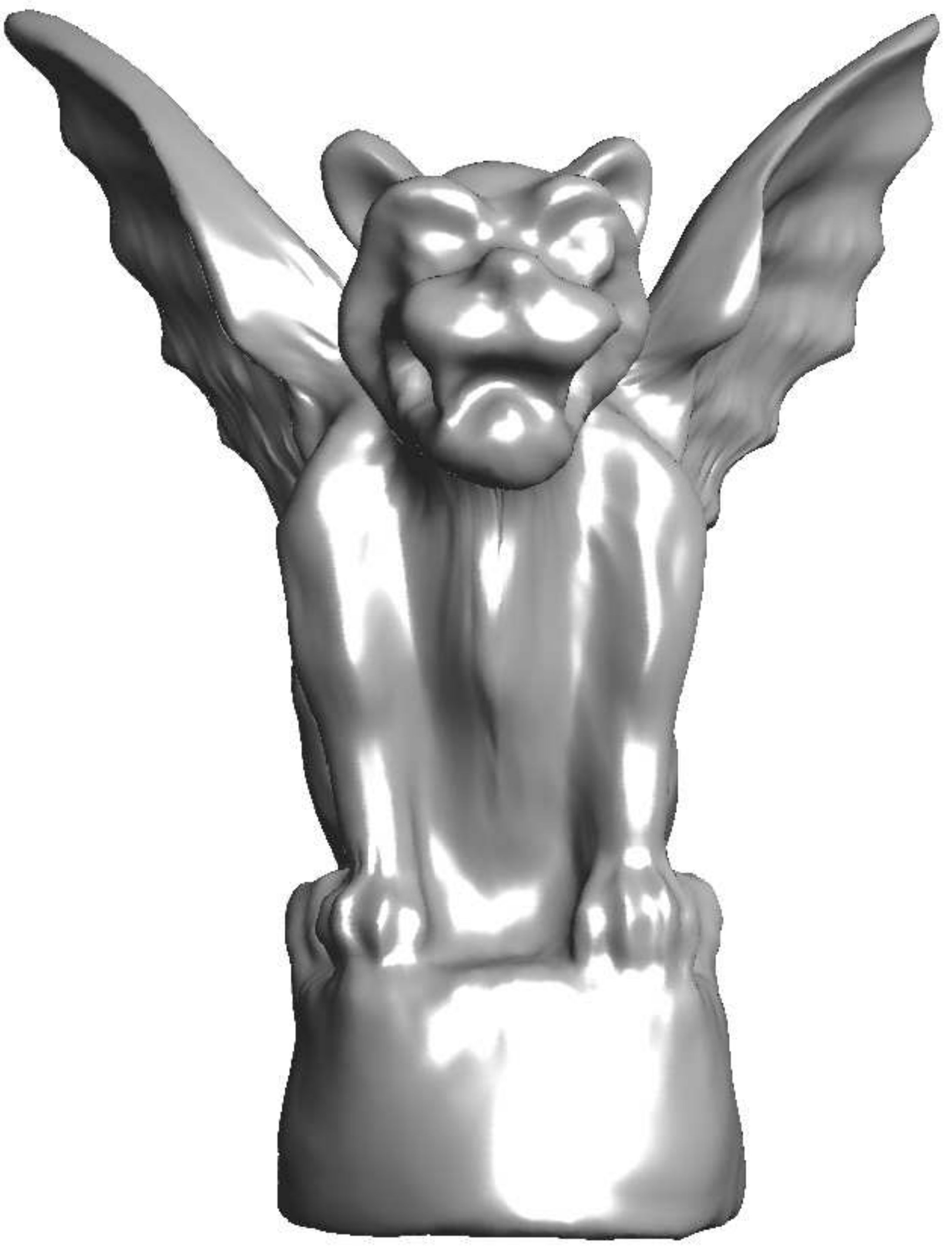}}
  \subfigure[\label{fig-comp-2}]{
    \includegraphics[height=5cm,keepaspectratio]{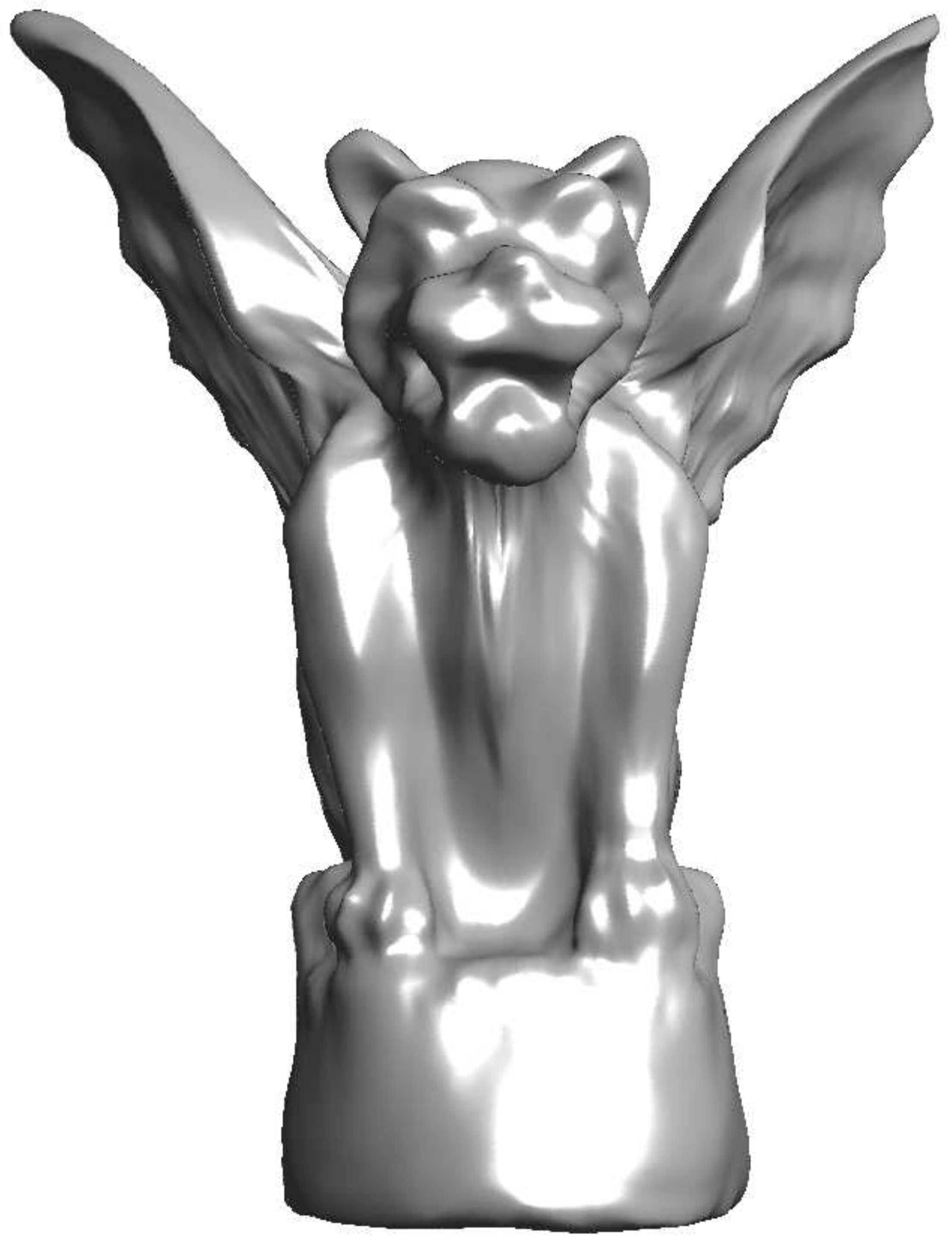}}
  \subfigure[\label{fig-comp-1}]{
    \includegraphics[height=5cm,keepaspectratio]{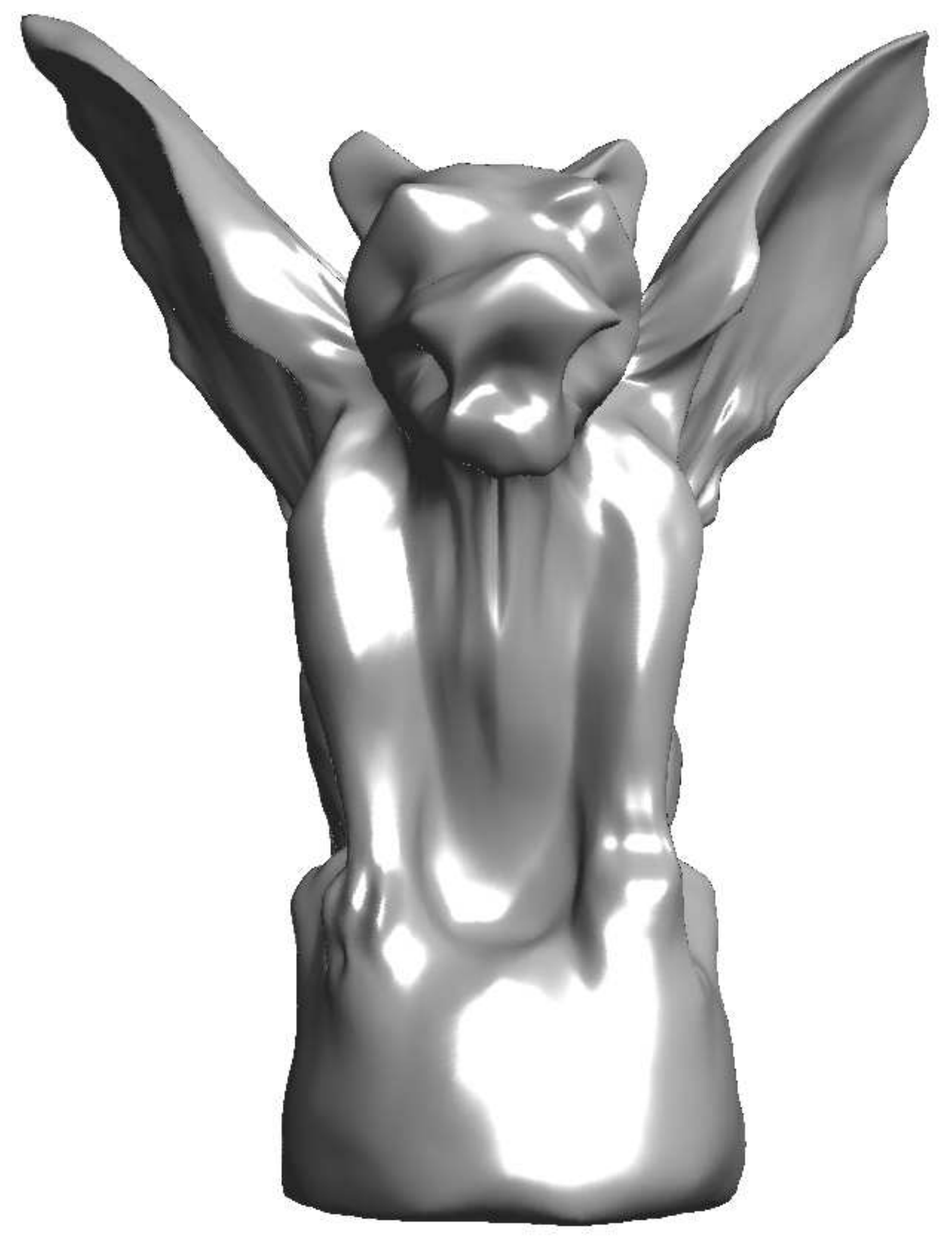}}
  \caption{Top row: example of semi-regular mesh $\{ \Mm^j \}_j$. Bottom row: example of surface approximation $f_M$ obtained by thresholding the lifted wavelets coefficients, where $N$ is the number of vertices in $\Mm^J$. (a) $\Mm^j, j=-4$. (b) $\Mm^j, j=-5$. (c) $\Mm^j, j=-6$. (d) $\Mm^j, j=-7$. (e) $M/N=100\%$. (f) $M/N=10\%$. (g) $M/N=5\%$ (h) $M/N=2\%$. \label{fig-semiregular}}
\end{figure}

Semi-regular meshes $\{ \Mm^j \}_{J \leq j \leq 0}$ are obtained by a
regular 1:4 refinement rule starting from an arbitrary control mesh
$\Mm^0$. Each edge of $\Mm^{j}$ is split into two sub-edges by vertex
insertion to obtain the refined mesh $\Mm^{j-1}$. The fine mesh
$\Mm^J$ is the sampling grid that stores the position of the surface
points in space, and a signal $f$ sampled at each grid point. Fig.~\ref{fig-semiregular}, top row, shows an example of such a
multiresolution mesh, obtained by a semi-regular remeshing of a high
resolution input mesh.

The lifting scheme described in Sec.~\ref{sec-lifting} can be
applied by storing the scaling coefficients $a_j$ on the grid point of
the mesh $\Mm^j$, while the detail coefficients are stored on the
complementary detail grid $\Dd^j$ where $\Mm^{j-1} = \Mm^j \cup
\Dd^j$. The splitting of $a_{j-1}$ into $a_j^o$ and $d_j^o$
corresponds to assigning the values stored in $\Mm^{j-1}$ to either
$\Mm^j$ or $\Dd^j$. The predict operator $P_j$ used to compute the
wavelet coefficients $d_j$ stored in $\Dd^j$ is a local polynomial
interpolator on a triangulation grid. The update operator $U_j$ is
computed by solving a linear system, to impose that moments of low
orders, such as the mean, are preserved when moving from $a_{j-1}$ to
$a_j$.

This lifting wavelet transform computes the coefficients $d_j[\bs n] =
\dotp{\psi_{(j,\bs n)}}{f}$ for all scales $0 < j < J$ and grid points
$\bs n \in \Dd^j$. It corresponds to the projection of the signal $f$
defined on the triangulated surface $\Mm^j$ onto a discrete
biorthogonal wavelet frame $\Basis = \{\psi_{(j,\bs n)}\}_{j,\bs
  n}$. These coefficients can be thresholded, and inverting the
lifting steps creates an approximated signal $f_M$ with $M$ non-zero coefficients.
Although this approach works well in practice, the frame bounds of the
resulting wavelet frame $\Basis$ are difficult to control, and $f_M$
might be far from the best $M$-terms approximation. It is also
difficult to guarantee the convergence of the wavelet atoms
$\psi_{(j,\bs n)}$ to smooth functions, when $J$ tends to $-\infty$, and
the mesh $\Mm^J$ approximates a smooth surface.

To perform surface approximation, one defines the signal $a_J$ at the finest 
scale as the position of the nodes on the surface. Each coefficient $a_J[\bs n] \in \RR^3$
is thus a point in 3D space. The lifting transform can be applied to 
this vector-valued signal. Thresholding the resulting wavelet coefficients 
allows one to approximate the surface using few coefficients, as shown
on Fig.~\ref{fig-semiregular}, bottom row.
If the lifting operators $P_j$ and $U_j$ do not depend
on the position of the points on the surface, the resulting lifting 
wavelets can be used to perform 3D mesh compression~\cite{Lounsbery_M_1997_j-acm-tog_mul_asatt,Schroder_P_1995_p-acm-siggraph_sph_werfs}.

\subsection{Wavelets on Graphs}
\label{sec:wavelet-graph}

Let us finally mention that wavelet transform has been extended to
functions defined on the vertices of an arbitrary finite weighted
graph. The latter may for instance generalize standard picture definition by  describing two-dimensional pixel adjacencies. Maggioni \etal introduced ``diffusion
wavelets'' \cite{Coifman_R_2006_j-acha_diff_w}, a general
theory for wavelet decompositions based on compressed representations
of powers of a diffusion operator such as the graph Laplacian. The
constructed wavelet basis is made orthogonal by combining graph
subsampling and Gram-Schmidt orthogonalization on each subsampled
space.

More recently, Hammond \etal \cite{Hammond_2011_j-acha_wav_gsgt}
developed a general wavelet frame theory on such graphs thanks to the
graph analogue of the Fourier domain, namely the spectral
decomposition of the discrete graph Laplacian. Wavelets are defined in
this frequency domain by dilating an ``admissible'' generating
kernel. The final representation is redundant but wavelets can be shaped by
changing the generating kernel. Moreover, for sparse graph Laplacian
matrix, a fast wavelet transform avoiding the Laplacian spectral
decomposition is developed.

\section{Conclusion}
\label{sec:conclusion}

A century after the discovery by Alfr\'ed Haar, and twenty years after
the emergence of wavelets as genuine processing tools,
major advances  have been made in the improvement of natural images
representations, aiming at enhanced understanding.

Their common characteristic resides in uncovering multiscale and
oriented features of natural images, through projections on a specific
set of elongated atoms. The resulting dictionaries are thus often
redundant, and may be coupled with sparsity enforcing priors, or
adaptivity. They reveal a striking similarity with low level vision,
where similar strategies are used to build powerful processing
architectures.

The availability of such a large number of transformations, that
potentially extend the standard wavelet framework, leaves open the
question of the best representation to process a given image. This
choice is unfortunately data dependent, since the geometry of edges
and textures varies significantly from natural to seismic or medical
images. Selection of a representation, as well as its parameterization
(number of scales, span of orientations, support in space or
frequency), is also application dependent, and applications to inverse
problems or pattern recognition typically impose strong design
requirements on the dictionary. Their exhaustive comparison thus
remains out of reach, with traditional methods from image processing
or approximation theory only providing a partial answer.

As a humble contribution to a subjective comparison, additional
materials, full scale decomposition images, related links and
associated toolboxes necessary to reproduce  illustrations provided
in this paper are available at 
\cite{Jacques_L_2011_url_panorama_addendum}.  Oddly enough, a common
etymology of Szeged resides in an old Hungarian word for
\emph{corner} (\emph{szeg}).  At a turn in a wavelet century, A. Haar and F. Riesz
might not have foreseen the harvest from their mathematical seeds.
Image understanding is at the beginning of reaping their fruits.

\section*{Acknowledgements}

Laurent Jacques is a postdoctoral researcher funded by the Belgian
National Science Foundation (F.R.S.-FNRS).  Professor K\'aroly
Szatm\'ary is warmly acknowledged for providing us with the recurrent
picture \imageHRP. 
We warmly thank
Pedro Corea (ICTEAM-TELE, UCL), J\'er\^ome Gauthier (CEA) and Gr\'egoire Hertz (Sup\'elec) for their thorough proofreading. The four
authors are also very grateful to the anonymous reviewers whose
important remarks and suggestions have greatly improve the quality of
this paper. They finally are indebted to  Jean-Pierre Antoine, Nick Kingsbury,  Fran\c{c}ois Meyer, Ivan Selesnick and guest editors Thierry Blu, Jean-Christophe Pesquet and Truong Q. Nguyen  for their constructive and insightful discussions and comments.

\end{document}